\documentclass[final,3p,times,twocolumn,authoryear]{elsarticle}
\usepackage{amssymb}
\usepackage{amsmath}
\usepackage{booktabs}
\usepackage{multirow}
\usepackage{adjustbox}
\usepackage{pifont}% http://ctan.org/pkg/pifont
\newcommand{\cmark}{\ding{51}}%
\begin{document}

\begin{frontmatter}

%% Title, authors and addresses

%% use the tnoteref command within \title for footnotes;
%% use the tnotetext command for theassociated footnote;
%% use the fnref command within \author or \affiliation for footnotes;
%% use the fntext command for theassociated footnote;
%% use the corref command within \author for corresponding author footnotes;
%% use the cortext command for theassociated footnote;
%% use the ead command for the email address,
%% and the form \ead[url] for the home page:
%% \title{Title\tnoteref{label1}}
%% \tnotetext[label1]{}
%% \author{Name\corref{cor1}\fnref{label2}}
%% \ead{email address}
%% \ead[url]{home page}
%% \fntext[label2]{}
%% \cortext[cor1]{}
%% \affiliation{organization={},
%%            addressline={}, 
%%            city={},
%%            postcode={}, 
%%            state={},
%%            country={}}
%% \fntext[label3]{}

\title{Deep Learning-Based Facial Expression Recognition for the Elderly: A Systematic Review}

\author[1]{F. Xavier Gaya-Morey\corref{cor1}}
\ead{francesc-xavier.gaya@uib.es}
% \ead[url]{https://orcid.org/0000-0003-1231-7235}
% \credit{Conceptualization, Methodology, Validation, Investigation, Writing - Original Draft, Writing - Review \& Editing Preparation, Visualization}

\affiliation[1]{%
    organization={Universitat de les Illes Balears},
    addressline={Carretera de Valldemossa, km 7.5}, 
    city={Palma, Illes Balears},
    postcode={07122}, 
    country={Spain}}

% Second author

% Third author
\author[1]{Jose M. Buades-Rubio}
\ead{josemaria.buades@uib.es}
% \ead[url]{https://orcid.org/0000-0002-6137-9558}
% \credit{Conceptualization, Methodology, Writing - Review \& Editing Preparation, Supervision, Project administration, Funding acquisition}

\author[2]{Philippe Palanque}
\ead{philippe.palanque@irit.fr}

\affiliation[2]{%
    organization={ICS-IRIT, University Toulouse 3, Paul Sabatier},
    addressline={118 Rte de Narbonne}, 
    city={Toulouse},
    postcode={31062}, 
    country={France}}

\author[3]{Raquel Lacuesta}
\ead{lacuesta@unizar.es}

\affiliation[3]{%
    organization={Universidad de Zaragoza},
    addressline={C. de Pedro Cerbuna, 12}, 
    city={Zaragoza, Aragón},
    postcode={50009}, 
    country={Spain}}
    
\author[1]{Cristina Manresa-Yee}
\ead{cristina.manresa@uib.es}
% \ead[url]{https://orcid.org/0000-0002-8482-7552}
% \credit{Conceptualization, Methodology, Writing - Review \& Editing Preparation, Supervision, Project administration, Funding acquisition}
    
% Corresponding author text
\cortext[cor1]{Corresponding author}

%% Abstract
\begin{abstract}
The rapid aging of the global population has highlighted the need for technologies to support elderly, particularly in healthcare and emotional well-being. Facial expression recognition (FER) systems offer a non-invasive means of monitoring emotional states, with applications in assisted living, mental health support, and personalized care. This study presents a systematic review of deep learning-based FER systems, focusing on their applications for the elderly population. Following a rigorous methodology, we analyzed 31 studies published over the last decade, addressing challenges such as the scarcity of elderly-specific datasets, class imbalances, and the impact of age-related facial expression differences. Our findings show that convolutional neural networks remain dominant in FER, and especially lightweight versions for resource-constrained environments. However, existing datasets often lack diversity in age representation, and real-world deployment remains limited. Additionally, privacy concerns and the need for explainable artificial intelligence emerged as key barriers to adoption. This review underscores the importance of developing age-inclusive datasets, integrating multimodal solutions, and adopting XAI techniques to enhance system usability, reliability, and trustworthiness. We conclude by offering recommendations for future research to bridge the gap between academic progress and real-world implementation in elderly care.
\end{abstract}

% %%Graphical abstract
% \begin{graphicalabstract}
% %\includegraphics{grabs}
% \end{graphicalabstract}

%%Research highlights
% \begin{highlights}
%     \item Systematic review of deep learning facial expression recognition systems in elderly
%     \item Most datasets suffer from elderly underrepresentation and class imbalance
%     \item Few studies include privacy protection methods
%     \item Most studies disregard the use of explainable artificial intelligence tools
%     \item Real-world deployment remains insufficiently explored
% \end{highlights}

%% Keywords
\begin{keyword}
%% keywords here, in the form: keyword \sep keyword
facial expression recognition \sep
deep learning \sep 
computer vision \sep 
elderly

%% PACS codes here, in the form: \PACS code \sep code

%% MSC codes here, in the form: \MSC code \sep code
%% or \MSC[2008] code \sep code (2000 is the default)

\end{keyword}

\end{frontmatter}

%% Add \usepackage{lineno} before \begin{document} and uncomment 
%% following line to enable line numbers
%% \linenumbers

%% main text
%%

\section{Introduction}

    The global population is aging at an unprecedented rate, with significant implications for multiple aspects of society, including healthcare, social services, and the economy \citep{bloom2016chapter}. According to the World Population Aging report \citep{WHO2024}, there will be 265 million persons aged 80 years or older by the mid-2030s, more than the number of infants (1 year of age or less), and by the late 2070s, the number of persons at ages 65 years and higher is projected to reach 2.2 billion, surpassing the number of children (under age 18). This shift in demographics, driven by increasing life expectancy and declining fertility rates, particularly in developed countries, is creating a significant demand for innovations in elderly care. As older adults are more susceptible to physical and cognitive health conditions, including dementia, Alzheimer’s disease, and other age-related illnesses \citep{langa2018cognitive}, they often require specialized care. Providing such care on a large scale poses unique challenges, calling for the development of technologies that can assist with monitoring, diagnosing, and enhancing the well-being of this rapidly growing segment of the population with very specific needs.

    One area where technology can have a significant impact is in the real-time recognition of emotional states in elderly individuals. Facial expression recognition (FER) systems offer a non-invasive means to monitor emotional well-being and identify potential mental health issues early on \citep{fei2019survey}. FER, in particular, has become an important tool in healthcare, enabling caregivers and medical professionals to better understand the emotional states of their patients and respond accordingly. Social robots, for example, have been shown to increase social engagement among the elderly and can help reduce loneliness and depression by providing emotional support \citep{joan2015what, wang2014towards}. By incorporating FER into these systems, they can interact more naturally with older adults, helping to recognize emotions and respond in ways that are empathetic and supportive. However, despite the growing need for such technology, research in FER focused on the elderly population remains underrepresented in the literature \citep{ma2019elderreact}. Most existing studies on FER target adults and younger adults, leaving a significant gap in understanding how to optimize these systems for older adults, whose facial features may change due to age-related factors \citep{ko2021changes}, making emotion recognition more challenging \citep{mary2016review}.
    
    The foundation of FER systems largely stems from the work of psychologist Paul Ekman, who identified six basic facial expressions—happiness, sadness, fear, anger, surprise, and disgust—that are universally recognized across cultures \citep{ekman1992argument}. These six expressions have since formed the basis of most emotion recognition systems, despite ongoing debates about their universality and the ability of facial expressions to fully capture the complexity of human emotions \citep{barrett2019emotional}. For older adults, FER systems have the potential to greatly improve quality of life by facilitating better communication between caregivers and patients, supporting emotional monitoring, and enabling more personalized care. FER can assist across an array of domains, including medical diagnosis and treatment \citep{grabowski2019emotional}, enabling timely intervention, which is particularly important for individuals suffering from cognitive decline, where emotional expression may be the primary means of communication.
    
    In recent years, Deep Learning (DL) has revolutionized the field of computer vision, including FER \citep{omahony2020deep}. DL techniques, especially Convolutional Neural Networks (CNNs), have demonstrated remarkable performance in recognizing facial expressions by automatically learning features from large datasets without the need for extensive manual feature engineering \citep{li2022deep}. The success of DL in tasks such as object detection, speech recognition, and natural language processing has motivated its application to FER, where it has achieved state-of-the-art performance. One of the key advantages of DL models is their ability to generalize well to complex data, such as facial images, which often involve variations in lighting, pose, and expression. For FER in elderly populations, DL offers the possibility of overcoming the challenges posed by age-related changes in facial structure, allowing for more accurate recognition across diverse populations and over a long period of time for the same population.
    
    However, while DL-based FER systems hold significant promise, they also come with challenges. A key limitation of DL models is their "black-box" nature, meaning that the decision-making process of these models is often opaque to users. In high-stakes applications, such as healthcare, understanding how a model arrives at a particular prediction is crucial for building trust in the system and ensuring its reliability \citep{barredoarrieta2020explainable, adadi2018peeking}. For instance, a medical professional may need to know why a system predicts that a patient is exhibiting signs of depression or distress before making treatment decisions based on that prediction. This lack of transparency in DL models has led to the rise of explainable artificial intelligence (XAI), a field that seeks to make AI systems more interpretable and understandable to humans \citep{barredoarrieta2020explainable}. XAI techniques aim to shed light on the internal workings of complex models by providing explanations for their predictions \citep{gunning2019xai—explainable}. In the context of FER for the elderly, XAI methods can offer caregivers insights into why a system interprets certain facial expressions in specific ways, allowing them to better assess the emotional well-being of their patients. Moreover, the integration of XAI can enhance the development of FER systems by allowing researchers to identify and mitigate biases in the data or model, leading to more robust and reliable systems \citep{burkart2021survey}. Thus, the adoption of XAI techniques is essential for making DL-based FER systems not only effective but also trustworthy and accountable. Furthermore, these techniques can provide insights into expression differences influenced by age.

    This paper presents a systematic review of the current state of DL-based facial expression recognition systems, with a specific focus on their applications for elderly people. The review follows the guidelines for conducting a systematic literature review (SLR) in software engineering, as outlined by \cite{kitchenham2007guidelines}, which provide a comprehensive framework for ensuring rigor in methodology, structure, and best practices. The structure of the paper is as follows: first, we summarize previous reviews on FER, alongside background information emphasizing the effects of aging in FER, highlighting the need for specialized solutions for this population. Next, we introduce the review questions. Following that, the methodology of the review is detailed, including data sources, the search strategy, criteria for study selection, quality assessment procedures, and the process of data extraction. Subsequently, the list of selected studies is thoroughly analyzed to provide insights that address the review questions. In the discussion section, we reflect on the key findings, outline the identified strengths and weaknesses of the analyzed studies, and offer recommendations for future works on the subject. Finally, last section of the paper presents the conclusions drawn from the SLR.

\section{Related Work}

    In this section, we examine the literature to identify studies that assess the impact of age on FER tasks, as well as prior reviews on FER. Our objective is twofold: first, to underscore the lack of FER reviews focused on specific age groups, particularly the elderly, and second, to provide evidence of the age bias often introduced by age-agnostic approaches. This analysis aims to highlight the need for age-sensitive methodologies in FER and to emphasize the importance of addressing this gap in existing reviews.

    \subsection{Effects of Aging on Facial Expression Recognition}

        Numerous psychological studies have explored the effects of aging on facial expression recognition, demonstrating that observers are influenced by their own age. For example, older individuals tend to exhibit deficits in decoding specific emotions \citep{isaacowitz2011bringing, ruffman2008meta}. However, aging impacts not only the observers but also the individuals displaying the target expressions. In this regard, \cite{fölster2014facial} examined how age-related changes in the face, such as wrinkles and folds, influence the decoding process of emotional expressions, concluding that the age of the face is a critical factor. This finding is supported by subsequent studies, such as \cite{ko2021changes}, which identified variations in facial expression intensities and muscle usage across different age groups. For instance, elderly individuals tend to display more negative emotions and engage more muscles in the lower face compared to younger people. Similarly, \cite{ngrondhuis2021having} used generative adversarial networks (GAN) to investigate the increased difficulty in identifying expressions of older adults and attributed this challenge to the decline in facial muscle function with age. \cite{battinisonmez2019computational} explored the effect of training FER models with data from different age groups and found that recognizing expressions in elderly faces posed the greatest difficulty. A review by \cite{raghebatallah2019review} on the effect of facial aging on FER identified several open challenges, including the scarcity of images capturing the same individuals across different ages (which hinders the learning of aging patterns), the tendency of models to overlook important facial features, and the considerable variation in aging effects across subjects. Moreover, the training of a Siamese CNN on the ElderReact and EmoReact datasets by \cite{rahatuljannat2021expression} highlighted differences in the expressions of elderly individuals and children. \cite{park2022facial} demonstrated the importance of addressing age bias in FER, showing that age-specific training yielded a 22\% improvement in accuracy compared to using a non-age-specific dataset.

        Automatic FER systems are heavily influenced by the age of the faces in the dataset, making the choice of dataset critical. Datasets such as FACES \citep{ebner2010faces} and LifeSpan \citep{minear2004lifespan}, are frequently employed \citep{guo2013facial, mary2016review, wu2015enhanced, wang2015facial, al_garaawi2016study, lopes2018facial, battinisonmez2019computational, caroppo2017facial, caroppo2019facial, caroppo2020comparison, al_garaawi2022fully}, largely due to their inclusion of subjects from a broad age range, which facilitates the development of age-invariant models. To address the effects of aging, several strategies have been proposed. One approach involves removing aging features through facial smoothing techniques that eliminate age-related details without compromising essential structural information \citep{guo2013facial, mary2016review}. Another method incorporates age information during training using Bayesian networks, while marginalizing over age during testing \citep{wu2015enhanced, wang2015facial}. \cite{al_garaawi2016study} first analyzed age-related differences in facial characteristics, then later incorporated age as a key feature in their model \citep{al_garaawi2022fully}, using a weighted combination of age group estimators and age-specific expression recognizers. 
        
        Other studies have focused on specific age groups rather than addressing aging effects comprehensively. Datasets utilized in such approaches include ElderReact \citep{ma2019elderreact} and Tsinghua \citep{yang2020tsinghua} for elderly subjects, CK+ \citep{lucey2010extended}, JAFFE \citep{lyons2020coding}, AFEW \citep{dhall2007collecting}, and FER-2013 \citep{goodfellow2013challenges} for adults, and LIRIS \citep{khan2019novel}, CAFE \citep{lobue2014child}, DEFSS \citep{meuwissen2017creation}, and EmoReact \citep{nojavanasghari2016emoreact} for children, among others.

    \subsection{Summary of Previous Reviews}
        % Background 
        
        Mining the literature, we identified several existing reviews addressing facial expression recognition. Although most of these do not focus specifically on elderly populations and therefore do not consider the effects of human aging discussed in the previous section, they do provide valuable insights into the datasets and techniques commonly employed in FER. In this section, we analyze relevant reviews and surveys from the past decade, highlighting that many critical aspects addressed in our current study have not been thoroughly examined in prior reviews.
    
        \begin{table*}
            \centering
            % \begin{adjustbox}{width=.65\textwidth}
            \begin{tabular*}{.71\textwidth}{l|ll|ll|lllll}
                 & \multicolumn{2}{c|}{\textbf{Process}} & \multicolumn{2}{c|}{\textbf{Focus}} &\multicolumn{5}{c}{\textbf{Researched data}} \\
                 \multicolumn{1}{c|}{} & \rotatebox{90}{Systematic} & \rotatebox{90}{Prev. reviews} & \rotatebox{90}{Elderly} & \multicolumn{1}{c|}{\rotatebox{90}{DL}} & \rotatebox{90}{Datasets} & \rotatebox{90}{Multimodality} & \rotatebox{90}{Deployment} & \rotatebox{90}{Privacy} & \rotatebox{90}{XAI}\\ \midrule
                \cite{ghayoumi2017quick} &  &  &  & \cmark &  &  &  &  \\
                \cite{asad2017recent} &  &  &  &  &  &  &  &  &  \\
                \cite{chulko2018brief} &  & \cmark &  &  & \cmark &  &  &  &  \\
                \cite{rajeswari2018literature} &  &  &  &  &  &  &  &  &  \\
                \cite{yantililiana2018review} &  &  &  &  &  &  &  &  &  \\
                \cite{fei2019survey} &  &  & \cmark &  &  &  &  &  &  \\
                \cite{bhattacharya2019survey} &  &  &  &  &  &  &  &  &  \\
                \cite{martinez2019automatic} &  & \cmark &  &  & \cmark &  &  &  &  \\
                \cite{canedo2019facial} & \cmark &  &  &  & \cmark & \cmark &  &  &  \\
                \cite{achinchanikar2019facial} &  &  &  & \cmark & \cmark &  &  &  &  \\
                \cite{svyas2019survey} &  &  &  &  & \cmark &  &  &  &  \\
                \cite{sari2020automated} &  &  &  & \cmark & \cmark &  &  &  &  \\
                \cite{li2022deep} &  &  &  & \cmark & \cmark & \cmark &  &  &  \\
                \cite{patel2020facial} &  & \cmark &  &  & \cmark &  &  &  &  \\
                \cite{ribeiroalexandre2020systematic} & \cmark & \cmark &  &  & \cmark & \cmark &  &  &  \\
                \cite{pranathi2021review} &  &  &  &  & \cmark &  &  &  &  \\
                \cite{dalvi2021survey} & \cmark & \cmark &  &  & \cmark & \cmark &  &  &  \\
                \cite{revina2021survey} &  &  &  &  & \cmark &  &  &  &  \\
                \cite{moolchandani2021survey} &  &  &  &  & \cmark &  &  &  &  \\
                \cite{ullah2021systematic} & \cmark &  &  &  & \cmark &  &  &  &  \\
                \cite{muazu2021systematic} &  &  &  &  & \cmark &  &  &  &  \\
                \cite{msaleemabdullah2021facial} &  &  &  & \cmark &  &  &  &  &  \\
                \cite{moehtay2021feature} &  &  &  &  & \cmark &  &  &  &  \\
                \cite{raijain2021recent} &  &  &  &  & \cmark & \cmark &  &  &  \\
                \cite{modi2022state-of-the-art} &  &  &  &  &  &  &  &  &  \\
                \cite{zagocanal2022survey} & \cmark & \cmark &  &  & \cmark &  &  &  &  \\
                \cite{azlinaabaziz2022systematic} & \cmark &  &  &  & \cmark &  & \cmark &  &  \\
                \cite{maithri2022automated} & \cmark &  &  &  & \cmark & \cmark &  &  &  \\
                \cite{rehmankhan2022facial} &  & \cmark &  &  & \cmark &  &  &  &  \\
                \cite{guerdelli2022macro} &  &  &  &  & \cmark & \cmark &  &  &  \\
                \cite{rashmiadyapady2023comprehensive} &  &  &  &  &  &  &  & \cmark &  \\
                \cite{liang2023survey} &  &  &  & \cmark & \cmark & \cmark &  &  &  \\
                \cite{labzour2023survey} &  &  & \cmark &  & \cmark &  &  &  &  \\
                \cite{vinicioslopespinto2023systematic} & \cmark &  &  & \cmark & \cmark &  &  &  &  \\
                \cite{boughanem2023facial} &  &  &  & \cmark & \cmark & \cmark &  &  &  \\
                \cite{chitleong2023facial} & \cmark & \cmark &  &  &  &  & \cmark &  &  \\
                \cite{almasoudi2023facial} &  &  &  &  & \cmark &  &  &  &  \\
                \cite{kumari2024emotion} &  &  &  & \cmark & \cmark &  &  &  &  \\
                \cite{kaur2024facial} & \cmark & \cmark &  &  & \cmark & \cmark &  &  &  \\
                \cite{mohana2024facial} & \cmark & \cmark &  &  & \cmark &  &  &  &  \\
                This study & \cmark & \cmark & \cmark & \cmark & \cmark & \cmark & \cmark & \cmark & \cmark \\
                \bottomrule
            \end{tabular*}
            % \end{adjustbox}
            \caption{Comparison of previous reviews with our work, considering the key aspects of the current systematic review. The columns represent the following criteria, from left to right: whether the review is systematic; it includes an exploration of prior reviews; it focuses on the elderly population; it focuses on DL-based approaches; it provides a list of datasets; it investigates the use of multimodal data; it examines the deployment of FER systems in real-world environments; it considers privacy a critical issue; and it discusses the application of XAI techniques in FER systems.}
            \label{tab:review-comparison}
        \end{table*}
        
        To begin with, it is important to consider the methodological rigor of prior review processes. A systematic review methodology promotes a thorough and unbiased selection and analysis of studies by precisely defining information sources, search strategies, study selection criteria, and data extraction procedures. Additionally, quality assessment steps are essential to ensure that only high-quality, relevant studies are included. However, only 12 of the 41 reviews listed in Table \ref{tab:review-comparison} followed a systematic review process. \cite{kitchenham2007guidelines} also emphasize the importance of analyzing existing reviews in the field to identify gaps in the literature and avoid duplicate efforts. Yet, only 11 studies accounted for previous reviews, as illustrated in the table.
        
        The main distinction between the current study and earlier reviews lies in our focus on FER applications for the elderly and deep learning tools. Only two previous reviews have concentrated specifically on elderly populations. \cite{labzour2023survey} discussed 11 studies on elderly-focused FER, covering both traditional and deep learning methods and their associated datasets. \cite{fei2019survey} also reviewed FER in elderly populations, primarily as a tool for early diagnosis of mild cognitive impairment. However, their work lacks coverage of recent studies from the last five years and focuses primarily on traditional computer vision approaches rather than DL-based methods. Consequently, there is a gap in the literature regarding recent DL-based FER research specific to elderly individuals.
        
        Ten reviews were identified that discuss DL in FER \citep{ghayoumi2017quick, achinchanikar2019facial, sari2020automated, li2022deep, msaleemabdullah2021facial, liang2023survey, vinicioslopespinto2023systematic, boughanem2023facial, kumari2024emotion}. Of these, only the work by Pinto et al. employed a systematic review process, and none of these reviews examined previous FER reviews, which may have contributed to overlapping lists of studies. While many of these reviews included extensive lists of commonly used datasets--an important consideration in DL-based approaches--none were focused on elderly populations, nor did they address critical aspects for the current work such as deployment in real-world environments, privacy preservation, or explainable AI techniques.
        
        Three critical aspects—deployment in real--world settings, privacy preservation, and XAI techniques--remain largely unexplored in prior reviews. Deployment in real-world scenarios was examined in only two reviews, which described applications involving mobile apps and robots \citep{chitleong2023facial} and smart home integration \citep{azlinaabaziz2022systematic}. Although ethical concerns regarding privacy were noted in two reviews \citep{kaur2024facial, almasoudi2023facial}, only one review explored this issue in depth \citep{rashmiadyapady2023comprehensive}. Furthermore, despite the growing importance of XAI, this area was not thoroughly investigated in any review, even though some listed it as a future challenge \citep{mohana2024facial, kaur2024facial, patel2020facial}.
        
        Given the evident lack of reviews focused on elderly populations using DL-based techniques for FER, the present study aims to address this gap. Additionally, as few reviews have examined factors critical for the practical deployment of FER systems in real-world environments, and none have explored XAI in detail, our work may offer valuable insights for the development of practical, reliable applications.

\section{Review Questions}
    % Identify primary and secondary review questions. Note this section may be included in the background section.

    We identified two primary review questions, each associated with multiple secondary questions, as shown in Table \ref{tab:questions}.

    \begin{table*}[h]
        \label{tab:questions}
        % \begin{adjustbox}{width=\textwidth}
        \centering
        \begin{tabular*}{.81\textwidth}{ll}
            \toprule
            \textbf{ID} & \textbf{Research Question} \\
            \midrule
            \textbf{RQ1} & 
            \textbf{What DL techniques are applied for FER in elderly populations?}\\
            
            RQ1.1 & 
            Which architectures are most commonly used?\\
    
            RQ1.2 & 
            Which datasets are most frequently employed?\\
    
            RQ1.3 & 
            What are the primary characteristics of the data utilized?\\
    
            RQ1.4 & 
            To what extent are facial landmarks and action units used?\\
    
            RQ1.5 & 
            What other tasks are commonly computed alongside FER?\\
            
            \midrule
    
            \textbf{RQ2} & 
            \textbf{How can these techniques be effectively deployed in real-world environments?}\\

            RQ2.1 &
            Are proposed solutions being deployed?\\
    
            RQ2.2 &
            Are aging biases addressed?\\
    
            RQ2.3 & 
            Is privacy a primary concern?\\
    
            RQ2.4 & 
            Is economic cost a primary consideration?\\
    
            RQ2.5 & 
            Are XAI techniques employed?\\
            
            \bottomrule
        \end{tabular*}
        % \end{adjustbox}
        \caption{Primary and secondary research questions used for this SLR.}
    \end{table*}
    
    RQ1 aims to find which deep learning techniques are applied specifically to facial expression recognition in elderly populations, and understand how are they used. The associated secondary questions delve into particular facets of these techniques, such as the prevalent architectures and datasets. Additionally, the questions examine data characteristics (e.g., image or video format, and multimodal integration) and assess the role of facial landmarks and action units, which are common features in FER research.

    In contrast, RQ2 focuses on practical considerations for deploying FER systems for elderly users in real-world settings. We investigate factors crucial to successful deployment, including the potential impact of age-related biases in training data that could lead to suboptimal performance in real-life scenarios. We also assess privacy and economic considerations. Finally, given the critical need for transparency and trust in DL applications within healthcare settings \citep{adadi2018peeking}, we explore the extent to which explainable AI techniques are integrated into these systems.

\section{Review Methods}

    In this section, we provide a detailed description of the systematic review process, adhering to the guidelines outlined by \cite{kitchenham2007guidelines}. First, we present the data sources and search strategy employed to compile the initial list of studies. Next, we explain the quality assessment and study selection procedures used to exclude low-quality and irrelevant works. Lastly, we outline the information extracted from each selected study.

    \subsection{Data sources}

\begin{figure}[h]
             \centering
             \includegraphics[width=\columnwidth]{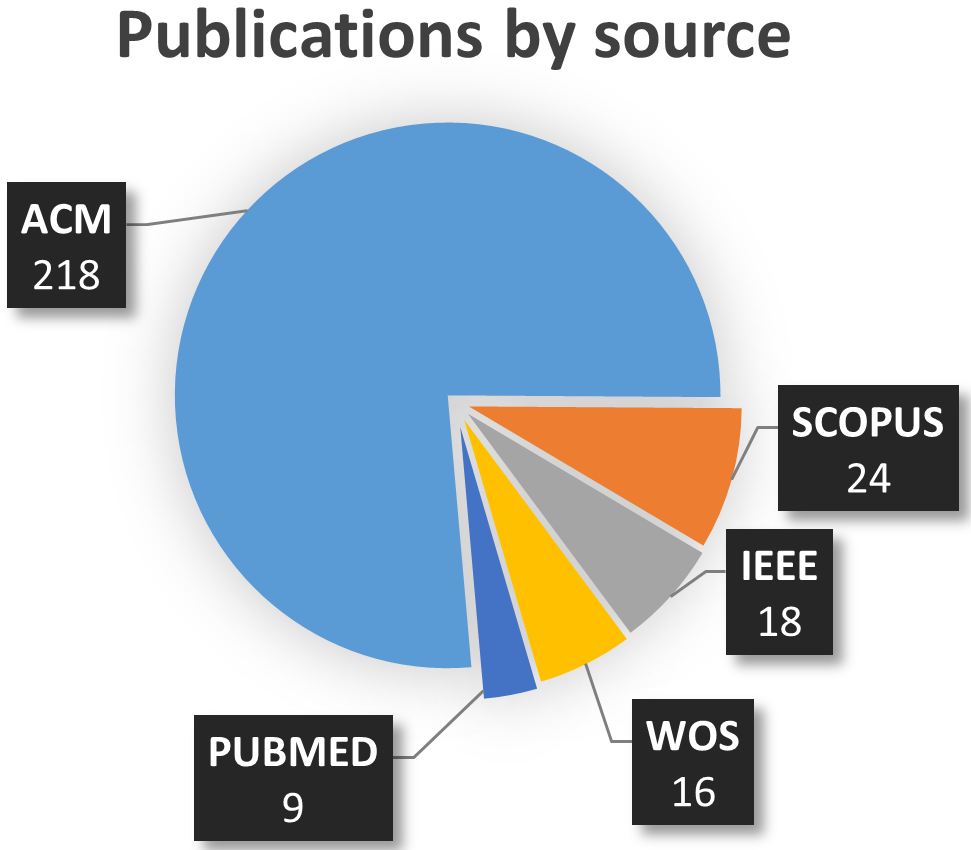}
             \caption{Distribution of publications retrieved across the five databases.}
            \label{fig:sources}
        \end{figure}

        To ensure a comprehensive collection of relevant studies, we utilized five distinct digital databases. SCOPUS and Web of Science (WOS) were selected for their broad interdisciplinary coverage, while the ACM Digital Library, IEEE Xplore Digital Library, and PubMed were chosen for their focus on computer science, technology, and biomedical research, respectively. Figure \ref{fig:sources} illustrates the distribution of studies retrieved from each source. Notably, the ACM Digital Library yielded the highest number of studies among the five databases.
        
    \subsection{Search strategy}

        To ensure a more precise search in the SCOPUS database, we limited our query to the title, abstract, and keywords, as searching the full text yielded a significant number of irrelevant results. In contrast, full-text searches were performed in the other databases. Although the query strings were adjusted according to the specific requirements of each search engine, they consistently included four key concepts: facial expression recognition, computer vision, elderly population, and deep learning. Each concept was paired with multiple synonyms and, where applicable, the "*" wildcard was used to capture various word terminations.

\begin{figure}[h]
             \centering
             \includegraphics[width=\columnwidth]{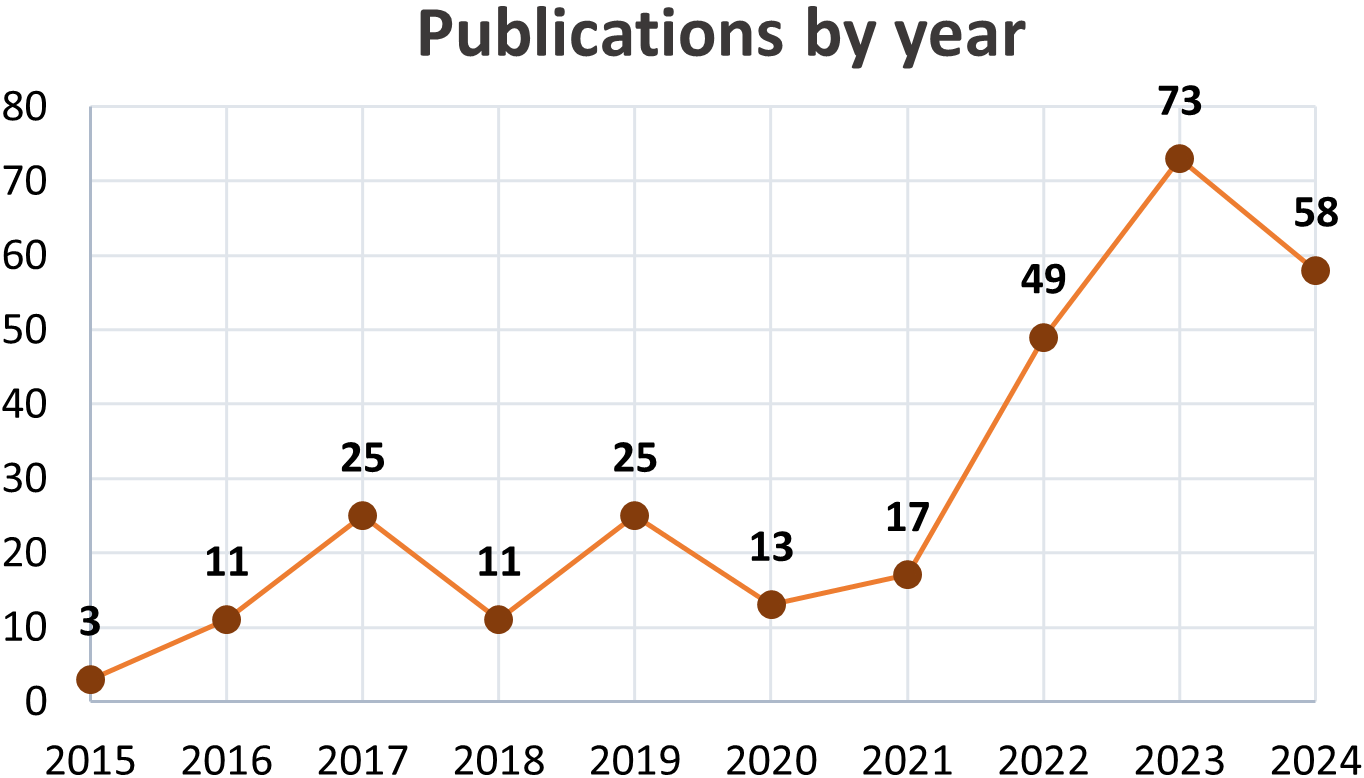}
             \caption{Number of publications found from 2015 to September 2024 (both included).}
            \label{fig:years}
        \end{figure}

        The search was constrained to publications from the past ten years, specifically from 2015 through September 2024, inclusive. Figure \ref{fig:years} presents the number of related studies identified by year, demonstrating a clear upward trend, particularly over the last three years.
        
    \subsection{Quality assessment} \label{section:quality}

        To assess the quality of the included studies, we followed the guidelines proposed by \cite{kitchenham2007guidelines}. The evaluation was based on four primary criteria:
        
        \begin{itemize}
            \item \textbf{Type of publication}. Since conference proceedings may not always undergo a rigorous peer-review process, unlike journal articles, the type of publication was considered as a quality indicator.
            
            \item \textbf{Reproducibility}. This criterion assesses whether the work can be replicated, either through detailed explanations of the methods used, by providing access to the programming code, or by making the collected dataset publicly available.
            
            \item \textbf{Benchmarking with the state-of-the-art}. Whether the proposed methods are compared with prior works under similar conditions, such as using the same performance metrics, dataset splits, and methodologies.
            
            \item \textbf{Performance on public datasets}. The use of publicly available datasets is important, as it allows other researchers to compare their results with those of prior studies in a standardized manner.
        \end{itemize}
        
        By applying these criteria, we were able to filter out studies that exhibited potential bias. This quality assessment was integrated into the study selection process, further detailed in Section \ref{section:selection}, and the list of extracted data from each study, outlined in Section \ref{section:extraction}.

    \subsection{Study selection} \label{section:selection}

        After gathering the search results from multiple databases, totaling 285 studies, we identified and removed 16 duplicate entries. The remaining studies were then screened individually to exclude publications that were not directly relevant to the research topics. Additionally, we eliminated works that did not meet our inclusion criteria, including those not written in English or Spanish, those without full-text access, or those that did not meet the quality standards outlined in Section \ref{section:quality}. 

\begin{figure*}[h]
            \centering
            \includegraphics[width=.8\textwidth]{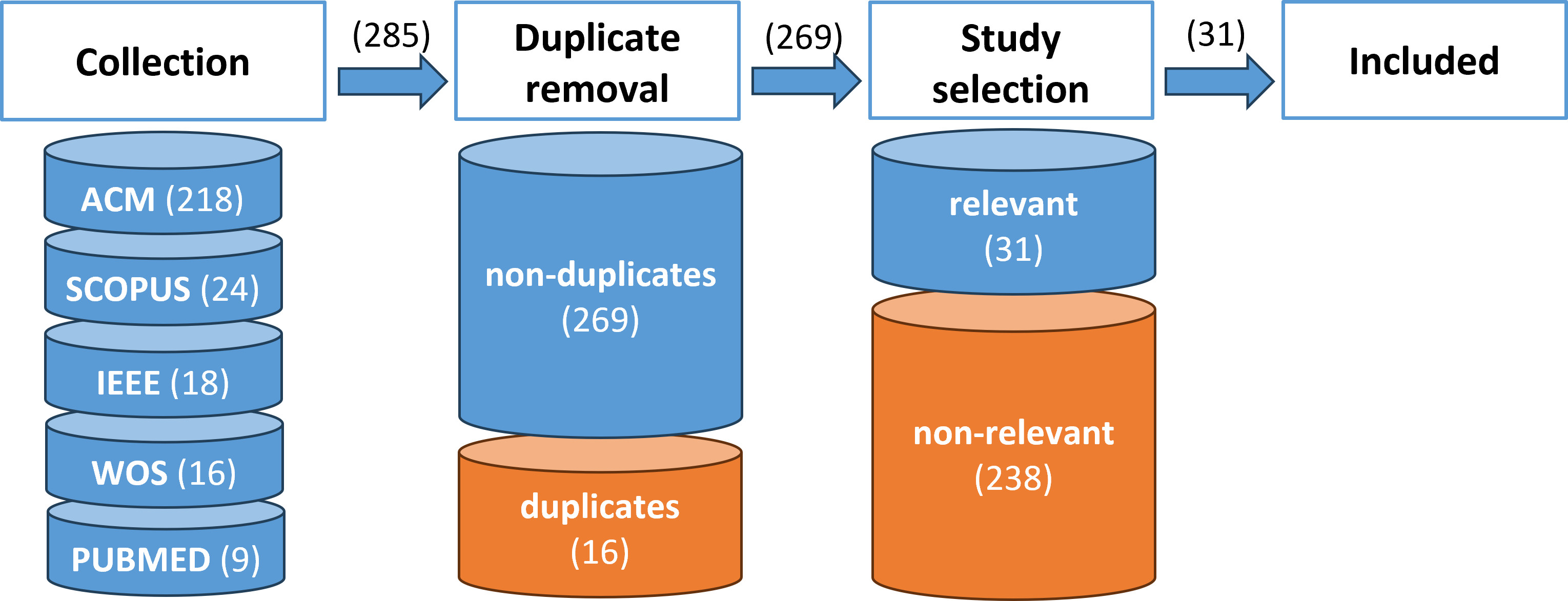}
            \caption{Summary of the systematic review process: collection of publications from the five databases, duplicate removal, and final study selection.}
            \label{fig:selection}
        \end{figure*}

        Figure \ref{fig:selection} illustrates the progression of the selection process, from the initial collection of studies to the final screening. Ultimately, 31 studies were selected for inclusion in this review, of which 15 were journal articles, and 16 were conference proceedings. As stated by \cite{kitchenham2007guidelines}, "publication bias can lead to systematic bias in systematic reviews unless special efforts are made to address this issue". Since scanning conference proceedings is a standard search strategy to address publication bias, they were retained for analysis among the relevant studies.
        
    \subsection{Data extraction and synthesis} 
    \label{section:extraction}

\begin{figure*}[h]
            \centering
            \includegraphics[width=.9\textwidth]{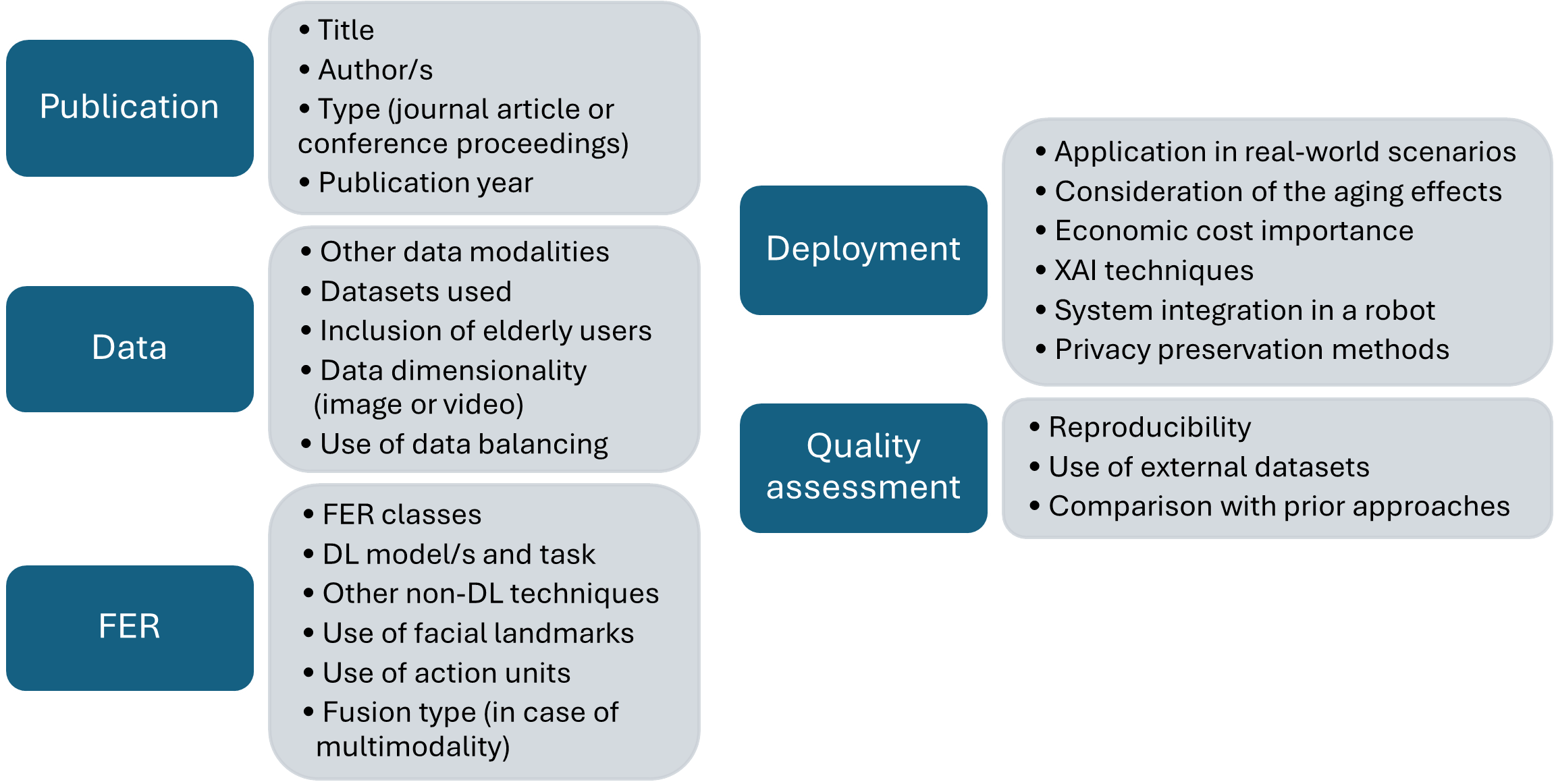}
            \caption{Information extracted from each relevant study.}
            \label{fig:extraction}
        \end{figure*}

        To synthesize the relevant studies, different information was extracted from each one. The gathered fields are shown in Figure \ref{fig:extraction}.
        
\section{Results} 
\label{section:results}
%     % Non-quantitative summaries should be provided to summarize each of the studies and presented in tabular form.
%     % Quantitative summary results should be presented in tables and graphs. 
%     % - Findings
%     %     Description of primary studies. Results of any quantitative summaries. Details of any meta-analysis.
%     % - Sensitivity analysis

    The complete list of relevant studies is provided in Table \ref{tab:studies}. In this section, we present a detailed analysis of these studies to address the research questions posed. For improved readability, the results have been organized into distinct subsections, each addressing a review question.

    \begin{table*}[h]
        \centering
        \begin{adjustbox}{width=\textwidth}
        \begin{tabular*}{1.05\textwidth}{l|cccccccc}
            \toprule
            \multirow{2}{*}{\textbf{Study}} & \multirow{2}{*}{\textbf{Elderly}} & \multirow{2}{3.2em}{\textbf{Aging effects}} & \multirow{2}{*}{\textbf{Video}} & \multirow{2}{4.5em}{\textbf{Economic cost}} & \multirow{2}{*}{\textbf{Deployed}} & \multirow{2}{*}{\textbf{XAI}} & \multirow{2}{3em}{\textbf{Multi-modal}} & \multirow{2}{*}{\textbf{Privacy}} \\
            \\
            \midrule
            \cite{gaya-morey2024ai-powered} &  &  &  & \cmark & \cmark & \cmark &  &  \\
            \cite{huang2024auto} & \cmark &  &  &  &  &  &  & \cmark \\
            \cite{huang2024facial} & \cmark & \cmark &  &  &  &  &  &  \\
            \cite{jayanthi2024enhanced} &  &  &  &  &  &  &  &  \\
            \cite{nadjadecarolis2024exploring} & \cmark &  & \cmark & \cmark & \cmark &  & \cmark & \cmark \\
            \cite{zhu2024defining} & \cmark &  &  &  &  &  &  & \cmark \\
            \cite{anand2023multi-label} & \cmark &  & \cmark &  &  & \cmark & \cmark &  \\
            \cite{chen2023emotion-reading} &  &  &  &  & \cmark &  &  & \cmark \\
            \cite{khajontantichaikun2023facial} & \cmark &  &  &  &  &  &  &  \\
            \cite{petrou2023lightweight} &  &  &  & \cmark &  &  &  &  \\
            \cite{bi2022dynamic} &  &  & \cmark &  &  &  &  &  \\
            \cite{carretopicón2022do} & \cmark &  &  &  &  &  &  &  \\
            \cite{ekosantoso2022facial} &  &  &  &  &  &  &  &  \\
            \cite{fahn2022image} &  &  &  & \cmark & \cmark &  &  &  \\
            \cite{fan2022fer-pcvt} & \cmark &  &  &  &  & \cmark &  &  \\
            \cite{fei2022novel} & \cmark &  &  & \cmark &  &  &  &  \\
            \cite{jiang2022automated} &  &  &  &  &  &  &  &  \\
            \cite{khajontantichaikun2022emotion} & \cmark &  &  &  &  &  &  &  \\
            \cite{sreevidya2022elder} & \cmark &  & \cmark &  &  &  & \cmark &  \\
            \cite{kim2021age} & \cmark & \cmark &  &  &  &  &  &  \\
            \cite{rahatuljannat2021expression} & \cmark & \cmark &  &  &  &  &  &  \\
            \cite{caroppo2020comparison} & \cmark & \cmark &  & \cmark &  &  &  & \cmark \\
            \cite{sharma2020audio-visual} & \cmark &  & \cmark &  &  &  & \cmark & \cmark \\
            \cite{yan2020deep} &  &  &  & \cmark & \cmark &  &  &  \\
            \cite{zhang2020facial} &  &  & \cmark &  &  &  &  &  \\
            \cite{skibelirokkones2019facial} &  &  & \cmark &  &  & \cmark &  &  \\
            \cite{caroppo2018facial} & \cmark &  &  & \cmark &  &  &  &  \\
            \cite{yang2018joint} & \cmark & \cmark &  &  &  &  &  &  \\
            \cite{ziauddin2017facial1} & \cmark &  &  &  &  &  &  & \cmark \\
            \cite{ziauddin2017facial2} &  &  &  &  &  & \cmark &  & \cmark \\
            \cite{wu2015enhanced} & \cmark & \cmark &  &  &  &  &  & \cmark \\
            \bottomrule
            \end{tabular*}
            \end{adjustbox}
        \caption{Relevant studies identified in the systematic review. The columns, from left to right, indicate whether the study includes experiments with elderly users, examines the effects of age on FER, utilizes the temporal dimension, considers economic cost, integrates the solution into a framework or deployment, employs XAI tools, incorporates additional modalities, and implements privacy protection methods.}
        \label{tab:studies}
    \end{table*}

    \subsection{RQ1. DL techniques for FER in elderly populations}
    
        Next, we present the findings related to the first research question, focusing on common DL architectures, datasets, data characteristics, the use of facial landmarks and action units, and related tasks in performing FER for elderly populations.

    \subsubsection{RQ1.1. Deep learning architectures}
    \label{sec:models}
    
        % \begin{itemize}
        %     \item Table with list of architectures.
        %     \item Taxonomy: CNNs, RNNs, Transformers
        %     \item Smaller architectures: MobileNet, EfficientNet, simple CNNs, mini-Xception...
        %     \item Unsupervised: Bayesian Network, AE...
        %     \item Software: Face++, Sighthound, Amazon Rekognition, Microsoft Face, OpenFace 2.0, FaceReader
        %     \item Classes used
        % \end{itemize}

        Table \ref{tab:models} provides an overview of the deep learning-based methods employed in the relevant studies. The table highlights the diversity in model architectures and their applications to various tasks within the facial expression recognition pipeline.

        \begin{table*}
            \centering
            % \begin{adjustbox}{width=\textwidth}
            \begin{tabular*}{\textwidth}{ll|llllll}
                \toprule
                \textbf{Reference} & \textbf{Model/Software} & \textbf{Task/s} & \textbf{Studies} \\
                \midrule
                \cite{VGG} & VGG & FER, feat. ext. & 6 \\
                \cite{GoogLeNet} & Inception & FER, feat. ext. & 5 \\
                \cite{lecun1999object} & CNN & FER, feat. ext. & 5 \\
                \cite{EfficientNet} & EfficientNet & FER, feat. ext. & 3 \\
                \cite{Xception} & Xception & FER & 3 \\
                \cite{MobileNet} & MobileNet & FER, FR, feat. ext.,   landmarks & 3 \\
                \cite{ren2017faster} & Faster R-CNN & FER, OD & 3 \\
                \cite{ResNet} & ResNet & FER, FR, feat. ext. & 3 \\
                \cite{YOLO} & YOLO & FER, OD & 3 \\
                \cite{zhang2016joint} & MTCNN & OD, landmarks & 3 \\
                \cite{noldus} & FaceReader & FER, landmarks, AUs & 2 \\
                \cite{liu2016ssd} & SSD & FER, OD & 2 \\
                \cite{GAN} & GAN & dataset gen., privacy prot. & 2 \\
                \cite{AlexNet} & AlexNet & FER & 2 \\
                \cite{hinton2006fast} & DBN & FER & 2 \\
                \cite{LSTM} & LSTM & FER & 2 \\
                \cite{kuprashevich2024mivolo} & MiVOLO & age, gender & 1 \\
                \cite{ruan2024scatnet} & SCaTNet & feat. ext. & 1 \\
                \cite{openai2024gpt-4} & GPT-4 & text emotion recognition, chatbot & 1 \\
                \cite{kabir2023spinalnet} & SpinalNet & FER & 1 \\
                \cite{SilNet2022} & SilNet & FER & 1 \\
                \cite{liu2022video} & Swin-Transformer & FER & 1 \\
                \cite{fan2022fer-pcvt} & FER-PCVT & FER & 1 \\
                \cite{prados-torreblanca2022shape} & SPIGA & landmarks & 1 \\
                \cite{dosovitskiy2021image} & ViT & feat. ext. & 1 \\
                \cite{wang2021pyramid} & PvT & FER & 1 \\
                \cite{wu2021cvt} & CvT & FER & 1 \\
                \cite{guo2021sample} & SCRFD & OD & 1 \\
                \cite{cao2021openpose} & OpenPose & pose estimation & 1 \\
                \cite{qin2020u2-net} & U2-Net & OS & 1 \\
                \cite{hernandez-ortega2019faceqnet} & FaceQnet & facial image quality & 1 \\
                \cite{baltrusaitis2018openface} & OpenFace 2.0 & landmarks, AUs & 1 \\
                \cite{MicrosoftFace} & Microsoft Face & FER & 1 \\
                \cite{Transformer} & Transformer & FER & 1 \\
                \cite{AmazonRekognition} & Amazon Rekognition & FER & 1 \\
                \cite{niandola2016squeezenet} & SqueezeNet & OD & 1 \\
                \cite{Sighthound} & Sighthound & FER & 1 \\
                \cite{WeiNet2015} & WeiNet & FER & 1 \\
                \cite{Siamese_CNNs} & Siamese NN & FER & 1 \\
                \cite{Song2014} & SongNet & FER & 1 \\
                \cite{GRU} & GRU & FER & 1 \\
                \cite{Face++} & Face++ & FER & 1 \\
                \cite{vincent2010stacked} & SDAE & feat. ext. & 1 \\
                \cite{pearl1988probabilistic} & Bayesian Network & FER & 1 \\
                \cite{jhopfield1982neural} & RNN & FER & 1 \\
                \bottomrule
                \end{tabular*}
                % \end{adjustbox}
            \caption{Deep learning methods and software, along with the tasks they were applied to and the number of studies in which they were utilized. The following acronyms and abbreviations where used for conciseness: facial expression recognition (FER), feature extraction (feat. ext.), object detection (OD), facial recognition (FR), object segmentation (OS).}
            \label{tab:models}
        \end{table*}

        The methods listed in the table were used for a wide range of tasks. The most common was FER itself, where models included a final classification layer to predict facial expressions. Convolutional models such as VGG, Inception, and ResNet were often employed to extract complex features, which were later used for classification tasks \citep{zhang2020facial,bi2022dynamic,huang2024facial}. Object detectors, including Faster R-CNN, YOLO, SSD, MTCNN, and SCRFD, were used in pre-processing steps to localize and crop faces before performing FER \citep{jiang2022automated,gaya-morey2024ai-powered}. Notably, Faster R-CNN, YOLO, and SSD were also used by Khajontantichaikun et al. \citeyearpar{khajontantichaikun2022emotion,khajontantichaikun2023facial} to simultaneously perform object detection and FER. In addition to FER, some studies focused on auxiliary tasks such as facial landmark estimation, for which MobileNet, SPIGA, MTCNN, FaceReader, and OpenFace 2.0 were employed \citep{zhu2024defining,chen2023emotion-reading,gaya-morey2024ai-powered}. Among these, FaceReader and OpenFace 2.0 also supported action unit estimation. Other applications included emotion recognition from text, as demonstrated by \cite{nadjadecarolis2024exploring}, who used GPT-4 for emotion classification and text generation in a social robot. \cite{gaya-morey2024ai-powered} also explored the integration of multiple tasks in a social robot, among which U2-Net and MiVOLO were used for object segmentation and for age and gender estimation, respectively. To address privacy concerns, \cite{huang2024auto} employed GAN to transform images and FaceQnet to estimate image quality. GAN were also used by \cite{zhu2024defining} to generate a FER dataset comprising multiple age groups.

        Three primary families of architectures were frequently employed: CNNs, recurrent neural networks (RNNs), and Transformers. Convolutional neural networks were the most widely used, appearing in 27 models across 22 of the reviewed studies. These feed-forward architectures are highly effective for computer vision tasks and were often used as either end-to-end FER solutions \citep{petrou2023lightweight,gaya-morey2024ai-powered} or feature extractors \citep{caroppo2020comparison,sreevidya2022elder}. On the other hand, RNNs such as GRU and LSTM were used in only four studies, where they excelled in handling temporal information for video-based FER \citep{zhang2020facial} or sequential non-visual data \citep{sharma2020audio-visual}. Transformers, known for their attention mechanisms, have gained traction in recent years. Nine models from the table incorporated attention-based methods, appearing in six studies. Transformers were used for tasks ranging from FER in images \citep{ekosantoso2022facial,huang2024auto} to multimodal analysis \citep{fan2022fer-pcvt} and hybrid convolutional-transformer models like the Convolutional Vision Transformer (CvT).

        A few studies incorporated unsupervised learning phases in their FER pipelines. Uddin et al. \citeyearpar{ziauddin2017facial2,ziauddin2017facial1} used deep belief networks (DBNs) with an initial unsupervised training phase for feature extraction learning, followed by supervised fine-tuning. Similarly, \cite{caroppo2018facial} utilized a Stacked Denoising Auto-Encoder (SDAE), which first learned to reconstruct input data in an unsupervised phase before being fine-tuned for FER.

        To enable deployment on low-resource devices or ensure real-time processing, some studies adopted lightweight architectures. MobileNet \citep{fei2022novel,gaya-morey2024ai-powered} was designed specifically for mobile applications, while EfficientNet \citep{nadjadecarolis2024exploring,gaya-morey2024ai-powered,ekosantoso2022facial} offered scalable performance with optimized resource usage. Similarly, the SqueezeNet \cite{niandola2016squeezenet} achieved the same accuracy than AlexNet with 50 times less parameters, thanks to their proposed \textit{Fire modules}. The mini-Xception network \citep{yan2020deep,petrou2023lightweight}, a compact version of  Xception, drastically reduced the number of trainable parameters, making it ideal for lightweight FER solutions.

        Commercial software also played a role in some studies. \cite{kim2021age} investigated the age bias in FER tools like Amazon Rekognition, Face++, Microsoft Face, and Sighthound, revealing significant performance drops when analyzing elderly faces. OpenFace 2.0 was employed by \cite{sharma2020audio-visual} for facial landmarks and action unit detection, which were subsequently used by DL models for FER. FaceReader, capable of estimating facial landmarks, action units, and expressions, was examined by \cite{zhu2024defining}, who defined expected outputs for smiling images. \cite{chen2023emotion-reading} further integrated FaceReader into an emotion-reading care environment.

\begin{figure}[h]
             \centering
             \includegraphics[width=\columnwidth]{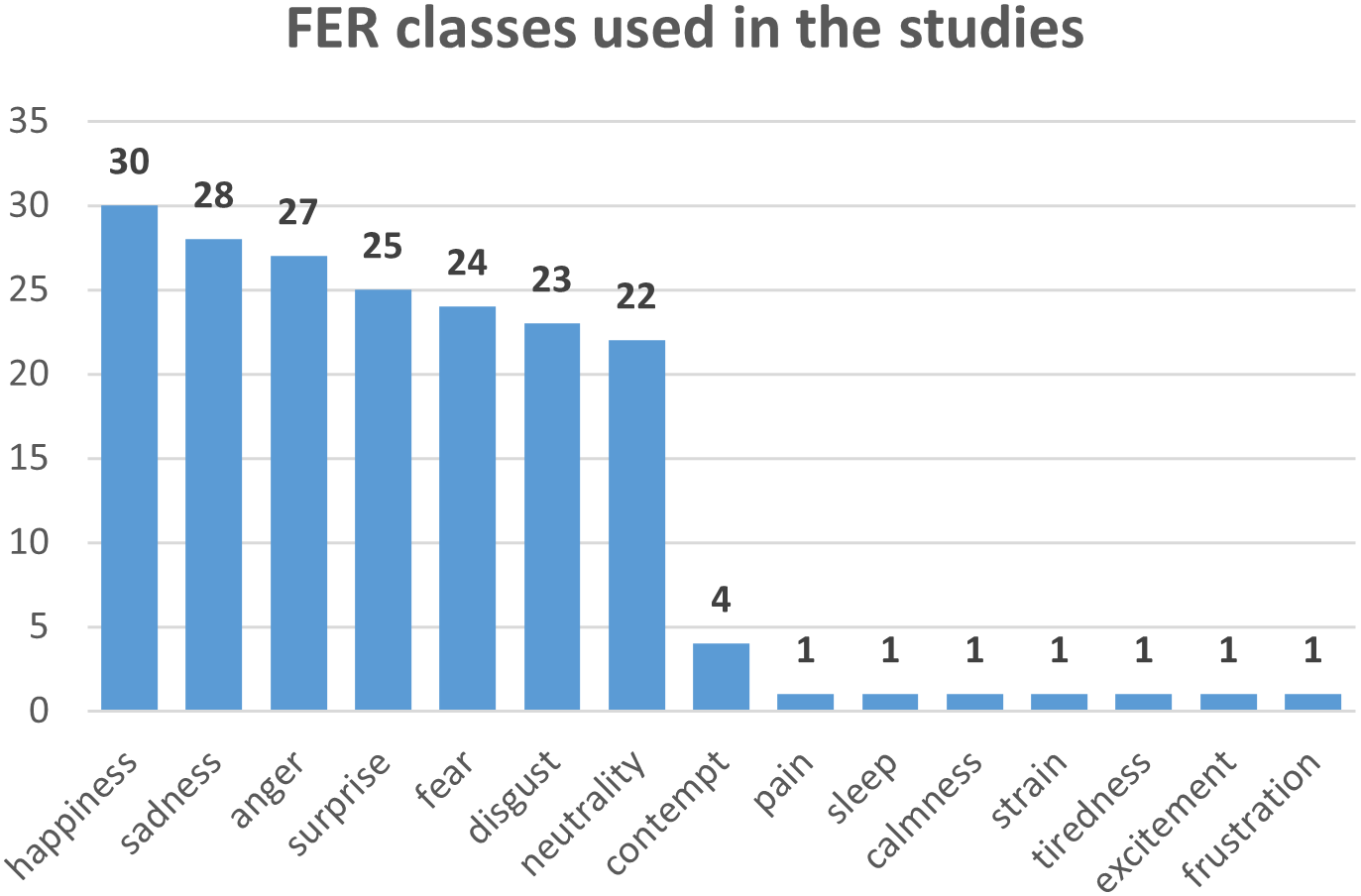}
             \caption{Distribution of FER classes observed in the analyzed studies.}
            \label{fig:classes}
        \end{figure}

        The number and type of output classes in FER models varied across studies. As shown in Figure \ref{fig:classes}, most studies focused on the six basic facial expressions--happiness, sadness, anger, surprise, fear, and disgust--along with the neutral expression. Additional expressions such as contempt \citep{zhang2020facial,skibelirokkones2019facial,fan2022fer-pcvt,jayanthi2024enhanced}, sleep, calmness \citep{chen2023emotion-reading}, strain, tiredness, pain \citep{fan2022fer-pcvt}, excitement, and frustration \citep{anand2023multi-label} were also explored in a smaller number of studies.
        
        Lastly, several studies employed traditional computer vision approaches coupled with deep learning for facial expression recognition. Support vector machines (SVMs) were frequently utilized for facial expression classification after feature extraction through other methods \citep{fei2022novel,caroppo2020comparison}. Principal component analysis was applied in some studies to reduce data dimensionality prior to classification \citep{ziauddin2017facial2,bi2022dynamic}. For face detection and cropping during pre-processing, the Viola-Jones algorithm was employed in multiple works \citep{fei2022novel,gaya-morey2024ai-powered}. Additionally, a variety of feature descriptors were implemented across studies, including local binary patterns \citep{caroppo2020comparison,zhang2020facial,ziauddin2017facial2}, local directional strength patterns \citep{skibelirokkones2019facial}, and the Geneva minimalistic acoustic parameter set \citep{sharma2020audio-visual}.
        
    \subsubsection{RQ1.2. Datasets}
        % \begin{itemize}
        %     \item Table with list of datasets.
        %     \item Automatic vs. controlled
        %     \item Image vs. Video
        %     \item Classes
        %     \item External vs. custom datasets.
        % \end{itemize}
        
        Table \ref{tab:datasets} provides an overview of the datasets used for training in the FER tasks identified in the reviewed studies. As highlighted, these datasets vary significantly in terms of annotations, data type, classes, and sample sizes.

        \begin{table*}
            \centering
            \begin{adjustbox}{width=\textwidth}
            \begin{tabular*}{1.075\textwidth}{ll|llllll}
                \toprule
                \textbf{Reference} & \textbf{Dataset} & \textbf{Users} & \textbf{Ages} & \textbf{Type} & \textbf{Samples} & \textbf{Classes} & \textbf{Studies}\\
                \midrule
                \cite{goodfellow2013challenges} & FER-2013 & N/A & N/A & Image & 35,887 & 7 & 7 \\
                \cite{cebner2010faces-a} & FACES & 171 & 19-80 & Image & 2,052 & 6 & 6 \\
                \cite{lucey2010extended} & CK+ & 123 & 18-50 & Video & 593 & 7 & 5 \\
                \cite{minear2004lifespan} & LifeSpan & 576 & 18-93 & Image & 1,354 & 8 & 4 \\
                \cite{ma2019elderreact} & ElderReact & 46 & N/A & Video & 1,323 & 6 & 3 \\
                \cite{mollahosseini2017affectnet} & AffectNet & N/A & N/A & Image & 440,000 & 7 & 3 \\
                \cite{nojavanasghari2016emoreact} & EmoReact & 63 & 4-14 & Video & 1,102 & 8 & 3 \\
                \cite{zhao2011facial} & Oulu-CASIA & 80 & 23-58 & Video & 480 & 6 & 2 \\
                \cite{langner2010presentation} & RaFD & 49 & N/A & Image & 5,880 & 8 & 2 \\
                \cite{fei2022novel} & CAD & N/A & N/A & Image & 165 & 5 & 1 \\
                \cite{fei2022novel} & CEPD & N/A & N/A & Image & 13,692 & 6 & 1 \\
                \cite{hussain2021human} & Jafar   Hussain & N/A & N/A & Image & 679 & 5 & 1 \\
                \cite{yang2020tsinghua} & TFED & 110 & 19-76 & Image & 1128 & 8 & 1 \\
                \cite{oliver2020uibvfed} & UIBVFED & 20 & 20-80 & Image & 640 & 6 & 1 \\
                \cite{jlyons2020coding} & JAFFE & 10 & N/A & Image & 219 & 7 & 1 \\
                \cite{li2018recursive} & CIFE & N/A & N/A & Image & 14,756 & 7 & 1 \\
                \cite{livingstone2018ryerson} & RAVDESS & 24 & 21-33 & Video & 7,356 & 8 & 1 \\
                \cite{bagherzadeh2018multimodal} & CMU-MOSEI & 1,000 & N/A & Video & 3,228 & 6 & 1 \\
                \cite{zhalehpour2017baum-1} & BAUM-1s & 31 & 19-65 & Video & 1,184 & 8 & 1 \\
                \cite{pantic2005web-based} & MMI & 75 & 19-62 & Video+Image & 848+740 & 6 & 1 \\
                \cite{lundqvist1998karolinska} & KDEF & 70 & 20-30 & Image & 4,900 & 7 & 1 \\
                \bottomrule
                \end{tabular*}
                \end{adjustbox}
            \caption{FER datasets used in the studies included in this review. The columns, from left to right, display the number of users, their age range, whether the dataset contains image or video data, the total number of samples, the number of facial expression classes, and the number of relevant studies that used each dataset.}
            \label{tab:datasets}
        \end{table*}

        Several datasets were automatically collected from the Internet, including FER-2013, ElderReact, AffectNet, EmoReact, Jafar Hussain, CIFE, and CMU-MOSEI. This collection approach enables the creation of large datasets, such as FER-2013 (35,887 samples) and AffectNet (440,000 samples). However, these datasets often suffer from issues such as poor expression annotations \citep{mejia-escobar2023towards}, the presence of duplicate samples \citep{ijimai}, and the absence of demographic information about users. Conversely, datasets collected under controlled conditions offer advantages such as user selection based on demographic criteria, multiple samples from the same individual, and controlled lighting conditions. For example, FACES and LifeSpan emphasize age diversity, supporting cross-age group studies, while ElderReact and EmoReact target elderly individuals and children, respectively. Additionally, UIBVFED stands out as the only synthetically generated dataset, allowing for precise control over the facial expressions of avatars, as well as their demographic characteristics, including age, gender, and ethnicity.

        Although the majority of datasets are in image format, several are video-based \citep{lucey2010extended, ma2019elderreact, nojavanasghari2016emoreact, zhao2011facial, livingstone2018ryerson, bagherzadeh2018multimodal, zhalehpour2017baum-1}. Notably, MMI \citep{pantic2005web-based} offers samples in both image and video formats. Video-based datasets enable the incorporation of temporal information into FER tasks and are particularly valuable for systems requiring real-time processing.

        Most datasets include the six basic facial expressions (happiness, sadness, anger, fear, disgust, and surprise). Additionally, the neutral expression is included in twelve datasets, while contempt is present in three. Other, less common expressions are found in only one dataset each, such as boredom \citep{zhalehpour2017baum-1}, content \citep{yang2020tsinghua}, calmness \citep{livingstone2018ryerson}, curiosity, uncertainty, excitement, frustration \citep{nojavanasghari2016emoreact}, annoyance, and grumpiness \citep{minear2004lifespan}.

        Instead of relying on the datasets listed in Table \ref{tab:datasets}, several of the analyzed studies chose to create custom datasets tailored to their specific research needs. These efforts provided unique opportunities to address specialized populations or expressions that were not well-represented in existing datasets. For instance, Uddin et al. \citeyearpar{ziauddin2017facial2,ziauddin2017facial1} developed a dataset containing 40 close-up RGB-D videos for each of the six basic facial expressions. This approach leveraged depth information to improve facial expression recognition while maintaining user privacy. Similarly, Khajontantichaikun et al. \citeyearpar{khajontantichaikun2022emotion,khajontantichaikun2023facial} curated a dataset of 900 images of Thai elderly individuals, with 150 samples per basic expression. These samples were collected from online media and annotated by five human evaluators. \cite{zhu2024defining}, on the other hand, employed a generative adversarial network to synthesize 20 images for each age group: children, young adults, middle-aged adults, and elderly individuals.
        
        Several studies focused on populations with specific medical or psychological conditions. For instance, \cite{huang2024auto} introduced the Parkinson's disease facial expression dataset, which comprises RGB images of the six basic expressions from 95 patients with Parkinson’s disease. \cite{fan2022fer-pcvt} constructed a dataset of 1,302 videos from stroke patients aged 18-85, capturing not only basic expressions (happiness, sadness, surprise, and anger) but also four condition-specific expressions: painful, strained, tired, and neutral. Additionally, this dataset included facial action coding system (FACS) labels, making it valuable for detailed facial action unit analysis. \cite{sharma2020audio-visual} gathered 140 videos from 70 elderly participants aged 65 and above, 28 of whom were identified as apathetic, enabling the study of apathy recognition in older adults.

        To address the issue of imbalanced data, several techniques are available, such as rebalancing through undersampling or oversampling and employing weighted loss functions. However, despite many studies using imbalanced datasets, such as AffectNet and LifeSpan, the majority either did not apply these methods or did not mention their use. Among the studies that explicitly addressed the imbalance, four utilized undersampling techniques \citep{sreevidya2022elder,wu2015enhanced,huang2024facial,fahn2022image}, while three opted to remove minority classes altogether \citep{yang2018joint,petrou2023lightweight,anand2023multi-label}.
    
    \subsubsection{RQ1.3. Data characteristics}        

        Seven studies leveraged the temporal dimension of visual data to perform facial expression recognition, while the remaining studies were limited to using static images. Most video-based solutions adopted a common approach: extracting features from individual frames and employing an RNN to capture temporal information. For instance, \cite{zhang2020facial} utilized the VGG16 network for feature extraction, followed by a long short-term memory (LSTM) network to process sequential data. Similarly, \cite{skibelirokkones2019facial} employed local directional strength patterns as frame descriptors, which were then fed into an RNN. \cite{bi2022dynamic} adopted a comparable methodology, using GoogleNet for feature extraction, applying principal component analysis to reduce data dimensionality, and processing the resulting data with a Bayesian probabilistic framework. In the same line, \cite{sreevidya2022elder} extracted features from multiple video frames using a convolutional neural network and utilizing an LSTM to handle the temporal aspects. 
        
        Other studies explored variations of this approach to optimize performance. For example, \cite{sharma2020audio-visual} combined visual features extracted by VGG16 with action units and facial landmarks obtained via OpenPose 2.0, processing the sequential data with a gated recurrent unit (GRU). While most studies relied on RNNs to model sequential information, \cite{anand2023multi-label} took a different route by employing a multimodal Transformer architecture \citep{Transformer} that integrated visual data, audio, and text. \cite{nadjadecarolis2024exploring}, despite performing FER on still images, incorporated temporal information by calculating the frequency of different expressions over a defined time period. They determined the final expression by selecting the most frequently occurring one, giving greater weight to the most recent frames.

        Only four studies opted for multimodal solutions, integrating non-visual data into the facial expression recognition pipeline. These approaches demonstrated the potential of combining modalities to enhance predictive performance. 
        
        \cite{sharma2020audio-visual} investigated apathy detection from a multimodal perspective. For the audio modality, they extracted features using the Geneva minimalistic acoustic parameter set, which includes 18 low-level descriptors based on frequency, energy, and spectral parameters. These features were processed through fully connected layers to learn meaningful representations. The audio features were then concatenated with visual features, action units, and facial landmarks before being passed through additional fully connected layers. Comparing their multimodal model with uni-modal approaches, they observed significant improvements in prediction accuracy when leveraging multiple modalities.
        
        \cite{sreevidya2022elder} proposed two distinct methods for processing audio signals. The first approach used a one-dimensional representation, extracting features such as prosody, spectral coefficients, and voice quality characteristics (e.g., tenseness and creakiness) and feeding them into a 1-D convolutional neural network. The second approach converted the audio signals into 2-D spectrograms, which were processed using the Inception-V2 \citep{szegedy2016rethinking} network. For the visual modality, a separate CNN was employed to compute FER. The final multimodal model fused intermediate layers from both modalities, with the fusion layers determined through a grid-based search. Results showed that combining features from both audio and visual data significantly improved overall performance.
        
        \cite{anand2023multi-label} introduced text as a third modality in their FER pipeline, proposing a multimodal Transformer network. The architecture consisted of modality-specific peer networks that independently learned discriminative features from each modality. These features were then fused by minimizing the Kullback-Leibler divergence between peer networks. This design enabled the network to learn mode-specific patterns while effectively leveraging multiple modalities simultaneously, leading to remarkable improvements in performance, particularly in cross-dataset settings.
        
        \cite{nadjadecarolis2024exploring} focused on integrating vision and text modalities. For the visual modality, they trained an EfficientNet model \citep{tan2020efficientnet} to process RGB images. For the text modality, they utilized GPT-4.0 \citep{openai2024gpt-4} to predict emotions from textual inputs. A late-fusion strategy was employed for the final prediction, using the confidence scores from both modalities. 
        
    \subsubsection{RQ1.4. Facial landmarks and action units for FER}
    \label{section:landmarks}
        % \begin{itemize}
        %     \item Only three use them for FER.
        %     \item List studies using landmarks and AUs and how.
        % \end{itemize}
        
        Facial landmarks and action units (AUs) are two fundamental tools frequently employed for facial expression recognition. Facial landmarks are visually identifiable points on the face, such as specific regions around the eyes, nose, and mouth, which help characterize facial geometry. AUs, on the other hand, were introduced in the facial action coding system \citep{ekman1978facial} to describe 46 basic muscle movements, providing a more granular understanding of facial expressions. These features serve as inputs for various computational approaches to facial expression recognition.  
        
        One notable application of landmarks and AUs is in the multimodal apathy detection solution by \cite{sharma2020audio-visual}. They integrated landmarks and AUs as features into a multimodal model alongside visual and audio data. They computed these features using the OpenFace 2.0 toolkit \citep{baltrusaitis2018openface}, and demonstrated that combining multiple modalities significantly enhanced system performance. In contrast, \cite{caroppo2020comparison} extracted geometrical features--linear, elliptical, and polygonal--from facial landmarks estimated using an improved version of active shape models \citep{milborrow2014active}. These features were then used to train various machine learning models. However, their results showed that the performance of geometrical landmark-based features lagged behind that of DL-based features, such as those computed by VGG16 \citep{simonyan2015very}. A different approach was adopted by \cite{wu2015enhanced}, who utilized manually annotated fiducial points provided by \cite{guo2013facial} to train a Bayesian network. By incorporating age labels during the training phase, Wu et al. demonstrated significant performance improvements, highlighting the potential of integrating demographic features like age into the FER task.
        
        On the other hand, some studies computed facial landmarks and AUs but did not use them directly for facial expression recognition. For instance, \cite{gaya-morey2024ai-powered}, \cite{zhu2024defining}, and \cite{chen2023emotion-reading} incorporated these features for other purposes, such as system development and integration, but refrained from leveraging them for FER tasks specifically.

    \subsubsection{RQ1.5. Other tasks}
        % \begin{itemize}
        %     \item List tasks that are computed alongside FER in the studies.
        % \end{itemize}
        
        Apart from the facial expression recognition task, which is the central focus of this review, several other tasks were computed alongside it in multiple studies.  
        
        Face detection emerged as the most common preprocessing step in the studies reviewed. This step involves cropping the image to focus on the face and removing irrelevant data for the FER task. Several approaches were used for face detection, with the Viola-Jones technique \citep{viola2001rapid} being the most prevalent due to its simplicity, high speed, and low computational requirements. Another common preprocessing task was facial landmark estimation, which is further addressed in Section \ref{section:landmarks}.  
         
        Age estimation was frequently employed to address facial expression differences among age groups. For example, \cite{caroppo2020comparison} automatically labeled images from the CIFE and FER-2013 datasets using landmark-based methods \citep{wu2012age} to study FER accuracy across different age groups. Similarly, \cite{zhu2024defining} utilized FaceReader \citep{noldus} to classify images into age groups and analyze performance variations. \cite{huang2024facial} not only used age estimation with ResNet18 \citep{he2016deep} for evaluation purposes but also integrated it into their FER system to compute age group-specific features. In contrast, \cite{yang2018joint} adopted a multi-tasking approach, simultaneously predicting both age and facial expressions to improve accuracy for both tasks. Because age estimation is included as a means to studying aging effects on facial expression recognition, these works are analyzed in more depth in Section \ref{section:aging-effect}
        
        FER was also employed as a tool for detecting cognitive impairment (CI), which is often characterized by abnormal emotional patterns. \cite{fei2022novel} used the occurrence of facial expressions in consecutive frames as input to an SVM, enabling CI detection in elderly individuals. \cite{jiang2022automated} adopted a similar approach, utilizing the VGG19 network \citep{simonyan2015very} to predict facial expressions. These predictions were then fed into a linear regression model and an SVM for CI detection. They found that CI participants displayed significantly fewer positive emotions, more negative emotions, and higher facial expressiveness. However, attempts to identify CI subtypes using the same features were unsuccessful. Apathy detection, closely related to FER, was also explored by \cite{sharma2020audio-visual}, who proposed a multimodal solution that combined RGB data, action units, and audio. Their results demonstrated that using multiple modalities improved performance compared to relying on a single modality.  
        
        Several studies integrated FER with additional tasks to create multifunctional systems. \cite{yan2020deep} combined FER with facial recognition and fall detection. For facial recognition, they used MobileNetV2 \citep{sandler2018mobilenetv2}, while fall detection was achieved by estimating body pose using OpenPose \citep{cao2021openpose} and analyzing shoulder and chest coordinates. This integration aimed to detect and respond to different warning situations, like falls and negative expressions prolonged in time. Similarly, \cite{fahn2022image} performed fall detection using body pose estimation and incorporated gaze, facial expression, and speech recognition into a social robot. \cite{nadjadecarolis2024exploring} extended FER by including emotion recognition from text and dialogue generation using GPT-4.0. These tasks, paired with FER, endowed a social robot with empathy capabilities. A user study involving 30 elderly participants revealed that the robot's empathic features improved usability and provided more positive user experiences. \cite{gaya-morey2024ai-powered} took this further by including eight tasks in their system: FER, face and person detection, age and gender estimation, facial recognition, facial landmark estimation, and background subtraction. They also provided multiple options for each task to accommodate different use cases. Their objective was to create a versatile, AI-powered module for any socially interactive agent.
    
    \subsection{RQ2. Successful deployment in real-world environments}

        We now present the findings related to the second research question, emphasizing framework integration, deployment, the impact of age on FER, privacy considerations, economic costs, and the use of explainable AI techniques.

    \subsubsection{RQ2.1. Deployment}

        % \begin{itemize}
        %     \item List studies deploying the system in a real-case scenario or implementing a framework.
        % \end{itemize}
    
        The deployment of FER systems in real-world scenarios, such as care systems, presents multiple challenges, including privacy concerns and the need for seamless integration with other tools. Several studies have addressed these issues, offering innovative approaches to improve the functionality and user experience of FER systems.
    
        \cite{chen2023emotion-reading} proposed an adaptive nursing care environment that adjusts environmental factors--such as music, lighting, temperature, and scent--based on the detected emotions of users. This system aims to transform negative emotions into positive ones, for example, guiding anger and fear toward calmness, and disgust and sadness toward happiness. By personalizing environmental settings, they highlighted how FER systems can actively enhance users' well-being and emotional state.
    
        Building on the theme of user interaction, \cite{nadjadecarolis2024exploring} focused on the role of empathy in human-robot interactions for elderly care. Their system recognized emotions using both visual data and text and enabled the robot to express emotions through speech and facial expressions displayed on a screen. GPT 4.0 \citep{openai2024gpt-4} was employed for emotional state analysis and dialogue generation. A user study involving 30 elderly participants found that the empathic version of the robot was more user-friendly and provided a better overall experience compared to a non-empathic version.
        
        Similarly, \cite{yan2020deep} developed a multi-functional system for clinical monitoring, combining facial expression, gesture, and face recognition to aid in patient treatment. This system can alert caregivers of critical situations, such as falls, the presence of unknown individuals, or prolonged negative emotions (e.g., sadness, anger, fear, or disgust).
        
        \cite{gaya-morey2024ai-powered} expanded the scope of FER applications by integrating multiple computer vision tasks into a single module designed for socially interactive agents. Their system supports a wide range of tasks, including facial expression recognition, facial recognition, facial landmark estimation, face and person detection, age and gender estimation, and background subtraction. Additionally, they incorporated explainable AI tools to enhance transparency and user trust.
        
        Lastly, \cite{fahn2022image} explored the integration of FER with complementary modalities, including gaze, speech, and pose recognition, within a social robot. While their work provided detailed descriptions of each task, it lacked specifics regarding their integration into the robot and did not include a user study.
    
    \subsubsection{RQ2.2. Effects of age on facial expression recognition}
    \label{section:aging-effect}
        % \begin{itemize}
        %     \item List studies directly facing (mentioning) the aging effect.
        %     \item List studies including elderly users in the training. Relate with dataset list.
        % \end{itemize}
        
        The age of users' faces significantly influences both human and automated methods for decoding facial expressions, as highlighted in numerous studies \citep{fölster2014facial, ko2021changes, battinisonmez2019computational}. Consequently, various works have introduced specific attention mechanisms to address the challenges posed by age differences in facial expression recognition.
        
        One of the earliest efforts to incorporate age into FER models was undertaken by \cite{wu2015enhanced}, who proposed a three-node Bayesian network that integrated age information during training. Utilizing the FACES and LifeSpan datasets, which include both expression and age labels, they evaluated the model's performance through within-group and between-group assessments. Their findings revealed a substantial bias in all between-group cases, underscoring the importance of age-specific modeling. Notably, incorporating age information during training improved recognition accuracy across most expressions, except for happiness.  
        
        Building on this concept, \cite{yang2018joint} adopted a multi-task learning framework to simultaneously predict age and expressions. Their approach, which leveraged two feature-extracting sub-networks (a convolutional neural network and a scatter network), allowed for shared feature representations across tasks. The model demonstrated superior performance on the FACES and LifeSpan datasets, outperforming earlier single-task solutions in both age and expression predictions.  
        
        Another innovative approach came from \cite{huang2024facial}, who integrated an age group classifier directly into the recognition pipeline. This allowed their model to specialize in age-specific features for distinct groups: babies (0–3 years), adolescents (4–19), young adults (20–39), middle-aged adults (40–69), and elderly adults (70+). Tested on the RAF-DB, AffectNet, and FACES datasets, their method achieved significant performance gains, particularly in recognizing facial expressions in the elderly population.  
        
        Some works have focused on evaluating and addressing age-related biases in commercial FER systems. For example, \cite{kim2021age} benchmarked four commercial systems using the FACES dataset across three age groups: young, middle-aged, and older adults. They found that these systems consistently performed best on younger individuals and worst on older users. Moreover, specific expression classes, such as anger and neutrality, exhibited significantly lower positive predictive values for older individuals, highlighting critical shortcomings in current FER technologies. \cite{zhu2024defining} also explored the performance of a commercial system, i.e., FaceReader \citep{noldus}. They used a GAN to generate 80 images distributed across four age groups: children, young adults, middle-aged adults, and elderly individuals. To validate the system, they conducted a survey with 496 participants via Amazon Mechanical Turk. While the system achieved strong agreement with survey responses for young, middle-aged, and elderly images (Cohen's kappa $>0.8$), it performed poorly for children's images (kappa = 0.4), suggesting unique challenges in recognizing expressions in younger faces.  
        
        A more targeted exploration of age-specific datasets was conducted by \cite{rahatuljannat2021expression}, who focused on children and elderly populations using the EmoReact and ElderReact datasets. By employing a siamese neural network trained on pairs of images (positive pairs with the same expression and negative pairs with different expressions), they achieved strong within-dataset performance. However, cross-dataset testing revealed interesting dynamics: training on ElderReact and testing on EmoReact led to decreased performance, while training on EmoReact and testing on ElderReact resulted in an unexpected improvement.  
        
        Finally, \cite{caroppo2020comparison} offered a broader perspective by studying traditional machine learning and deep learning methods across four datasets: FACES, LifeSpan, FER-2013, and CIFE. They analyzed performance across four age groups: young (18–29 years), middle-aged (30–49 years), old (50–69 years), and very old (70–93 years). Since FER-2013 and CIFE lacked age labels, they computed them automatically through landmark-based methods \citep{wu2012age} to conduct their evaluations. Among the models tested, a combination of the VGG16 deep architecture with random forest yielded the best results, particularly for aging adults. However, cross-dataset evaluations revealed notable performance drops, especially for younger individuals, with per-class results varying significantly by dataset. 
    
    % \subsubsection{Data capture devices}
        % \begin{itemize}
        %     \item It was not specified in any study. Maybe this question can be removed. 
        % \end{itemize}
    
    \subsubsection{RQ2.3. User privacy protection}
        % \begin{itemize}
        %     \item Figure showing proportion of studies with and without privacy concerns.
        %     \item Describe the three main strategies: depth-only, RGB with privacy protection, and RGB with landmarks.
        %     \item List and explain all studies in the text.
        % \end{itemize}
        
        Privacy preservation is a critical concern in multiple fields, especially in healthcare applications, and particularly when visual data is involved. Multiple reviewed studies addressed this issue through diverse strategies aimed at protecting user identities while maintaining the effectiveness of facial expression recognition systems.
        
        \cite{huang2024auto} tackled privacy concerns by employing a Star Generative Adversarial Network (StarGAN) to generate synthetic facial expression images of Parkinson’s disease patients. This technique preserved the patients’ identities while retaining essential expression data, ensuring privacy without compromising system performance. Another effective strategy involved the use of depth images instead of RGB data. Uddin et al. \citeyearpar{ziauddin2017facial2,ziauddin2017facial1} demonstrated the feasibility of FER with depth images, which inherently obfuscate user identities. They employed various feature descriptors, such as local directional position patterns and modified local directional patterns, to achieve robust FER performance.
        
        Several studies leveraged facial landmarks or action units to enhance FER while reducing reliance on raw RGB data. For instance, \cite{sharma2020audio-visual}, \cite{caroppo2020comparison}, and \cite{wu2015enhanced} utilized these intermediate features, which allowed anonymization of the data after the initial RGB image processing stage. This approach offers a balance between accuracy and privacy by removing identifiable information in later stages of processing.
        
        \cite{nadjadecarolis2024exploring} adopted a different approach, embedding their FER system into a small robot. Their privacy-preservation efforts focused on local processing, ensuring that all computations occurred within the robot itself, thereby eliminating the need for transmitting user data over the Internet.
    
    \subsubsection{RQ2.4. Economic cost}
        % \begin{itemize}
        %     \item Say how many studies included economic concerns, tried to lower hardware requirements, etc.
        % \end{itemize}
    
        The economic cost is a common concern when deploying a system, and it can be addressed in various ways, such as reducing hardware requirements or utilizing low-cost equipment. We identified eight studies that mentioned economic cost in some capacity.
        
        \cite{nadjadecarolis2024exploring} conducted their experiment using the Ubtech Alpha Mini robot, chosen for its affordability. To further reduce costs, they selected the EfficientNet model \citep{tan2020efficientnet} for the FER task, which helped lower hardware requirements while increasing inference speed. Similarly, \cite{petrou2023lightweight} used the mini-Xception model for FER, notable for requiring only 58,000 parameters and having a size smaller than 1 MB, making it ideal for low-specification environments. \cite{yan2020deep} also leveraged cost-effective models, utilizing mini-Xception for FER, SqueezeNet \citep{niandola2016squeezenet} for face detection, and MobileNetV2 \citep{sandler2018mobilenetv2} for face recognition. These models were deployed on a Raspberry Pi with a Pi Camera, both highly affordable components. \cite{fei2022novel} and \cite{fahn2022image} adopted similar strategies to reduce hardware needs by employing compact models. Fei et al. utilized MobileNetV2, while Fahn et al. developed a small custom CNN for FER, showcasing how smaller models can effectively address economic constraints.
        
        \cite{gaya-morey2024ai-powered} approached economic cost differently, offering a framework with multiple options for each task, accommodating a wide range of hardware requirements. This flexibility allows their framework to adapt to various environments and budgets. Additionally, their client-server scheme significantly reduces the hardware demands on the robot, as computationally intensive tasks are handled by an external computer.
        
        Lastly, Caroppo et al. \citeyearpar{caroppo2018facial,caroppo2020comparison} highlighted in their future work plans to integrate their system into a cost-effective application, further emphasizing the importance of affordability in FER system design.
    
    \subsubsection{RQ2.5. Explainable AI techniques}
        % \begin{itemize}
        %     \item List the XAI methods used: LIME, Grad-CAM, and clustering visualization.
        %     \item Explain the purpose of their utilization: find system faults, increase user trust, etc.
        % \end{itemize}
    
        As noted in the work by \cite{barredoarrieta2020explainable}, the search for explainable AI models is driven by goals such as trustworthiness, causality among data variables, transferability, informativeness for decision-making, confidence, and fairness, among others. Among relevant studies, we identified six that address these goals.
    
        \cite{anand2023multi-label} developed a multimodal solution to FER, integrating text, audio, and visual data, and providing explanations separately for each modality. They employed local interpretable model-agnostic explanations (LIME) \citep{ribeiro2016why} for the text and vision modalities, and its audio counterpart, audioLIME \citep{haunschmid2020audiolime}, for audio. LIME explains the predictions of any classifier by learning an interpretable model locally around a prediction. The explanations consisted of visual plots (video frames and spectrograms) where the most relevant features for correct predictions were highlighted in green. In the visual modality, these features primarily corresponded to the face and arms. \cite{gaya-morey2024ai-powered} also adopted model-agnostic XAI methods for integration into their general-purpose computer vision module designed for socially interactive agents. They incorporated three approaches: LIME (previously mentioned), randomized input sampling for explanation of black-box models (RISE) \citep{petsiuk2018rise}, and Shapley additive explanations (SHAP) \citep{lundberg2017unified}. Being model-agnostic, these methods are versatile and can be applied to any technique used for any task. The explanations they generate share a common format: visual plots that highlight the importance of different image regions using color gradients.
    
        In contrast to model-agnostic methods, \cite{fan2022fer-pcvt} adopted model-specific XAI techniques tailored to the architectures they evaluated. Their study introduced a patch-convolutional vision transformer (FER-PCVT) for FER, which they compared against ResNet18 \citep{he2016deep} and Vision Transformer (ViT) \citep{dosovitskiy2021image}. To explain ResNet18, they used Gradient-weighted Class Activation Mapping (Grad-CAM) \citep{selvaraju2017grad}, a method specifically designed for convolutional neural networks commonly used in computer vision tasks. Grad-CAM uses the gradients flowing into the final convolutional layer to produce a coarse localization map that highlights important regions in the image for a target prediction. For transformer-based models, such as ViT and FER-PCVT, they aggregated attention weights across all layers. All explanations were visualized as heat maps, with hot regions corresponding to the most relevant areas for a specific prediction. FER-PCVT demonstrated the best results, extracting more specific facial features, while ViT attributed importance to data on the periphery of the image for certain classes, and ResNet18 focused on more extensive facial regions.
    
        Some studies also analyzed the clustering ability of the computed features in the last layer before classification. Although different from the aforementioned XAI methods, these approaches provide insights into the models' inner workings by examining how well the models separate classes and display the perceived similarities between them. In this context, \cite{fan2022fer-pcvt} used t-distributed stochastic neighbor embedding to visualize the last layer of the model by mapping each data point to a two-dimensional space. Similarly, Uddin et al. explored various visual descriptors such as local directional strength patterns, local directional position patterns, and modified local directional patterns in several works \citeyearpar{ziauddin2017facial2,ziauddin2017facial1,skibelirokkones2019facial}, employing either local discriminant analysis or generalized discriminant analysis to get 3D plots where classes were clearly grouped into distinct clusters.

\section{Discussion}

    % Strengths, weaknesses, recommendations for future studies...
    In this section, we discuss the major findings of the systematic literature review, and analyze strengths and weaknesses of the found approaches for facial expression recognition on the elderly, as well as their main characteristics, attending to common aspects to take into account for deployment in real scenarios. Additionally, we give recommendations for future works, based on the findings.

    % Architectures, classes
    Convolutional architectures remain the most frequently utilized for vision-based data, as evidenced by the reviewed studies. Popular models such as VGG, Inception, and ResNet have proven to be robust feature extractors \citep{ekosantoso2022facial,bi2022dynamic}. Recurrent architectures, on the other hand, are predominantly employed when temporal information is crucial for recognition tasks \citep{skibelirokkones2019facial,zhang2020facial}. Meanwhile, Transformer architectures are gaining traction for both visual and sequential data \citep{fan2022fer-pcvt,anand2023multi-label}. For resource-constrained environments, lightweight models such as MobileNet, EfficientNet, and mini-Xception, designed for efficient deployment on low-power devices, have emerged as leading solutions \citep{fei2022novel,nadjadecarolis2024exploring,yan2020deep}.

    % Datasets and problems with imbalanced datasets
    Deep learning models thrive on data-driven learning, but they are inherently susceptible to biases in the training datasets. Three widely used datasets--FER-2013, CK+, and AffectNet--lack adequate representation of elderly individuals and do not include age annotations. This is a notable limitation given the demonstrated differences in facial expressions across age groups (see Section \ref{section:aging-effect}). In contrast, datasets such as FACES and LifeSpan provide age annotations, making them valuable for improving prediction accuracy for elderly populations \citep{caroppo2020comparison,wu2015enhanced}. Similarly, ElderReact, while lacking explicit age annotations, contains a large volume of videos featuring elderly individuals. Class imbalance is another significant challenge in FER datasets, exemplified by AffectNet (with over a 100,000-sample difference between the happiness and disgust classes) and LifeSpan (580 samples for the neutral class versus only 7 for disgust). However, many studies lack information on how they addressed this imbalance \citep{jiang2022automated,nadjadecarolis2024exploring,caroppo2018facial,caroppo2020comparison}. When tackled, class imbalance was typically addressed by undersampling \citep{sreevidya2022elder,wu2015enhanced,huang2024facial,fahn2022image} or by removing minority classes entirely \citep{anand2023multi-label,petrou2023lightweight,yang2018joint}. While these methods can help in solving the imbalance problem, they reduce dataset diversity. Alternative approaches, such as oversampling techniques \citep{bach2017study,mohammed2020machine} or class-weighted loss functions \citep{cui2019class-balanced}, are preferable as they maintain dataset size and variability. Additionally, metrics like accuracy should be avoided for evaluating imbalanced data, as they fail to reflect true model performance across classes. Instead, metrics like the F1 score, which balances precision and recall for each class, are more appropriate \citep{branco2016survey}.

    % Data characteristics: image vs. video and multiple modalities
    The literature reveals no consensus on whether static images suffice for FER or whether motion information is essential. For instance, while \cite{alves2013recognition} argued that dynamic expressions elicit stronger facial and physiological responses in observers, Gold et al. \citeyearpar{gold2013efficiency} found no additional benefit in using dynamic data for emotion recognition beyond what a single static snapshot provides. Nonetheless, static images dominate the analyzed studies, likely due to the scarcity of large-scale video datasets. Developing video datasets for FER could not only provide new insights into the importance of dynamic expressions but also allow direct comparisons with static datasets. Similarly, multimodal solutions combining visual data with other modalities, such as audio, have demonstrated greater robustness to errors in a single modality. For instance, studies leveraging the ElderReact dataset showed that combining visual and audio data outperformed single-modality solutions \citep{sreevidya2022elder,anand2023multi-label}.

    % Other tasks and facial landmarks
    Facial expression recognition can be combined with other tasks, as demonstrated in various studies, such as facial recognition \citep{yan2020deep}, fall detection \citep{fahn2022image}, and speech recognition \citep{nadjadecarolis2024exploring}. These combinations can be particularly valuable in healthcare environments, enabling systems to automatically identify users, detect potentially hazardous situations, and provide tools to mitigate loneliness. Integrating these tasks is advantageous when developing smart care systems, as they can operate using the same input data (e.g., RGB video from a camera) and share similar resource requirements. Additionally, tasks such as face detection and age estimation have been primarily utilized as preprocessing steps to enhance the performance of expression recognition systems \citep{huang2024facial,sreevidya2022elder}.

    % Aging effects, deployment
    Given the impact of aging on FER, as discussed in Section \ref{section:aging-effect}, these studies incorporate training with datasets that include elderly individuals. Some go further by improving results through approaches such as incorporating user age into training \citep{huang2024facial,wu2015enhanced} or adopting multi-tasking frameworks \citep{yang2018joint}. However, we believe these approaches primarily address the scarcity of data representing diverse age groups, which often results in underrepresentation of elderly users. This highlights the critical need for more comprehensive datasets tailored to this demographic or the integration of multiple datasets that include elderly people. Moreover, only a few studies were found to integrate the developed methods into real-world applications, such as mobile applications, assisted living devices, and healthcare centers \citep{nadjadecarolis2024exploring,chen2023emotion-reading}. Since the ultimate goal of these studies is societal adoption and practical deployment, we argue that further research is required to understand elder users’ acceptance and trust in these technologies. Additionally, investigations into their potential benefits, risks of failure, and other real-world considerations are essential for ensuring successful implementation.

    % Privacy and economic cost
    As highlighted by \cite{little2009pervasive}, the proliferation of technology raises critical concerns about usability, trust, and privacy for elderly users. Notably, most reviewed studies (22 in total) relied on RGB data with identifiable user information, yet very few incorporated privacy-preserving measures. Techniques such as anonymization through facial landmark estimation or action units \citep{sharma2020audio-visual} and privacy-preserving transformations \citep{huang2024auto} should be more widely adopted. Local processing is another recommended strategy to reduce privacy risks associated with external data exchange \citep{nadjadecarolis2024exploring}. Economic considerations were also prevalent, with many studies focusing on reducing hardware requirements \citep{yan2020deep,fahn2022image,nadjadecarolis2024exploring}. Combining privacy-preserving techniques with lightweight, local solutions could simultaneously address privacy and cost concerns, making the systems more accessible.

    % XAI
    A notable gap in the reviewed studies is the limited use of explainable AI techniques. XAI is crucial for justifying model predictions and fostering user trust, particularly with complex and opaque models \citep{barredoarrieta2020explainable}. While some studies analyzed the clustering ability of features \citep{skibelirokkones2019facial,ziauddin2017facial2}, this approach only measures classification effectiveness and neglects input feature contributions. Advanced XAI methods, such as LIME \citep{ribeiro2016why} or Grad-CAM \citep{selvaraju2017grad}, were employed in a few studies \citep{fan2022fer-pcvt,gaya-morey2024ai-powered,anand2023multi-label} and should be more widely adopted for future FER research.

\section{Conclusion}

    This systematic review has provided a comprehensive analysis of deep learning-based facial expression recognition systems with a focus on elderly populations. Our findings reveal significant progress in leveraging deep learning techniques, particularly convolutional neural networks, for vision-based FER tasks. Lightweight architectures, such as MobileNet and mini-Xception, have demonstrated potential for deployment in resource-constrained environments, while emerging approaches like Transformer-based models and multimodal frameworks offer promising avenues for enhancing robustness and accuracy. Despite these advances, several critical challenges remain.

    A notable limitation across studies is the underrepresentation of elderly individuals in FER datasets. The most widely used data sets, such as FER-2013 and AffectNet, lack sufficient diversity in age representation and fail to account for age-related facial changes that impact expression recognition. This highlights the urgent need to develop comprehensive and age-inclusive datasets or to merge existing ones into larger and more diverse datasets to improve FER performance across diverse age groups. Addressing class imbalance, another prevalent issue, will require broader adoption of more advanced rebalancing techniques, than simple undersampling and minority class removal, as found.
    
    The real-world deployment of FER systems for the elderly remains insufficiently explored. Privacy concerns, usability challenges, and the lack of explainable AI adoption hinder the integration of these technologies into healthcare and assisted living applications, which are of high stake for elderly users. Incorporating privacy-preserving measures, such as anonymized facial data or local processing, alongside lightweight models, could address privacy concerns and improve accessibility (as age typically increases disabilities). Furthermore, XAI techniques must be implemented more widely to foster user trust and transparency, allowing caregivers and medical professionals to better understand what is detected by the system and the predictions that are forecasted.
    
    Finally, while FER systems have immense potential to improve elderly care, additional research is needed to evaluate their societal acceptance, trustworthiness, and their impact on elder users when they are deployed. Future work should prioritize interdisciplinary efforts that combine technical innovation with human-centered design to ensure that these technologies are effective, ethical, and accessible. By addressing these challenges, FER systems can play a transformative role in enhancing the emotional well-being and quality of life of the aging population.

\section*{CRediT authorship contribution statement}
    \textbf{F. Xavier Gaya-Morey:} Conceptualization, Methodology, Validation, Investigation, Writing - Original Draft, Writing - Review \& Editing Preparation, Visualization. \textbf{Jose M. Buades-Rubio:} Conceptualization, Methodology, Writing - Review \& Editing Preparation, Supervision, Project administration, Funding acquisition. \textbf{Philippe Palanque:} Writing - Review \& Editing Preparation, Supervision. \textbf{Raquel Lacuesta:} Writing - Review \& Editing Preparation, Supervision. \textbf{Cristina Manresa-Yee:} Conceptualization, Methodology, Writing - Review \& Editing Preparation, Supervision, Project administration, Funding acquisition.

\section*{Funding}
   This work is part of the Project PID2023-149079OB-I00 (EXPLAINME) funded by MICIU/AEI/10.13039/ 501100011033/ and FEDER, EU and of Project PID2022-136779OB-C32 (PLEISAR) funded by MICIU/ AEI /10.13039/501100011033/ and FEDER, EU. F. X. Gaya-Morey was supported by an FPU scholarship from the Ministry of European Funds, University and Culture of the Government of the Balearic Islands.
    
\section*{Declaration of Competing Interest}
    The authors declare that they have no known competing financial interests or personal relationships that could have appeared to influence the work reported in this paper.
   
\section*{Data availability}
    All data relevant to the study are included in this document. No additional data were generated.

\section*{Declaration of generative AI and AI-assisted technologies in the writing process}
    During the preparation of this work the authors used ChatGPT in order to improve the readability and language of the manuscript. After using this tool, the authors reviewed and edited the content as needed and take full responsibility for the content of the published article.
    
%% If you have bib database file and want bibtex to generate the
%% bibitems, please use
%%
 \bibliographystyle{elsarticle-harv} 
 \bibliography{bibliography}

\begin{thebibliography}{193}
\expandafter\ifx\csname natexlab\endcsname\relax\def\natexlab#1{#1}\fi
\providecommand{\url}[1]{\texttt{#1}}
\providecommand{\href}[2]{#2}
\providecommand{\path}[1]{#1}
\providecommand{\DOIprefix}{doi:}
\providecommand{\ArXivprefix}{arXiv:}
\providecommand{\URLprefix}{URL: }
\providecommand{\Pubmedprefix}{pmid:}
\providecommand{\doi}[1]{\href{http://dx.doi.org/#1}{\path{#1}}}
\providecommand{\Pubmed}[1]{\href{pmid:#1}{\path{#1}}}
\providecommand{\bibinfo}[2]{#2}
\ifx\xfnm\relax \def\xfnm[#1]{\unskip,\space#1}\fi
%Type = Article
\bibitem[{Abdullah and Abdulazeez(2021)}]{msaleemabdullah2021facial}
\bibinfo{author}{Abdullah, S.M.S.}, \bibinfo{author}{Abdulazeez, A.M.}, \bibinfo{year}{2021}.
\newblock \bibinfo{title}{Facial expression recognition based on deep learning convolution neural network: A review}.
\newblock \bibinfo{journal}{Journal of Soft Computing and Data Mining} \bibinfo{volume}{2}, \bibinfo{pages}{53--65}.
%Type = Article
\bibitem[{Adadi and Berrada(2018)}]{adadi2018peeking}
\bibinfo{author}{Adadi, A.}, \bibinfo{author}{Berrada, M.}, \bibinfo{year}{2018}.
\newblock \bibinfo{title}{Peeking inside the black-box: A survey on explainable artificial intelligence (xai)}.
\newblock \bibinfo{journal}{IEEE Access} \bibinfo{volume}{6}, \bibinfo{pages}{52138--52160}.
\newblock \DOIprefix\doi{10.1109/ACCESS.2018.2870052}.
%Type = Article
\bibitem[{Adyapady and Annappa(2023)}]{rashmiadyapady2023comprehensive}
\bibinfo{author}{Adyapady, R.R.}, \bibinfo{author}{Annappa, B.}, \bibinfo{year}{2023}.
\newblock \bibinfo{title}{A comprehensive review of facial expression recognition techniques}.
\newblock \bibinfo{journal}{Multimedia Systems} \bibinfo{volume}{29}, \bibinfo{pages}{73--103}.
\newblock \DOIprefix\doi{10.1007/s00530-022-00984-w}.
%Type = Inproceedings
\bibitem[{Al-Garaawi and Morris(2016)}]{al_garaawi2016study}
\bibinfo{author}{Al-Garaawi, N.}, \bibinfo{author}{Morris, T.}, \bibinfo{year}{2016}.
\newblock \bibinfo{title}{Study on aging effect on facial expression recognition}, in: \bibinfo{booktitle}{Proceedings of the World Congress on Engineering}.
%Type = Article
\bibitem[{Al-Garaawi et~al.(2022)Al-Garaawi, Morris and Cootes}]{al_garaawi2022fully}
\bibinfo{author}{Al-Garaawi, N.}, \bibinfo{author}{Morris, T.}, \bibinfo{author}{Cootes, T.F.}, \bibinfo{year}{2022}.
\newblock \bibinfo{title}{Fully automated age-weighted expression classification using real and apparent age}.
\newblock \bibinfo{journal}{Pattern Analysis and Applications} \bibinfo{volume}{25}, \bibinfo{pages}{451--466}.
\newblock \DOIprefix\doi{10.1007/s10044-021-01044-1}.
%Type = Article
\bibitem[{Alexandre et~al.(2020)Alexandre, Soares and Thé}]{ribeiroalexandre2020systematic}
\bibinfo{author}{Alexandre, G.R.}, \bibinfo{author}{Soares, J.M.}, \bibinfo{author}{Thé, G.A.P.}, \bibinfo{year}{2020}.
\newblock \bibinfo{title}{Systematic review of 3d facial expression recognition methods}.
\newblock \bibinfo{journal}{Pattern Recognition} \bibinfo{volume}{100}, \bibinfo{pages}{107108}.
\newblock \DOIprefix\doi{10.1016/j.patcog.2019.107108}.
%Type = Article
\bibitem[{Almasoudi et~al.(2023)Almasoudi, Baowidan and Sarhan}]{almasoudi2023facial}
\bibinfo{author}{Almasoudi, A.}, \bibinfo{author}{Baowidan, S.}, \bibinfo{author}{Sarhan, S.}, \bibinfo{year}{2023}.
\newblock \bibinfo{title}{Facial expressions decoded: A survey of facial emotion recognition}.
\newblock \bibinfo{journal}{International Journal of Computer Applications} \bibinfo{volume}{185}, \bibinfo{pages}{1--11}.
\newblock \DOIprefix\doi{10.5120/ijca2023922765}.
%Type = Article
\bibitem[{Alves(2013)}]{alves2013recognition}
\bibinfo{author}{Alves, N.}, \bibinfo{year}{2013}.
\newblock \bibinfo{title}{Recognition of static and dynamic facial expressions: A study review}.
\newblock \bibinfo{journal}{Estudos de Psicologia (Natal)} \bibinfo{volume}{18}, \bibinfo{pages}{125--130}.
\newblock \DOIprefix\doi{10.1590/S1413-294X2013000100020}.
%Type = Misc
\bibitem[{{Amazon}(2016)}]{AmazonRekognition}
\bibinfo{author}{{Amazon}}, \bibinfo{year}{2016}.
\newblock \bibinfo{title}{What is amazon rekognition?}
\newblock \URLprefix \url{https://docs.aws.amazon.com/rekognition/latest/dg/what-is.html}. \bibinfo{note}{{Accessed 22nd Nov 2024}}.
%Type = Inproceedings
\bibitem[{Anand et~al.(2023)Anand, Devulapally, Bhattacharjee and Yuan}]{anand2023multi-label}
\bibinfo{author}{Anand, S.}, \bibinfo{author}{Devulapally, N.K.}, \bibinfo{author}{Bhattacharjee, S.D.}, \bibinfo{author}{Yuan, J.}, \bibinfo{year}{2023}.
\newblock \bibinfo{title}{Multi-label emotion analysis in conversation via multimodal knowledge distillation}, in: \bibinfo{booktitle}{Proceedings of the 31st ACM International Conference on Multimedia}, \bibinfo{publisher}{Association for Computing Machinery}. pp. \bibinfo{pages}{6090--6100}.
\newblock \DOIprefix\doi{10.1145/3581783.3612517}.
%Type = Inproceedings
\bibitem[{Asad et~al.(2017)Asad, Kashyap and Singh}]{asad2017recent}
\bibinfo{author}{Asad, U.}, \bibinfo{author}{Kashyap, N.}, \bibinfo{author}{Singh, S.N.}, \bibinfo{year}{2017}.
\newblock \bibinfo{title}{Recent advancements in facial expression recognition systems: A survey}, in: \bibinfo{booktitle}{2017 International Conference on Computing, Communication and Automation (ICCCA)}, pp. \bibinfo{pages}{1203--1208}.
\newblock \DOIprefix\doi{10.1109/CCAA.2017.8229981}.
%Type = Article
\bibitem[{Atallah et~al.(2019)Atallah, Kamsin and Ismail}]{raghebatallah2019review}
\bibinfo{author}{Atallah, R.R.}, \bibinfo{author}{Kamsin, A.}, \bibinfo{author}{Ismail, M.A.}, \bibinfo{year}{2019}.
\newblock \bibinfo{title}{A review study: The effect of face aging at estimating age and face recognition}.
\newblock \bibinfo{journal}{Journal of Physics: Conference Series} \bibinfo{volume}{1339}, \bibinfo{pages}{12006}.
\newblock \DOIprefix\doi{10.1088/1742-6596/1339/1/012006}.
%Type = Inproceedings
\bibitem[{Aziz et~al.(2022)Aziz, Khan, Aziz, Hasnul and Ismail}]{azlinaabaziz2022systematic}
\bibinfo{author}{Aziz, N.A.A.}, \bibinfo{author}{Khan, T.}, \bibinfo{author}{Aziz, K.A.}, \bibinfo{author}{Hasnul, M.A.}, \bibinfo{author}{Ismail, S.N.M.S.}, \bibinfo{year}{2022}.
\newblock \bibinfo{title}{A systematic review on facial expression based emotion recognition system for smart homes}, in: \bibinfo{booktitle}{Proceedings of the Multimedia University Engineering Conference (MECON 2022)}, \bibinfo{publisher}{Atlantis Press}. pp. \bibinfo{pages}{28--37}.
\newblock \DOIprefix\doi{10.2991/978-94-6463-082-4\_5}.
%Type = Article
\bibitem[{Bach et~al.(2017)Bach, Werner, Zywiec and Pluskiewicz}]{bach2017study}
\bibinfo{author}{Bach, M.}, \bibinfo{author}{Werner, A.}, \bibinfo{author}{Zywiec, J.}, \bibinfo{author}{Pluskiewicz, W.}, \bibinfo{year}{2017}.
\newblock \bibinfo{title}{The study of under- and over-sampling methods’ utility in analysis of highly imbalanced data on osteoporosis}.
\newblock \bibinfo{journal}{Information Sciences} \bibinfo{volume}{384}, \bibinfo{pages}{174--190}.
\newblock \DOIprefix\doi{10.1016/j.ins.2016.09.038}.
%Type = Inproceedings
\bibitem[{Bagher~Zadeh et~al.(2018)Bagher~Zadeh, Liang, Poria, Cambria and Morency}]{bagherzadeh2018multimodal}
\bibinfo{author}{Bagher~Zadeh, A.}, \bibinfo{author}{Liang, P.P.}, \bibinfo{author}{Poria, S.}, \bibinfo{author}{Cambria, E.}, \bibinfo{author}{Morency, L.P.}, \bibinfo{year}{2018}.
\newblock \bibinfo{title}{Multimodal language analysis in the wild: {CMU}-{MOSEI} dataset and interpretable dynamic fusion graph}, in: \bibinfo{editor}{Gurevych, I.}, \bibinfo{editor}{Miyao, Y.} (Eds.), \bibinfo{booktitle}{Proceedings of the 56th Annual Meeting of the Association for Computational Linguistics (Volume 1: Long Papers)}, \bibinfo{publisher}{Association for Computational Linguistics}, \bibinfo{address}{Melbourne, Australia}. pp. \bibinfo{pages}{2236--2246}.
\newblock \DOIprefix\doi{10.18653/v1/P18-1208}.
%Type = Inproceedings
\bibitem[{Baltrusaitis et~al.(2018)Baltrusaitis, Zadeh, Lim and Morency}]{baltrusaitis2018openface}
\bibinfo{author}{Baltrusaitis, T.}, \bibinfo{author}{Zadeh, A.}, \bibinfo{author}{Lim, Y.C.}, \bibinfo{author}{Morency, L.P.}, \bibinfo{year}{2018}.
\newblock \bibinfo{title}{Openface 2.0: Facial behavior analysis toolkit}, in: \bibinfo{booktitle}{2018 13th IEEE International Conference on Automatic Face \& Gesture Recognition (FG 2018)}, pp. \bibinfo{pages}{59--66}.
\newblock \DOIprefix\doi{10.1109/FG.2018.00019}.
%Type = Article
\bibitem[{{Barredo Arrieta} et~al.(2020){Barredo Arrieta}, Díaz-Rodríguez, {Del Ser}, Bennetot, Tabik, Barbado, Garcia, Gil-Lopez, Molina, Benjamins, Chatila and Herrera}]{barredoarrieta2020explainable}
\bibinfo{author}{{Barredo Arrieta}, A.}, \bibinfo{author}{Díaz-Rodríguez, N.}, \bibinfo{author}{{Del Ser}, J.}, \bibinfo{author}{Bennetot, A.}, \bibinfo{author}{Tabik, S.}, \bibinfo{author}{Barbado, A.}, \bibinfo{author}{Garcia, S.}, \bibinfo{author}{Gil-Lopez, S.}, \bibinfo{author}{Molina, D.}, \bibinfo{author}{Benjamins, R.}, \bibinfo{author}{Chatila, R.}, \bibinfo{author}{Herrera, F.}, \bibinfo{year}{2020}.
\newblock \bibinfo{title}{Explainable artificial intelligence (xai): Concepts, taxonomies, opportunities and challenges toward responsible ai}.
\newblock \bibinfo{journal}{Information Fusion} \bibinfo{volume}{58}, \bibinfo{pages}{82--115}.
%Type = Article
\bibitem[{Barrett et~al.(2019)Barrett, Adolphs, Marsella, Martinez and Pollak}]{barrett2019emotional}
\bibinfo{author}{Barrett, L.F.}, \bibinfo{author}{Adolphs, R.}, \bibinfo{author}{Marsella, S.}, \bibinfo{author}{Martinez, A.M.}, \bibinfo{author}{Pollak, S.D.}, \bibinfo{year}{2019}.
\newblock \bibinfo{title}{Emotional expressions reconsidered: Challenges to inferring emotion from human facial movements}.
\newblock \bibinfo{journal}{Psychological Science in the Public Interest} \bibinfo{volume}{20}, \bibinfo{pages}{1--68}.
\newblock \DOIprefix\doi{10.1177/1529100619832930}.
%Type = Inproceedings
\bibitem[{Bhattacharya and Gupta(2019)}]{bhattacharya2019survey}
\bibinfo{author}{Bhattacharya, S.}, \bibinfo{author}{Gupta, M.}, \bibinfo{year}{2019}.
\newblock \bibinfo{title}{A survey on: Facial emotion recognition invariant to pose, illumination and age}, in: \bibinfo{booktitle}{2019 Second International Conference on Advanced Computational and Communication Paradigms (ICACCP)}, pp. \bibinfo{pages}{1--6}.
\newblock \DOIprefix\doi{10.1109/ICACCP.2019.8883015}.
%Type = Article
\bibitem[{Bi et~al.(2022)Bi, Tian, Wang and Zhang}]{bi2022dynamic}
\bibinfo{author}{Bi, A.Q.}, \bibinfo{author}{Tian, X.Y.}, \bibinfo{author}{Wang, S.H.}, \bibinfo{author}{Zhang, Y.D.}, \bibinfo{year}{2022}.
\newblock \bibinfo{title}{Dynamic transfer exemplar based facial emotion recognition model toward online video}.
\newblock \bibinfo{journal}{ACM Trans. Multimedia Comput. Commun. Appl.} \bibinfo{volume}{18}.
\newblock \DOIprefix\doi{10.1145/3538385}.
%Type = Incollection
\bibitem[{Bloom and Luca(2016)}]{bloom2016chapter}
\bibinfo{author}{Bloom, D.}, \bibinfo{author}{Luca, D.}, \bibinfo{year}{2016}.
\newblock \bibinfo{title}{Chapter 1 - the global demography of aging: Facts, explanations, future}, in: \bibinfo{editor}{Piggott, J.}, \bibinfo{editor}{Woodland, A.} (Eds.), \bibinfo{booktitle}{Handbook of the Economics of Population Aging}. \bibinfo{publisher}{North-Holland}. volume~\bibinfo{volume}{1}, pp. \bibinfo{pages}{3--56}.
%Type = Article
\bibitem[{Boughanem et~al.(2023)Boughanem, Ghazouani and Barhoumi}]{boughanem2023facial}
\bibinfo{author}{Boughanem, H.}, \bibinfo{author}{Ghazouani, H.}, \bibinfo{author}{Barhoumi, W.}, \bibinfo{year}{2023}.
\newblock \bibinfo{title}{Facial emotion recognition in-the-wild using deep neural networks: A comprehensive review}.
\newblock \bibinfo{journal}{SN Computer Science} \bibinfo{volume}{5}, \bibinfo{pages}{96}.
\newblock \DOIprefix\doi{10.1007/s42979-023-02423-7}.
%Type = Article
\bibitem[{Branco et~al.(2016)Branco, Torgo and Ribeiro}]{branco2016survey}
\bibinfo{author}{Branco, P.}, \bibinfo{author}{Torgo, L.}, \bibinfo{author}{Ribeiro, R.P.}, \bibinfo{year}{2016}.
\newblock \bibinfo{title}{A survey of predictive modeling on imbalanced domains}.
\newblock \bibinfo{journal}{ACM Comput. Surv.} \bibinfo{volume}{49}.
\newblock \DOIprefix\doi{10.1145/2907070}.
%Type = Article
\bibitem[{Burkart and Huber(2021)}]{burkart2021survey}
\bibinfo{author}{Burkart, N.}, \bibinfo{author}{Huber, M.F.}, \bibinfo{year}{2021}.
\newblock \bibinfo{title}{A survey on the explainability of supervised machine learning}.
\newblock \bibinfo{journal}{Journal of Artificial Intelligence Research} \bibinfo{volume}{70}, \bibinfo{pages}{245--317}.
\newblock \DOIprefix\doi{10.1613/jair.1.12228}.
%Type = Article
\bibitem[{Canal et~al.(2022)Canal, Müller, Matias, Scotton, de~Sa~Junior, Pozzebon and Sobieranski}]{zagocanal2022survey}
\bibinfo{author}{Canal, F.Z.}, \bibinfo{author}{Müller, T.R.}, \bibinfo{author}{Matias, J.C.}, \bibinfo{author}{Scotton, G.G.}, \bibinfo{author}{de~Sa~Junior, A.R.}, \bibinfo{author}{Pozzebon, E.}, \bibinfo{author}{Sobieranski, A.C.}, \bibinfo{year}{2022}.
\newblock \bibinfo{title}{A survey on facial emotion recognition techniques: A state-of-the-art literature review}.
\newblock \bibinfo{journal}{Information Sciences} \bibinfo{volume}{582}, \bibinfo{pages}{593--617}.
\newblock \DOIprefix\doi{10.1016/j.ins.2021.10.005}.
%Type = Article
\bibitem[{Canedo and Neves(2019)}]{canedo2019facial}
\bibinfo{author}{Canedo, D.}, \bibinfo{author}{Neves, A.J.R.}, \bibinfo{year}{2019}.
\newblock \bibinfo{title}{Facial expression recognition using computer vision: A systematic review}.
\newblock \bibinfo{journal}{Applied Sciences} \bibinfo{volume}{9}.
\newblock \DOIprefix\doi{10.3390/app9214678}.
%Type = Article
\bibitem[{Cao et~al.(2021)Cao, Hidalgo, Simon, Wei and Sheikh}]{cao2021openpose}
\bibinfo{author}{Cao, Z.}, \bibinfo{author}{Hidalgo, G.}, \bibinfo{author}{Simon, T.}, \bibinfo{author}{Wei, S.E.}, \bibinfo{author}{Sheikh, Y.}, \bibinfo{year}{2021}.
\newblock \bibinfo{title}{Openpose: Realtime multi-person 2d pose estimation using part affinity fields}.
\newblock \bibinfo{journal}{IEEE Transactions on Pattern Analysis and Machine Intelligence} \bibinfo{volume}{43}, \bibinfo{pages}{172--186}.
\newblock \DOIprefix\doi{10.1109/TPAMI.2019.2929257}.
%Type = Inproceedings
\bibitem[{Carolis et~al.(2024)Carolis, Palestra and Castellano}]{nadjadecarolis2024exploring}
\bibinfo{author}{Carolis, B.N.D.}, \bibinfo{author}{Palestra, G.}, \bibinfo{author}{Castellano, G.}, \bibinfo{year}{2024}.
\newblock \bibinfo{title}{Exploring the role of empathy in designing social robots for elderly people}, in: \bibinfo{booktitle}{Adjunct Proceedings of the 32nd ACM Conference on User Modeling, Adaptation and Personalization}, \bibinfo{publisher}{Association for Computing Machinery}. pp. \bibinfo{pages}{120--125}.
\newblock \DOIprefix\doi{10.1145/3631700.3664887}.
%Type = Inproceedings
\bibitem[{Caroppo et~al.(2017)Caroppo, Leone and Siciliano}]{caroppo2017facial}
\bibinfo{author}{Caroppo, A.}, \bibinfo{author}{Leone, A.}, \bibinfo{author}{Siciliano, P.}, \bibinfo{year}{2017}.
\newblock \bibinfo{title}{Facial expression recognition in older adults using deep machine learning}, in: \bibinfo{booktitle}{AI*AAL@AI*IA}.
\newblock \URLprefix \url{https://api.semanticscholar.org/CorpusID:3526345}.
%Type = Inproceedings
\bibitem[{Caroppo et~al.(2018)Caroppo, Leone and Siciliano}]{caroppo2018facial}
\bibinfo{author}{Caroppo, A.}, \bibinfo{author}{Leone, A.}, \bibinfo{author}{Siciliano, P.}, \bibinfo{year}{2018}.
\newblock \bibinfo{title}{Facial expression recognition in older adults using deep machine learning}, in: \bibinfo{booktitle}{CEUR Workshop Proceedings}, p. \bibinfo{pages}{30 – 43}.
%Type = Inproceedings
\bibitem[{Caroppo et~al.(2019)Caroppo, Leone and Siciliano}]{caroppo2019facial}
\bibinfo{author}{Caroppo, A.}, \bibinfo{author}{Leone, A.}, \bibinfo{author}{Siciliano, P.}, \bibinfo{year}{2019}.
\newblock \bibinfo{title}{Facial expression recognition in ageing adults: A comparative study}, in: \bibinfo{editor}{Leone, A.}, \bibinfo{editor}{Caroppo, A.}, \bibinfo{editor}{Rescio, G.}, \bibinfo{editor}{Diraco, G.}, \bibinfo{editor}{Siciliano, P.} (Eds.), \bibinfo{booktitle}{Ambient Assisted Living}, \bibinfo{publisher}{Springer International Publishing}. pp. \bibinfo{pages}{349--359}.
%Type = Article
\bibitem[{Caroppo et~al.(2020)Caroppo, Leone and Siciliano}]{caroppo2020comparison}
\bibinfo{author}{Caroppo, A.}, \bibinfo{author}{Leone, A.}, \bibinfo{author}{Siciliano, P.}, \bibinfo{year}{2020}.
\newblock \bibinfo{title}{Comparison between deep learning models and traditional machine learning approaches for facial expression recognition in ageing adults}.
\newblock \bibinfo{journal}{Journal of Computer Science and Technology} \bibinfo{volume}{35}, \bibinfo{pages}{1127 – 1146}.
\newblock \DOIprefix\doi{10.1007/s11390-020-9665-4}.
%Type = Article
\bibitem[{Chen and Chen(2023)}]{chen2023emotion-reading}
\bibinfo{author}{Chen, S.Y.}, \bibinfo{author}{Chen, C.C.}, \bibinfo{year}{2023}.
\newblock \bibinfo{title}{Emotion-reading nursing care environment based on facial expression recognition}.
\newblock \bibinfo{journal}{SENSORS AND MATERIALS} \bibinfo{volume}{35}, \bibinfo{pages}{1859--1869}.
\newblock \DOIprefix\doi{10.18494/SAM4314}.
%Type = Article
\bibitem[{Chinchanikar(2019)}]{achinchanikar2019facial}
\bibinfo{author}{Chinchanikar, N.A.}, \bibinfo{year}{2019}.
\newblock \bibinfo{title}{Facial expression recognition using deep learning: a review}.
\newblock \bibinfo{journal}{International Research Journal of Engineering and Technology} \bibinfo{volume}{3274}.
%Type = Inproceedings
\bibitem[{Cho et~al.(2014)Cho, van Merri{\"e}nboer, Bahdanau and Bengio}]{GRU}
\bibinfo{author}{Cho, K.}, \bibinfo{author}{van Merri{\"e}nboer, B.}, \bibinfo{author}{Bahdanau, D.}, \bibinfo{author}{Bengio, Y.}, \bibinfo{year}{2014}.
\newblock \bibinfo{title}{On the properties of neural machine translation: Encoder{--}decoder approaches}, in: \bibinfo{editor}{Wu, D.}, \bibinfo{editor}{Carpuat, M.}, \bibinfo{editor}{Carreras, X.}, \bibinfo{editor}{Vecchi, E.M.} (Eds.), \bibinfo{booktitle}{Proceedings of {SSST}-8, Eighth Workshop on Syntax, Semantics and Structure in Statistical Translation}, \bibinfo{publisher}{Association for Computational Linguistics}, \bibinfo{address}{Doha, Qatar}. pp. \bibinfo{pages}{103--111}.
\newblock \DOIprefix\doi{10.3115/v1/W14-4012}.
%Type = Inproceedings
\bibitem[{Chollet(2017)}]{Xception}
\bibinfo{author}{Chollet, F.}, \bibinfo{year}{2017}.
\newblock \bibinfo{title}{Xception: Deep learning with depthwise separable convolutions}, in: \bibinfo{booktitle}{2017 IEEE Conference on Computer Vision and Pattern Recognition (CVPR)}, pp. \bibinfo{pages}{1800--1807}.
\newblock \DOIprefix\doi{10.1109/CVPR.2017.195}.
%Type = Inproceedings
\bibitem[{Cui et~al.(2019)Cui, Jia, Lin, Song and Belongie}]{cui2019class-balanced}
\bibinfo{author}{Cui, Y.}, \bibinfo{author}{Jia, M.}, \bibinfo{author}{Lin, T.Y.}, \bibinfo{author}{Song, Y.}, \bibinfo{author}{Belongie, S.}, \bibinfo{year}{2019}.
\newblock \bibinfo{title}{Class-balanced loss based on effective number of samples}, in: \bibinfo{booktitle}{2019 IEEE/CVF Conference on Computer Vision and Pattern Recognition (CVPR)}, pp. \bibinfo{pages}{9260--9269}.
\newblock \DOIprefix\doi{10.1109/CVPR.2019.00949}.
%Type = Article
\bibitem[{Dalvi et~al.(2021)Dalvi, Rathod, Patil, Gite and Kotecha}]{dalvi2021survey}
\bibinfo{author}{Dalvi, C.}, \bibinfo{author}{Rathod, M.}, \bibinfo{author}{Patil, S.}, \bibinfo{author}{Gite, S.}, \bibinfo{author}{Kotecha, K.}, \bibinfo{year}{2021}.
\newblock \bibinfo{title}{A survey of ai-based facial emotion recognition: Features, ml \& dl techniques, age-wise datasets and future directions}.
\newblock \bibinfo{journal}{IEEE Access} \bibinfo{volume}{9}, \bibinfo{pages}{165806--165840}.
\newblock \DOIprefix\doi{10.1109/ACCESS.2021.3131733}.
%Type = Misc
\bibitem[{Dhall et~al.(2007)Dhall, Member, Goecke, Lucey and Gedeon}]{dhall2007collecting}
\bibinfo{author}{Dhall, A.}, \bibinfo{author}{Member, S.}, \bibinfo{author}{Goecke, R.}, \bibinfo{author}{Lucey, S.}, \bibinfo{author}{Gedeon, T.}, \bibinfo{year}{2007}.
\newblock \bibinfo{title}{Collecting large, richly annotated facial-expression databases from movies}.
%Type = Misc
\bibitem[{Dosovitskiy et~al.(2021)Dosovitskiy, Beyer, Kolesnikov, Weissenborn, Zhai, Unterthiner, Dehghani, Minderer, Heigold, Gelly, Uszkoreit and Houlsby}]{dosovitskiy2021image}
\bibinfo{author}{Dosovitskiy, A.}, \bibinfo{author}{Beyer, L.}, \bibinfo{author}{Kolesnikov, A.}, \bibinfo{author}{Weissenborn, D.}, \bibinfo{author}{Zhai, X.}, \bibinfo{author}{Unterthiner, T.}, \bibinfo{author}{Dehghani, M.}, \bibinfo{author}{Minderer, M.}, \bibinfo{author}{Heigold, G.}, \bibinfo{author}{Gelly, S.}, \bibinfo{author}{Uszkoreit, J.}, \bibinfo{author}{Houlsby, N.}, \bibinfo{year}{2021}.
\newblock \bibinfo{title}{An image is worth 16x16 words: Transformers for image recognition at scale}.
\newblock \DOIprefix\doi{10.48550/arXiv.2010.11929}.
%Type = Article
\bibitem[{Ebner et~al.(2010a)Ebner, Riediger and Lindenberger}]{ebner2010faces}
\bibinfo{author}{Ebner, N.C.}, \bibinfo{author}{Riediger, M.}, \bibinfo{author}{Lindenberger, U.}, \bibinfo{year}{2010}a.
\newblock \bibinfo{title}{Faces-a database of facial expressions in young, middle-aged, and older women and men: Development and validation}.
\newblock \bibinfo{journal}{Behavior Research Methods} \bibinfo{volume}{42}, \bibinfo{pages}{351--362}.
\newblock \DOIprefix\doi{10.3758/BRM.42.1.351}.
%Type = Article
\bibitem[{Ebner et~al.(2010b)Ebner, Riediger and Lindenberger}]{cebner2010faces-a}
\bibinfo{author}{Ebner, N.C.}, \bibinfo{author}{Riediger, M.}, \bibinfo{author}{Lindenberger, U.}, \bibinfo{year}{2010}b.
\newblock \bibinfo{title}{Faces-a database of facial expressions in young, middle-aged, and older women and men: Development and validation}.
\newblock \bibinfo{journal}{Behavior Research Methods} \bibinfo{volume}{42}, \bibinfo{pages}{351--362}.
\newblock \DOIprefix\doi{10.3758/BRM.42.1.351}.
%Type = Article
\bibitem[{Ekman(1992)}]{ekman1992argument}
\bibinfo{author}{Ekman, P.}, \bibinfo{year}{1992}.
\newblock \bibinfo{title}{An argument for basic emotions}.
\newblock \bibinfo{journal}{Cognition and Emotion} \bibinfo{volume}{6}, \bibinfo{pages}{169--200}.
\newblock \DOIprefix\doi{10.1080/02699939208411068}.
%Type = Article
\bibitem[{Ekman and Friesen(1978)}]{ekman1978facial}
\bibinfo{author}{Ekman, P.}, \bibinfo{author}{Friesen, W.V.}, \bibinfo{year}{1978}.
\newblock \bibinfo{title}{Facial action coding system: a technique for the measurement of facial movement}.
\newblock \bibinfo{journal}{Palo Alto} \bibinfo{volume}{3}, \bibinfo{pages}{5}.
%Type = Article
\bibitem[{Fahn et~al.(2022)Fahn, Chen, Wu, Chu, Li, Hsu, Wang and Tsai}]{fahn2022image}
\bibinfo{author}{Fahn, C.S.}, \bibinfo{author}{Chen, S.C.}, \bibinfo{author}{Wu, P.Y.}, \bibinfo{author}{Chu, T.L.}, \bibinfo{author}{Li, C.H.}, \bibinfo{author}{Hsu, D.Q.}, \bibinfo{author}{Wang, H.H.}, \bibinfo{author}{Tsai, H.M.}, \bibinfo{year}{2022}.
\newblock \bibinfo{title}{Image and speech recognition technology in the development of an elderly care robot: Practical issues review and improvement strategies}.
\newblock \bibinfo{journal}{Healthcare (Switzerland)} \bibinfo{volume}{10}.
\newblock \DOIprefix\doi{10.3390/healthcare10112252}.
%Type = Article
\bibitem[{Fan et~al.(2022)Fan, Wang, Zhu, Cao, Yi, Chen, Jia and Lu}]{fan2022fer-pcvt}
\bibinfo{author}{Fan, Y.}, \bibinfo{author}{Wang, H.}, \bibinfo{author}{Zhu, X.}, \bibinfo{author}{Cao, X.}, \bibinfo{author}{Yi, C.}, \bibinfo{author}{Chen, Y.}, \bibinfo{author}{Jia, J.}, \bibinfo{author}{Lu, X.}, \bibinfo{year}{2022}.
\newblock \bibinfo{title}{Fer-pcvt: Facial expression recognition with patch-convolutional vision transformer for stroke patients}.
\newblock \bibinfo{journal}{Brain Sciences} \bibinfo{volume}{12}.
\newblock \DOIprefix\doi{10.3390/brainsci12121626}.
%Type = Article
\bibitem[{Fei et~al.(2019)Fei, Yang, Li, Butler, Ijomah and Zhou}]{fei2019survey}
\bibinfo{author}{Fei, Z.}, \bibinfo{author}{Yang, E.}, \bibinfo{author}{Li, D.D.U.}, \bibinfo{author}{Butler, S.}, \bibinfo{author}{Ijomah, W.}, \bibinfo{author}{Zhou, H.}, \bibinfo{year}{2019}.
\newblock \bibinfo{title}{A survey on computer vision techniques for detecting facial features towards the early diagnosis of mild cognitive impairment in the elderly}.
\newblock \bibinfo{journal}{Systems Science \& Control Engineering} \bibinfo{volume}{7}, \bibinfo{pages}{252--263}.
\newblock \DOIprefix\doi{10.1080/21642583.2019.1647577}.
%Type = Article
\bibitem[{Fei et~al.(2022)Fei, Yang, Yu, Li, Zhou and Zhou}]{fei2022novel}
\bibinfo{author}{Fei, Z.}, \bibinfo{author}{Yang, E.}, \bibinfo{author}{Yu, L.}, \bibinfo{author}{Li, X.}, \bibinfo{author}{Zhou, H.}, \bibinfo{author}{Zhou, W.}, \bibinfo{year}{2022}.
\newblock \bibinfo{title}{A novel deep neural network-based emotion analysis system for automatic detection of mild cognitive impairment in the elderly}.
\newblock \bibinfo{journal}{Neurocomputing} \bibinfo{volume}{468}, \bibinfo{pages}{306 – 316}.
\newblock \DOIprefix\doi{10.1016/j.neucom.2021.10.038}.
%Type = Article
\bibitem[{Fölster et~al.(2014)Fölster, Hess and Werheid}]{fölster2014facial}
\bibinfo{author}{Fölster, M.}, \bibinfo{author}{Hess, U.}, \bibinfo{author}{Werheid, K.}, \bibinfo{year}{2014}.
\newblock \bibinfo{title}{Facial age affects emotional expression decoding}.
\newblock \bibinfo{journal}{Frontiers in Psychology} \bibinfo{volume}{5}.
\newblock \DOIprefix\doi{10.3389/fpsyg.2014.00030}.
%Type = Inproceedings
\bibitem[{Gaya-Morey et~al.(2024)Gaya-Morey, Manresa-Yee and Buades-Rubio}]{gaya-morey2024ai-powered}
\bibinfo{author}{Gaya-Morey, F.X.}, \bibinfo{author}{Manresa-Yee, C.}, \bibinfo{author}{Buades-Rubio, J.M.}, \bibinfo{year}{2024}.
\newblock \bibinfo{title}{An ai-powered computer vision module for social interactive agents}, in: \bibinfo{booktitle}{Proceedings of the XXIV International Conference on Human Computer Interaction}, \bibinfo{publisher}{Association for Computing Machinery}.
\newblock \DOIprefix\doi{10.1145/3657242.3658601}.
%Type = Article
\bibitem[{Ghayoumi(2017)}]{ghayoumi2017quick}
\bibinfo{author}{Ghayoumi, M.}, \bibinfo{year}{2017}.
\newblock \bibinfo{title}{A quick review of deep learning in facial expression}.
\newblock \bibinfo{journal}{Journal of Communication and Computer} \bibinfo{volume}{14}.
\newblock \DOIprefix\doi{10.17265/1548-7709/2017.01.004}.
%Type = Article
\bibitem[{Gold et~al.(2013)Gold, Barker, Barr, Bittner, Bromfield, Chu, Goode, Lee, Simmons and Srinath}]{gold2013efficiency}
\bibinfo{author}{Gold, J.M.}, \bibinfo{author}{Barker, J.D.}, \bibinfo{author}{Barr, S.}, \bibinfo{author}{Bittner, J.L.}, \bibinfo{author}{Bromfield, W.D.}, \bibinfo{author}{Chu, N.}, \bibinfo{author}{Goode, R.A.}, \bibinfo{author}{Lee, D.}, \bibinfo{author}{Simmons, M.}, \bibinfo{author}{Srinath, A.}, \bibinfo{year}{2013}.
\newblock \bibinfo{title}{The efficiency of dynamic and static facial expression recognition}.
\newblock \bibinfo{journal}{Journal of Vision} \bibinfo{volume}{13}, \bibinfo{pages}{23--23}.
\newblock \DOIprefix\doi{10.1167/13.5.23}.
%Type = Misc
\bibitem[{Goodfellow et~al.(2014)Goodfellow, Pouget-Abadie, Mirza, Xu, Warde-Farley, Ozair, Courville and Bengio}]{GAN}
\bibinfo{author}{Goodfellow, I.}, \bibinfo{author}{Pouget-Abadie, J.}, \bibinfo{author}{Mirza, M.}, \bibinfo{author}{Xu, B.}, \bibinfo{author}{Warde-Farley, D.}, \bibinfo{author}{Ozair, S.}, \bibinfo{author}{Courville, A.}, \bibinfo{author}{Bengio, Y.}, \bibinfo{year}{2014}.
\newblock \bibinfo{title}{Generative adversarial nets}.
%Type = Inproceedings
\bibitem[{Goodfellow et~al.(2013)Goodfellow, Erhan, Carrier, Courville, Mirza, Hamner, Cukierski, Tang, Thaler, Lee, Zhou, Ramaiah, Feng, Li, Wang, Athanasakis, Shawe-Taylor, Milakov, Park, Ionescu, Popescu, Grozea, Bergstra, Xie, Romaszko, Xu, Chuang and Bengio}]{goodfellow2013challenges}
\bibinfo{author}{Goodfellow, I.J.}, \bibinfo{author}{Erhan, D.}, \bibinfo{author}{Carrier, P.L.}, \bibinfo{author}{Courville, A.}, \bibinfo{author}{Mirza, M.}, \bibinfo{author}{Hamner, B.}, \bibinfo{author}{Cukierski, W.}, \bibinfo{author}{Tang, Y.}, \bibinfo{author}{Thaler, D.}, \bibinfo{author}{Lee, D.H.}, \bibinfo{author}{Zhou, Y.}, \bibinfo{author}{Ramaiah, C.}, \bibinfo{author}{Feng, F.}, \bibinfo{author}{Li, R.}, \bibinfo{author}{Wang, X.}, \bibinfo{author}{Athanasakis, D.}, \bibinfo{author}{Shawe-Taylor, J.}, \bibinfo{author}{Milakov, M.}, \bibinfo{author}{Park, J.}, \bibinfo{author}{Ionescu, R.}, \bibinfo{author}{Popescu, M.}, \bibinfo{author}{Grozea, C.}, \bibinfo{author}{Bergstra, J.}, \bibinfo{author}{Xie, J.}, \bibinfo{author}{Romaszko, L.}, \bibinfo{author}{Xu, B.}, \bibinfo{author}{Chuang, Z.}, \bibinfo{author}{Bengio, Y.}, \bibinfo{year}{2013}.
\newblock \bibinfo{title}{Challenges in representation learning: A report on three machine learning contests}, in: \bibinfo{editor}{Lee, M.}, \bibinfo{editor}{Hirose, A.}, \bibinfo{editor}{Hou, Z.G.}, \bibinfo{editor}{Kil, R.M.} (Eds.), \bibinfo{booktitle}{Neural Information Processing}, \bibinfo{publisher}{Springer Berlin Heidelberg}, \bibinfo{address}{Berlin, Heidelberg}. pp. \bibinfo{pages}{117--124}.
%Type = Article
\bibitem[{Grabowski et~al.(2019)Grabowski, Rynkiewicz, Lassalle, Baron-Cohen, Schuller, Cummins, Baird, Podg{\'{o}}rska-Bednarz, Pieni{\c{a}}{\.{z}}ek and {\L}ucka}]{grabowski2019emotional}
\bibinfo{author}{Grabowski, K.}, \bibinfo{author}{Rynkiewicz, A.}, \bibinfo{author}{Lassalle, A.}, \bibinfo{author}{Baron-Cohen, S.}, \bibinfo{author}{Schuller, B.}, \bibinfo{author}{Cummins, N.}, \bibinfo{author}{Baird, A.}, \bibinfo{author}{Podg{\'{o}}rska-Bednarz, J.}, \bibinfo{author}{Pieni{\c{a}}{\.{z}}ek, A.}, \bibinfo{author}{{\L}ucka, I.}, \bibinfo{year}{2019}.
\newblock \bibinfo{title}{Emotional expression in psychiatric conditions: New technology for clinicians}.
\newblock \bibinfo{journal}{Psychiatry and Clinical Neurosciences} \bibinfo{volume}{73}, \bibinfo{pages}{50--62}.
\newblock \DOIprefix\doi{10.1111/pcn.12799}.
%Type = Article
\bibitem[{Grondhuis et~al.(2021)Grondhuis, Jimmy, Teague and Brunet}]{ngrondhuis2021having}
\bibinfo{author}{Grondhuis, S.N.}, \bibinfo{author}{Jimmy, A.}, \bibinfo{author}{Teague, C.}, \bibinfo{author}{Brunet, N.M.}, \bibinfo{year}{2021}.
\newblock \bibinfo{title}{Having difficulties reading the facial expression of older individuals? blame it on the facial muscles, not the wrinkles}.
\newblock \bibinfo{journal}{Frontiers in Psychology} \bibinfo{volume}{12}.
\newblock \DOIprefix\doi{10.3389/fpsyg.2021.620768}.
%Type = Article
\bibitem[{Guerdelli et~al.(2022)Guerdelli, Ferrari, Barhoumi, Ghazouani and Berretti}]{guerdelli2022macro}
\bibinfo{author}{Guerdelli, H.}, \bibinfo{author}{Ferrari, C.}, \bibinfo{author}{Barhoumi, W.}, \bibinfo{author}{Ghazouani, H.}, \bibinfo{author}{Berretti, S.}, \bibinfo{year}{2022}.
\newblock \bibinfo{title}{Macro- and micro-expressions facial datasets: A survey}.
\newblock \bibinfo{journal}{Sensors} \bibinfo{volume}{22}.
\newblock \DOIprefix\doi{10.3390/s22041524}.
%Type = Article
\bibitem[{Gunning et~al.(2019)Gunning, Stefik, Choi, Miller, Stumpf and Yang}]{gunning2019xai—explainable}
\bibinfo{author}{Gunning, D.}, \bibinfo{author}{Stefik, M.}, \bibinfo{author}{Choi, J.}, \bibinfo{author}{Miller, T.}, \bibinfo{author}{Stumpf, S.}, \bibinfo{author}{Yang, G.Z.}, \bibinfo{year}{2019}.
\newblock \bibinfo{title}{Xai—explainable artificial intelligence}.
\newblock \bibinfo{journal}{Science Robotics} \bibinfo{volume}{4}, \bibinfo{pages}{eaay7120}.
\newblock \DOIprefix\doi{10.1126/scirobotics.aay7120}.
%Type = Article
\bibitem[{Guo et~al.(2013)Guo, Guo and Li}]{guo2013facial}
\bibinfo{author}{Guo, G.}, \bibinfo{author}{Guo, R.}, \bibinfo{author}{Li, X.}, \bibinfo{year}{2013}.
\newblock \bibinfo{title}{Facial expression recognition influenced by human aging}.
\newblock \bibinfo{journal}{IEEE Transactions on Affective Computing} \bibinfo{volume}{4}, \bibinfo{pages}{291--298}.
\newblock \DOIprefix\doi{10.1109/T-AFFC.2013.13}.
%Type = Misc
\bibitem[{Guo et~al.(2021)Guo, Deng, Lattas and Zafeiriou}]{guo2021sample}
\bibinfo{author}{Guo, J.}, \bibinfo{author}{Deng, J.}, \bibinfo{author}{Lattas, A.}, \bibinfo{author}{Zafeiriou, S.}, \bibinfo{year}{2021}.
\newblock \bibinfo{title}{Sample and computation redistribution for efficient face detection}.
\newblock \DOIprefix\doi{10.48550/arXiv.2105.04714}.
%Type = Misc
\bibitem[{Haunschmid et~al.(2020)Haunschmid, Manilow and Widmer}]{haunschmid2020audiolime}
\bibinfo{author}{Haunschmid, V.}, \bibinfo{author}{Manilow, E.}, \bibinfo{author}{Widmer, G.}, \bibinfo{year}{2020}.
\newblock \bibinfo{title}{audiolime: Listenable explanations using source separation}.
\newblock \DOIprefix\doi{10.48550/arXiv.2008.00582}, \href{http://arxiv.org/abs/2008.00582}{{\tt arXiv:2008.00582}}.
%Type = Inproceedings
\bibitem[{He et~al.(2016a)He, Zhang, Ren and Sun}]{ResNet}
\bibinfo{author}{He, K.}, \bibinfo{author}{Zhang, X.}, \bibinfo{author}{Ren, S.}, \bibinfo{author}{Sun, J.}, \bibinfo{year}{2016}a.
\newblock \bibinfo{title}{Deep residual learning for image recognition}, in: \bibinfo{booktitle}{2016 IEEE Conference on Computer Vision and Pattern Recognition (CVPR)}, pp. \bibinfo{pages}{770--778}.
%Type = Inproceedings
\bibitem[{He et~al.(2016b)He, Zhang, Ren and Sun}]{he2016deep}
\bibinfo{author}{He, K.}, \bibinfo{author}{Zhang, X.}, \bibinfo{author}{Ren, S.}, \bibinfo{author}{Sun, J.}, \bibinfo{year}{2016}b.
\newblock \bibinfo{title}{Deep residual learning for image recognition}, in: \bibinfo{booktitle}{2016 IEEE Conference on Computer Vision and Pattern Recognition (CVPR)}, pp. \bibinfo{pages}{770--778}.
\newblock \DOIprefix\doi{10.1109/CVPR.2016.90}.
%Type = Inproceedings
\bibitem[{Hernandez-Ortega et~al.(2019)Hernandez-Ortega, Galbally, Fierrez, Haraksim and Beslay}]{hernandez-ortega2019faceqnet}
\bibinfo{author}{Hernandez-Ortega, J.}, \bibinfo{author}{Galbally, J.}, \bibinfo{author}{Fierrez, J.}, \bibinfo{author}{Haraksim, R.}, \bibinfo{author}{Beslay, L.}, \bibinfo{year}{2019}.
\newblock \bibinfo{title}{Faceqnet: Quality assessment for face recognition based on deep learning}, in: \bibinfo{booktitle}{2019 International Conference on Biometrics (ICB)}, pp. \bibinfo{pages}{1--8}.
\newblock \DOIprefix\doi{10.1109/ICB45273.2019.8987255}.
%Type = Article
\bibitem[{Hinton et~al.(2006)Hinton, Osindero and Teh}]{hinton2006fast}
\bibinfo{author}{Hinton, G.E.}, \bibinfo{author}{Osindero, S.}, \bibinfo{author}{Teh, Y.W.}, \bibinfo{year}{2006}.
\newblock \bibinfo{title}{A fast learning algorithm for deep belief nets}.
\newblock \bibinfo{journal}{Neural Computation} \bibinfo{volume}{18}, \bibinfo{pages}{1527--1554}.
\newblock \DOIprefix\doi{10.1162/neco.2006.18.7.1527}.
%Type = Article
\bibitem[{Hochreiter and Schmidhuber(1997)}]{LSTM}
\bibinfo{author}{Hochreiter, S.}, \bibinfo{author}{Schmidhuber, J.}, \bibinfo{year}{1997}.
\newblock \bibinfo{title}{Long short-term memory}.
\newblock \bibinfo{journal}{Neural Computation} \bibinfo{volume}{9}, \bibinfo{pages}{1735--1780}.
%Type = Article
\bibitem[{Hopfield(1982)}]{jhopfield1982neural}
\bibinfo{author}{Hopfield, J.J.}, \bibinfo{year}{1982}.
\newblock \bibinfo{title}{Neural networks and physical systems with emergent collective computational abilities.}
\newblock \bibinfo{journal}{Proceedings of the National Academy of Sciences} \bibinfo{volume}{79}, \bibinfo{pages}{2554--2558}.
\newblock \DOIprefix\doi{10.1073/pnas.79.8.2554}.
%Type = Misc
\bibitem[{Howard et~al.(2017)Howard, Zhu, Chen, Kalenichenko, Wang, Weyand, Andreetto and Adam}]{MobileNet}
\bibinfo{author}{Howard, A.G.}, \bibinfo{author}{Zhu, M.}, \bibinfo{author}{Chen, B.}, \bibinfo{author}{Kalenichenko, D.}, \bibinfo{author}{Wang, W.}, \bibinfo{author}{Weyand, T.}, \bibinfo{author}{Andreetto, M.}, \bibinfo{author}{Adam, H.}, \bibinfo{year}{2017}.
\newblock \bibinfo{title}{Mobilenets: Efficient convolutional neural networks for mobile vision applications}.
\newblock \DOIprefix\doi{10.48550/arXiv.1704.04861}.
%Type = Article
\bibitem[{Htay(2021)}]{moehtay2021feature}
\bibinfo{author}{Htay, M.M.}, \bibinfo{year}{2021}.
\newblock \bibinfo{title}{Feature extraction and classification methods of facial expression: A survey}.
\newblock \bibinfo{journal}{Computer Science and Information Technologies} \bibinfo{volume}{2}, \bibinfo{pages}{26--32}.
\newblock \DOIprefix\doi{10.11591/csit.v2i1.p26-32}.
%Type = Article
\bibitem[{Huang et~al.(2024a)Huang, Xu, Wan, Zhang, Zha and Pang}]{huang2024auto}
\bibinfo{author}{Huang, W.}, \bibinfo{author}{Xu, W.}, \bibinfo{author}{Wan, R.}, \bibinfo{author}{Zhang, P.}, \bibinfo{author}{Zha, Y.}, \bibinfo{author}{Pang, M.}, \bibinfo{year}{2024}a.
\newblock \bibinfo{title}{Auto diagnosis of parkinson's disease via a deep learning model based on mixed emotional facial expressions}.
\newblock \bibinfo{journal}{IEEE Journal of Biomedical and Health Informatics} \bibinfo{volume}{28}, \bibinfo{pages}{2547--2557}.
\newblock \DOIprefix\doi{10.1109/JBHI.2023.3239780}.
%Type = Inproceedings
\bibitem[{Huang et~al.(2024b)Huang, Peng, Cai, Guo, Chen and Tan}]{huang2024facial}
\bibinfo{author}{Huang, Y.}, \bibinfo{author}{Peng, J.}, \bibinfo{author}{Cai, Z.}, \bibinfo{author}{Guo, J.}, \bibinfo{author}{Chen, G.}, \bibinfo{author}{Tan, S.}, \bibinfo{year}{2024}b.
\newblock \bibinfo{title}{Facial expression recognition with age-group expression feature learning}, in: \bibinfo{booktitle}{2024 International Joint Conference on Neural Networks (IJCNN)}, pp. \bibinfo{pages}{1--8}.
\newblock \DOIprefix\doi{10.1109/IJCNN60899.2024.10649944}.
%Type = Misc
\bibitem[{Hussain(2021)}]{hussain2021human}
\bibinfo{author}{Hussain, J.}, \bibinfo{year}{2021}.
\newblock \bibinfo{title}{Human emotions}.
\newblock \URLprefix \url{https://www.kaggle.com/datasets/jafarhussain786}. \bibinfo{note}{{Accessed: 27th Nov 2024}}.
%Type = Misc
\bibitem[{Iandola et~al.(2016)Iandola, Han, Moskewicz, Ashraf, Dally and Keutzer}]{niandola2016squeezenet}
\bibinfo{author}{Iandola, F.N.}, \bibinfo{author}{Han, S.}, \bibinfo{author}{Moskewicz, M.W.}, \bibinfo{author}{Ashraf, K.}, \bibinfo{author}{Dally, W.J.}, \bibinfo{author}{Keutzer, K.}, \bibinfo{year}{2016}.
\newblock \bibinfo{title}{Squeezenet: Alexnet-level accuracy with 50x fewer parameters and $<0.5${MB} model size}.
\newblock \DOIprefix\doi{10.48550/arXiv.1602.07360}.
%Type = Article
\bibitem[{Isaacowitz and Stanley(2011)}]{isaacowitz2011bringing}
\bibinfo{author}{Isaacowitz, D.M.}, \bibinfo{author}{Stanley, J.T.}, \bibinfo{year}{2011}.
\newblock \bibinfo{title}{Bringing an ecological perspective to the study of aging and recognition of emotional facial expressions: Past, current, and future methods}.
\newblock \bibinfo{journal}{Journal of Nonverbal Behavior} \bibinfo{volume}{35}, \bibinfo{pages}{261--278}.
\newblock \DOIprefix\doi{10.1007/s10919-011-0113-6}.
%Type = Inproceedings
\bibitem[{Jain et~al.(2021)Jain, Quadri and Lalit}]{raijain2021recent}
\bibinfo{author}{Jain, P.R.}, \bibinfo{author}{Quadri, S.M.K.}, \bibinfo{author}{Lalit, M.}, \bibinfo{year}{2021}.
\newblock \bibinfo{title}{Recent trends in artificial intelligence for emotion detection using facial image analysis}, in: \bibinfo{booktitle}{Proceedings of the 2021 Thirteenth International Conference on Contemporary Computing}, \bibinfo{publisher}{Association for Computing Machinery}. pp. \bibinfo{pages}{18--36}.
\newblock \DOIprefix\doi{10.1145/3474124.3474205}.
%Type = Inproceedings
\bibitem[{Jannat and Canavan(2021)}]{rahatuljannat2021expression}
\bibinfo{author}{Jannat, S.R.}, \bibinfo{author}{Canavan, S.}, \bibinfo{year}{2021}.
\newblock \bibinfo{title}{Expression recognition across age}, in: \bibinfo{booktitle}{2021 16th IEEE International Conference on Automatic Face and Gesture Recognition (FG 2021)}, pp. \bibinfo{pages}{1--5}.
\newblock \DOIprefix\doi{10.1109/FG52635.2021.9667062}.
%Type = Inproceedings
\bibitem[{Jayanthi et~al.(2024)Jayanthi, Nirmaladevi, Bharathi, Krithiknimalan and Vimalkumar}]{jayanthi2024enhanced}
\bibinfo{author}{Jayanthi, P.}, \bibinfo{author}{Nirmaladevi, K.}, \bibinfo{author}{Bharathi, K.C.K.}, \bibinfo{author}{Krithiknimalan, S.K.}, \bibinfo{author}{Vimalkumar, S.}, \bibinfo{year}{2024}.
\newblock \bibinfo{title}{An enhanced method for recognition of facial expressions using convolutional neural network}, in: \bibinfo{booktitle}{Proceedings - 2024 5th International Conference on Intelligent Communication Technologies and Virtual Mobile Networks, ICICV 2024}, p. \bibinfo{pages}{173 – 175}.
\newblock \DOIprefix\doi{10.1109/ICICV62344.2024.00033}.
%Type = Article
\bibitem[{Jiang et~al.(2022)Jiang, Seyedi, Haque, Pongos, Vickers, Manzanares, Lah, Levey and Clifford}]{jiang2022automated}
\bibinfo{author}{Jiang, Z.}, \bibinfo{author}{Seyedi, S.}, \bibinfo{author}{Haque, R.U.}, \bibinfo{author}{Pongos, A.L.}, \bibinfo{author}{Vickers, K.L.}, \bibinfo{author}{Manzanares, C.M.}, \bibinfo{author}{Lah, J.J.}, \bibinfo{author}{Levey, A.I.}, \bibinfo{author}{Clifford, G.D.}, \bibinfo{year}{2022}.
\newblock \bibinfo{title}{Automated analysis of facial emotions in subjects with cognitive impairment}.
\newblock \bibinfo{journal}{PLoS ONE} \bibinfo{volume}{17}.
\newblock \DOIprefix\doi{10.1371/journal.pone.0262527}.
%Type = Article
\bibitem[{Kabir et~al.(2023)Kabir, Abdar, Khosravi, Jalali, Atiya, Nahavandi and Srinivasan}]{kabir2023spinalnet}
\bibinfo{author}{Kabir, H.M.D.}, \bibinfo{author}{Abdar, M.}, \bibinfo{author}{Khosravi, A.}, \bibinfo{author}{Jalali, S.M.J.}, \bibinfo{author}{Atiya, A.F.}, \bibinfo{author}{Nahavandi, S.}, \bibinfo{author}{Srinivasan, D.}, \bibinfo{year}{2023}.
\newblock \bibinfo{title}{Spinalnet: Deep neural network with gradual input}.
\newblock \bibinfo{journal}{IEEE Transactions on Artificial Intelligence} \bibinfo{volume}{4}, \bibinfo{pages}{1165--1177}.
\newblock \DOIprefix\doi{10.1109/TAI.2022.3185179}.
%Type = Article
\bibitem[{Kaur and Kumar(2024)}]{kaur2024facial}
\bibinfo{author}{Kaur, M.}, \bibinfo{author}{Kumar, M.}, \bibinfo{year}{2024}.
\newblock \bibinfo{title}{Facial emotion recognition: A comprehensive review}.
\newblock \bibinfo{journal}{Expert Systems} \bibinfo{volume}{41}, \bibinfo{pages}{e13670}.
\newblock \DOIprefix\doi{10.1111/exsy.13670}.
%Type = Inproceedings
\bibitem[{Khajontantichaikun et~al.(2023)Khajontantichaikun, Jaiyen, Yamsaengsung, Mongkolnam and Chirapornchai}]{khajontantichaikun2023facial}
\bibinfo{author}{Khajontantichaikun, T.}, \bibinfo{author}{Jaiyen, S.}, \bibinfo{author}{Yamsaengsung, S.}, \bibinfo{author}{Mongkolnam, P.}, \bibinfo{author}{Chirapornchai, T.}, \bibinfo{year}{2023}.
\newblock \bibinfo{title}{Facial emotion detection for thai elderly people using yolov7}, in: \bibinfo{booktitle}{2023 15th International Conference on Knowledge and Smart Technology (KST)}, pp. \bibinfo{pages}{1--4}.
\newblock \DOIprefix\doi{10.1109/KST57286.2023.10086786}.
%Type = Inproceedings
\bibitem[{Khajontantichaikun et~al.(2022)Khajontantichaikun, Jaiyen, Yamsaengsung, Mongkolnam and Ninrutsirikun}]{khajontantichaikun2022emotion}
\bibinfo{author}{Khajontantichaikun, T.}, \bibinfo{author}{Jaiyen, S.}, \bibinfo{author}{Yamsaengsung, S.}, \bibinfo{author}{Mongkolnam, P.}, \bibinfo{author}{Ninrutsirikun, U.}, \bibinfo{year}{2022}.
\newblock \bibinfo{title}{Emotion detection of thai elderly facial expressions using hybrid object detection}, in: \bibinfo{booktitle}{2022 26th International Computer Science and Engineering Conference (ICSEC)}, pp. \bibinfo{pages}{219--223}.
\newblock \DOIprefix\doi{10.1109/ICSEC56337.2022.10049334}.
%Type = Article
\bibitem[{Khan(2022)}]{rehmankhan2022facial}
\bibinfo{author}{Khan, A.R.}, \bibinfo{year}{2022}.
\newblock \bibinfo{title}{Facial emotion recognition using conventional machine learning and deep learning methods: Current achievements, analysis and remaining challenges}.
\newblock \bibinfo{journal}{Information} \bibinfo{volume}{13}.
\newblock \DOIprefix\doi{10.3390/info13060268}.
%Type = Article
\bibitem[{Khan et~al.(2019)Khan, Crenn, Meyer and Bouakaz}]{khan2019novel}
\bibinfo{author}{Khan, R.A.}, \bibinfo{author}{Crenn, A.}, \bibinfo{author}{Meyer, A.}, \bibinfo{author}{Bouakaz, S.}, \bibinfo{year}{2019}.
\newblock \bibinfo{title}{A novel database of children's spontaneous facial expressions (liris-cse)}.
\newblock \bibinfo{journal}{Image and Vision Computing} \bibinfo{volume}{83-84}, \bibinfo{pages}{61--69}.
\newblock \DOIprefix\doi{10.1016/j.imavis.2019.02.004}.
%Type = Inproceedings
\bibitem[{Kim et~al.(2021)Kim, Bryant, Srikanth and Howard}]{kim2021age}
\bibinfo{author}{Kim, E.}, \bibinfo{author}{Bryant, D.}, \bibinfo{author}{Srikanth, D.}, \bibinfo{author}{Howard, A.}, \bibinfo{year}{2021}.
\newblock \bibinfo{title}{Age bias in emotion detection: An analysis of facial emotion recognition performance on young, middle-aged, and older adults}, in: \bibinfo{booktitle}{Proceedings of the 2021 AAAI/ACM Conference on AI, Ethics, and Society}, \bibinfo{publisher}{Association for Computing Machinery}. pp. \bibinfo{pages}{638--644}.
\newblock \DOIprefix\doi{10.1145/3461702.3462609}.
%Type = Misc
\bibitem[{Kitchenham and Charters(2007)}]{kitchenham2007guidelines}
\bibinfo{author}{Kitchenham, B.}, \bibinfo{author}{Charters, S.}, \bibinfo{year}{2007}.
\newblock \bibinfo{title}{Guidelines for performing systematic literature reviews in software engineering}.
%Type = Article
\bibitem[{Ko(2018)}]{chulko2018brief}
\bibinfo{author}{Ko, B.C.}, \bibinfo{year}{2018}.
\newblock \bibinfo{title}{A brief review of facial emotion recognition based on visual information}.
\newblock \bibinfo{journal}{Sensors} \bibinfo{volume}{18}.
\newblock \DOIprefix\doi{10.3390/s18020401}.
%Type = Article
\bibitem[{Ko et~al.(2021)Ko, Kim, Bae, Seo, Nam, Park, Park, Ihm and Lee}]{ko2021changes}
\bibinfo{author}{Ko, H.}, \bibinfo{author}{Kim, K.}, \bibinfo{author}{Bae, M.}, \bibinfo{author}{Seo, M.G.}, \bibinfo{author}{Nam, G.}, \bibinfo{author}{Park, S.}, \bibinfo{author}{Park, S.}, \bibinfo{author}{Ihm, J.}, \bibinfo{author}{Lee, J.Y.}, \bibinfo{year}{2021}.
\newblock \bibinfo{title}{Changes in computer-analyzed facial expressions with age}.
\newblock \bibinfo{journal}{Sensors} \bibinfo{volume}{21}.
\newblock \DOIprefix\doi{10.3390/s21144858}.
%Type = Inproceedings
\bibitem[{Koch et~al.(2015)Koch, Zemel, Salakhutdinov et~al.}]{Siamese_CNNs}
\bibinfo{author}{Koch, G.}, \bibinfo{author}{Zemel, R.}, \bibinfo{author}{Salakhutdinov, R.}, et~al., \bibinfo{year}{2015}.
\newblock \bibinfo{title}{Siamese neural networks for one-shot image recognition}, in: \bibinfo{booktitle}{ICML deep learning workshop}, \bibinfo{organization}{Lille}. pp. \bibinfo{pages}{1--30}.
%Type = Article
\bibitem[{Krizhevsky et~al.(2017)Krizhevsky, Sutskever and Hinton}]{AlexNet}
\bibinfo{author}{Krizhevsky, A.}, \bibinfo{author}{Sutskever, I.}, \bibinfo{author}{Hinton, G.E.}, \bibinfo{year}{2017}.
\newblock \bibinfo{title}{Imagenet classification with deep convolutional neural networks}.
\newblock \bibinfo{journal}{Commun. ACM} \bibinfo{volume}{60}, \bibinfo{pages}{84–90}.
\newblock \DOIprefix\doi{10.1145/3065386}.
%Type = Inproceedings
\bibitem[{Kumari et~al.(2024)Kumari, Kapoor and Saini}]{kumari2024emotion}
\bibinfo{author}{Kumari, S.}, \bibinfo{author}{Kapoor, N.}, \bibinfo{author}{Saini, R.}, \bibinfo{year}{2024}.
\newblock \bibinfo{title}{Emotion recognition from facial expression using deep learning model : A review}, in: \bibinfo{booktitle}{2024 5th International Conference for Emerging Technology (INCET)}, pp. \bibinfo{pages}{1--6}.
\newblock \DOIprefix\doi{10.1109/INCET61516.2024.10592970}.
%Type = Inproceedings
\bibitem[{Kuprashevich and Tolstykh(2024)}]{kuprashevich2024mivolo}
\bibinfo{author}{Kuprashevich, M.}, \bibinfo{author}{Tolstykh, I.}, \bibinfo{year}{2024}.
\newblock \bibinfo{title}{Mivolo: Multi-input transformer for age and gender estimation}, in: \bibinfo{editor}{Ignatov, D.I.}, \bibinfo{editor}{Khachay, M.}, \bibinfo{editor}{Kutuzov, A.}, \bibinfo{editor}{Madoyan, H.}, \bibinfo{editor}{Makarov, I.}, \bibinfo{editor}{Nikishina, I.}, \bibinfo{editor}{Panchenko, A.}, \bibinfo{editor}{Panov, M.}, \bibinfo{editor}{Pardalos, P.M.}, \bibinfo{editor}{Savchenko, A.V.}, \bibinfo{editor}{Tsymbalov, E.}, \bibinfo{editor}{Tutubalina, E.}, \bibinfo{editor}{Zagoruyko, S.} (Eds.), \bibinfo{booktitle}{Analysis of Images, Social Networks and Texts}, \bibinfo{publisher}{Springer Nature Switzerland}, \bibinfo{address}{Cham}. pp. \bibinfo{pages}{212--226}.
%Type = Inproceedings
\bibitem[{Labzour et~al.(2023)Labzour, Fkihi, Benaissa, Zennayi and Bourja}]{labzour2023survey}
\bibinfo{author}{Labzour, N.}, \bibinfo{author}{Fkihi, S.E.}, \bibinfo{author}{Benaissa, S.}, \bibinfo{author}{Zennayi, Y.}, \bibinfo{author}{Bourja, O.}, \bibinfo{year}{2023}.
\newblock \bibinfo{title}{A survey on facial emotion recognition for the elderly}, in: \bibinfo{editor}{Motahhir, S.}, \bibinfo{editor}{Bossoufi, B.} (Eds.), \bibinfo{booktitle}{Digital Technologies and Applications}, \bibinfo{publisher}{Springer Nature Switzerland}. pp. \bibinfo{pages}{561--575}.
%Type = Inproceedings
\bibitem[{Langa(2018)}]{langa2018cognitive}
\bibinfo{author}{Langa, K.M.}, \bibinfo{year}{2018}.
\newblock \bibinfo{title}{Cognitive aging, dementia, and the future of an aging population}, in: \bibinfo{booktitle}{Future directions for the demography of aging: Proceedings of a workshop}, \bibinfo{organization}{National Academies Press Washington, DC, USA}. pp. \bibinfo{pages}{249--268}.
%Type = Article
\bibitem[{Langner et~al.(2010)Langner, Dotsch, Bijlstra, Wigboldus, Hawk and van Knippenberg}]{langner2010presentation}
\bibinfo{author}{Langner, O.}, \bibinfo{author}{Dotsch, R.}, \bibinfo{author}{Bijlstra, G.}, \bibinfo{author}{Wigboldus, D.H.}, \bibinfo{author}{Hawk, S.T.}, \bibinfo{author}{van Knippenberg, A.}, \bibinfo{year}{2010}.
\newblock \bibinfo{title}{Presentation and validation of the radboud faces database}.
\newblock \bibinfo{journal}{Cognition and Emotion} \bibinfo{volume}{24}, \bibinfo{pages}{1377--1388}.
\newblock \DOIprefix\doi{10.1080/02699930903485076}.
%Type = Inbook
\bibitem[{LeCun et~al.(1999)LeCun, Haffner, Bottou and Bengio}]{lecun1999object}
\bibinfo{author}{LeCun, Y.}, \bibinfo{author}{Haffner, P.}, \bibinfo{author}{Bottou, L.}, \bibinfo{author}{Bengio, Y.}, \bibinfo{year}{1999}.
\newblock \bibinfo{title}{Object Recognition with Gradient-Based Learning}. \bibinfo{publisher}{Springer Berlin Heidelberg}, \bibinfo{address}{Berlin, Heidelberg}.
\newblock pp. \bibinfo{pages}{319--345}.
\newblock \DOIprefix\doi{10.1007/3-540-46805-6\_19}.
%Type = Article
\bibitem[{Leong et~al.(2023)Leong, Tang, Lai and Lee}]{chitleong2023facial}
\bibinfo{author}{Leong, S.C.}, \bibinfo{author}{Tang, Y.M.}, \bibinfo{author}{Lai, C.H.}, \bibinfo{author}{Lee, C.K.M.}, \bibinfo{year}{2023}.
\newblock \bibinfo{title}{Facial expression and body gesture emotion recognition: A systematic review on the use of visual data in affective computing}.
\newblock \bibinfo{journal}{Computer Science Review} \bibinfo{volume}{48}, \bibinfo{pages}{100545}.
\newblock \DOIprefix\doi{10.1016/j.cosrev.2023.100545}.
%Type = Article
\bibitem[{Li and Deng(2022)}]{li2022deep}
\bibinfo{author}{Li, S.}, \bibinfo{author}{Deng, W.}, \bibinfo{year}{2022}.
\newblock \bibinfo{title}{Deep facial expression recognition: A survey}.
\newblock \bibinfo{journal}{IEEE Transactions on Affective Computing} \bibinfo{volume}{13}, \bibinfo{pages}{1195--1215}.
\newblock \DOIprefix\doi{10.1109/TAFFC.2020.2981446}.
%Type = Inproceedings
\bibitem[{Li et~al.(2015)Li, Li, Su and Zhu}]{WeiNet2015}
\bibinfo{author}{Li, W.}, \bibinfo{author}{Li, M.}, \bibinfo{author}{Su, Z.}, \bibinfo{author}{Zhu, Z.}, \bibinfo{year}{2015}.
\newblock \bibinfo{title}{A deep-learning approach to facial expression recognition with candid images}, in: \bibinfo{booktitle}{2015 14th IAPR International Conference on Machine Vision Applications (MVA)}, \bibinfo{organization}{IEEE}. pp. \bibinfo{pages}{279--282}.
%Type = Article
\bibitem[{Li et~al.(2018)Li, Tsangouri, Abtahi and Zhu}]{li2018recursive}
\bibinfo{author}{Li, W.}, \bibinfo{author}{Tsangouri, C.}, \bibinfo{author}{Abtahi, F.}, \bibinfo{author}{Zhu, Z.}, \bibinfo{year}{2018}.
\newblock \bibinfo{title}{A recursive framework for expression recognition: from web images to deep models to game dataset}.
\newblock \bibinfo{journal}{Machine Vision and Applications} \bibinfo{volume}{29}, \bibinfo{pages}{489--502}.
\newblock \DOIprefix\doi{10.1007/s00138-017-0904-9}.
%Type = Article
\bibitem[{Liang and Dong(2023)}]{liang2023survey}
\bibinfo{author}{Liang, C.}, \bibinfo{author}{Dong, J.}, \bibinfo{year}{2023}.
\newblock \bibinfo{title}{A survey of deep learning-based facial expression recognition research}.
\newblock \bibinfo{journal}{Frontiers in Computing and Intelligent Systems} \bibinfo{volume}{5}, \bibinfo{pages}{56--60}.
\newblock \DOIprefix\doi{10.54097/fcis.v5i2.12445}.
%Type = Inproceedings
\bibitem[{Liliana and Basaruddin(2018)}]{yantililiana2018review}
\bibinfo{author}{Liliana, D.Y.}, \bibinfo{author}{Basaruddin, T.}, \bibinfo{year}{2018}.
\newblock \bibinfo{title}{Review of automatic emotion recognition through facial expression analysis}, in: \bibinfo{booktitle}{2018 International Conference on Electrical Engineering and Computer Science (ICECOS)}, pp. \bibinfo{pages}{231--236}.
\newblock \DOIprefix\doi{10.1109/ICECOS.2018.8605222}.
%Type = Inproceedings
\bibitem[{Little and Briggs(2009)}]{little2009pervasive}
\bibinfo{author}{Little, L.}, \bibinfo{author}{Briggs, P.}, \bibinfo{year}{2009}.
\newblock \bibinfo{title}{Pervasive healthcare: the elderly perspective}, in: \bibinfo{booktitle}{Proceedings of the 2nd International Conference on PErvasive Technologies Related to Assistive Environments}, \bibinfo{publisher}{Association for Computing Machinery}, \bibinfo{address}{New York, NY, USA}.
\newblock \DOIprefix\doi{10.1145/1579114.1579185}.
%Type = Inproceedings
\bibitem[{Liu and Deng(2015)}]{VGG}
\bibinfo{author}{Liu, S.}, \bibinfo{author}{Deng, W.}, \bibinfo{year}{2015}.
\newblock \bibinfo{title}{Very deep convolutional neural network based image classification using small training sample size}, in: \bibinfo{booktitle}{2015 3rd IAPR Asian Conference on Pattern Recognition (ACPR)}, pp. \bibinfo{pages}{730--734}.
\newblock \DOIprefix\doi{10.1109/ACPR.2015.7486599}.
%Type = Inproceedings
\bibitem[{Liu et~al.(2016)Liu, Anguelov, Erhan, Szegedy, Reed, Fu and Berg}]{liu2016ssd}
\bibinfo{author}{Liu, W.}, \bibinfo{author}{Anguelov, D.}, \bibinfo{author}{Erhan, D.}, \bibinfo{author}{Szegedy, C.}, \bibinfo{author}{Reed, S.}, \bibinfo{author}{Fu, C.Y.}, \bibinfo{author}{Berg, A.C.}, \bibinfo{year}{2016}.
\newblock \bibinfo{title}{Ssd: Single shot multibox detector}, in: \bibinfo{editor}{Leibe, B.}, \bibinfo{editor}{Matas, J.}, \bibinfo{editor}{Sebe, N.}, \bibinfo{editor}{Welling, M.} (Eds.), \bibinfo{booktitle}{Computer Vision -- ECCV 2016}, \bibinfo{publisher}{Springer International Publishing}, \bibinfo{address}{Cham}. pp. \bibinfo{pages}{21--37}.
%Type = Inproceedings
\bibitem[{Liu et~al.(2022)Liu, Ning, Cao, Wei, Zhang, Lin and Hu}]{liu2022video}
\bibinfo{author}{Liu, Z.}, \bibinfo{author}{Ning, J.}, \bibinfo{author}{Cao, Y.}, \bibinfo{author}{Wei, Y.}, \bibinfo{author}{Zhang, Z.}, \bibinfo{author}{Lin, S.}, \bibinfo{author}{Hu, H.}, \bibinfo{year}{2022}.
\newblock \bibinfo{title}{Video swin transformer}, in: \bibinfo{booktitle}{2022 IEEE/CVF Conference on Computer Vision and Pattern Recognition (CVPR)}, pp. \bibinfo{pages}{3192--3201}.
\newblock \DOIprefix\doi{10.1109/CVPR52688.2022.00320}.
%Type = Article
\bibitem[{Livingstone and Russo(2018)}]{livingstone2018ryerson}
\bibinfo{author}{Livingstone, S.R.}, \bibinfo{author}{Russo, F.A.}, \bibinfo{year}{2018}.
\newblock \bibinfo{title}{The ryerson audio-visual database of emotional speech and song (ravdess): A dynamic, multimodal set of facial and vocal expressions in north american english}.
\newblock \bibinfo{journal}{PLOS ONE} \bibinfo{volume}{13}, \bibinfo{pages}{1--35}.
\newblock \DOIprefix\doi{10.1371/journal.pone.0196391}.
%Type = Article
\bibitem[{LoBue and Thrasher(2014)}]{lobue2014child}
\bibinfo{author}{LoBue, V.}, \bibinfo{author}{Thrasher, C.}, \bibinfo{year}{2014}.
\newblock \bibinfo{title}{The child affective facial expression (cafe) set: Validity and reliability from untrained adults}.
\newblock \bibinfo{journal}{Frontiers in Psychology} \bibinfo{volume}{5}.
\newblock \DOIprefix\doi{10.3389/fpsyg.2014.01532}.
%Type = Inproceedings
\bibitem[{Lopes et~al.(2018)Lopes, Silva, Khanal, Reis, Barroso, Filipe and Sampaio}]{lopes2018facial}
\bibinfo{author}{Lopes, N.}, \bibinfo{author}{Silva, A.}, \bibinfo{author}{Khanal, S.R.}, \bibinfo{author}{Reis, A.}, \bibinfo{author}{Barroso, J.}, \bibinfo{author}{Filipe, V.}, \bibinfo{author}{Sampaio, J.}, \bibinfo{year}{2018}.
\newblock \bibinfo{title}{Facial emotion recognition in the elderly using a svm classifier}, in: \bibinfo{booktitle}{2018 2nd International Conference on Technology and Innovation in Sports, Health and Wellbeing (TISHW)}, pp. \bibinfo{pages}{1--5}.
\newblock \DOIprefix\doi{10.1109/TISHW.2018.8559494}.
%Type = Misc
\bibitem[{Lucey et~al.(2010)Lucey, Cohn, Kanade, Saragih, Ambadar and Matthews}]{lucey2010extended}
\bibinfo{author}{Lucey, P.}, \bibinfo{author}{Cohn, J.F.}, \bibinfo{author}{Kanade, T.}, \bibinfo{author}{Saragih, J.}, \bibinfo{author}{Ambadar, Z.}, \bibinfo{author}{Matthews, I.}, \bibinfo{year}{2010}.
\newblock \bibinfo{title}{The extended cohn-kanade dataset (ck+): A complete dataset for action unit and emotion-specified expression}.
%Type = Inproceedings
\bibitem[{Lundberg and Lee(2017)}]{lundberg2017unified}
\bibinfo{author}{Lundberg, S.M.}, \bibinfo{author}{Lee, S.I.}, \bibinfo{year}{2017}.
\newblock \bibinfo{title}{A unified approach to interpreting model predictions}, in: \bibinfo{booktitle}{Proceedings of the 31st International Conference on Neural Information Processing Systems}, \bibinfo{publisher}{Curran Associates Inc.}, \bibinfo{address}{Red Hook, NY, USA}. p. \bibinfo{pages}{4768–4777}.
%Type = Article
\bibitem[{Lundqvist et~al.(1998)Lundqvist, Flykt and {\"O}hman}]{lundqvist1998karolinska}
\bibinfo{author}{Lundqvist, D.}, \bibinfo{author}{Flykt, A.}, \bibinfo{author}{{\"O}hman, A.}, \bibinfo{year}{1998}.
\newblock \bibinfo{title}{Karolinska directed emotional faces}.
\newblock \bibinfo{journal}{PsycTESTS Dataset} \bibinfo{volume}{91}, \bibinfo{pages}{630}.
\newblock \DOIprefix\doi{10.1037/t27732-000}.
%Type = Misc
\bibitem[{Lyons et~al.(2020a)Lyons, Kamachi and Gyoba}]{lyons2020coding}
\bibinfo{author}{Lyons, M.J.}, \bibinfo{author}{Kamachi, M.}, \bibinfo{author}{Gyoba, J.}, \bibinfo{year}{2020}a.
\newblock \bibinfo{title}{Coding facial expressions with gabor wavelets (ivc special issue)}.
\newblock \DOIprefix\doi{10.5281/ZENODO.4029679}.
%Type = Misc
\bibitem[{Lyons et~al.(2020b)Lyons, Kamachi and Gyoba}]{jlyons2020coding}
\bibinfo{author}{Lyons, M.J.}, \bibinfo{author}{Kamachi, M.}, \bibinfo{author}{Gyoba, J.}, \bibinfo{year}{2020}b.
\newblock \bibinfo{title}{Coding facial expressions with gabor wavelets (ivc special issue)}.
\newblock \DOIprefix\doi{10.5281/zenodo.4029679}.
%Type = Inproceedings
\bibitem[{Ma et~al.(2019)Ma, Wang, Yang, Zhang, Girard and Morency}]{ma2019elderreact}
\bibinfo{author}{Ma, K.}, \bibinfo{author}{Wang, X.}, \bibinfo{author}{Yang, X.}, \bibinfo{author}{Zhang, M.}, \bibinfo{author}{Girard, J.M.}, \bibinfo{author}{Morency, L.P.}, \bibinfo{year}{2019}.
\newblock \bibinfo{title}{Elderreact: A multimodal dataset for recognizing emotional response in aging adults}, in: \bibinfo{booktitle}{ICMI 2019 - Proceedings of the 2019 International Conference on Multimodal Interaction}, \bibinfo{publisher}{Association for Computing Machinery, Inc}. pp. \bibinfo{pages}{349--357}.
\newblock \DOIprefix\doi{10.1145/3340555.3353747}.
%Type = Misc
\bibitem[{Maithri et~al.(2022)Maithri, Raghavendra, Gudigar, Samanth, Barua, Murugappan, Chakole and Acharya}]{maithri2022automated}
\bibinfo{author}{Maithri, M.}, \bibinfo{author}{Raghavendra, U.}, \bibinfo{author}{Gudigar, A.}, \bibinfo{author}{Samanth, J.}, \bibinfo{author}{Barua, P.D.}, \bibinfo{author}{Murugappan, M.}, \bibinfo{author}{Chakole, Y.}, \bibinfo{author}{Acharya, U.R.}, \bibinfo{year}{2022}.
\newblock \bibinfo{title}{Automated emotion recognition: Current trends and future perspectives}.
\newblock \DOIprefix\doi{10.1016/j.cmpb.2022.106646}.
%Type = Article
\bibitem[{Manresa-Yee et~al.(2023)Manresa-Yee, Ramis and Buades}]{ijimai}
\bibinfo{author}{Manresa-Yee, C.}, \bibinfo{author}{Ramis, S.}, \bibinfo{author}{Buades, J.M.}, \bibinfo{year}{2023}.
\newblock \bibinfo{title}{Analysis of gender differences in facial expression recognition based on deep learning using explainable artificial intelligence}.
\newblock \bibinfo{journal}{International Journal of Interactive Multimedia and Artificial Intelligence} \bibinfo{volume}{In Press}, \bibinfo{pages}{1--10}.
\newblock \DOIprefix\doi{10.9781/ijimai.2023.04.003}.
%Type = Article
\bibitem[{Martinez et~al.(2019)Martinez, Valstar, Jiang and Pantic}]{martinez2019automatic}
\bibinfo{author}{Martinez, B.}, \bibinfo{author}{Valstar, M.F.}, \bibinfo{author}{Jiang, B.}, \bibinfo{author}{Pantic, M.}, \bibinfo{year}{2019}.
\newblock \bibinfo{title}{Automatic analysis of facial actions: A survey}.
\newblock \bibinfo{journal}{IEEE Transactions on Affective Computing} \bibinfo{volume}{10}, \bibinfo{pages}{325--347}.
\newblock \DOIprefix\doi{10.1109/TAFFC.2017.2731763}.
%Type = Inproceedings
\bibitem[{Mary and Jayakumar(2016)}]{mary2016review}
\bibinfo{author}{Mary, R.}, \bibinfo{author}{Jayakumar, T.V.}, \bibinfo{year}{2016}.
\newblock \bibinfo{title}{A review on how human aging influences facial expression recognition (fer)}, in: \bibinfo{booktitle}{Advances in Intelligent Systems and Computing}, \bibinfo{publisher}{Springer Verlag}. pp. \bibinfo{pages}{313--322}.
\newblock \DOIprefix\doi{10.1007/978-3-319-28031-8\_27}.
%Type = Misc
\bibitem[{{Megvii Technology}(2012)}]{Face++}
\bibinfo{author}{{Megvii Technology}}, \bibinfo{year}{2012}.
\newblock \bibinfo{title}{Face++}.
\newblock \URLprefix \url{https://www.faceplusplus.com/emotion-recognition/}. \bibinfo{note}{{Accessed 22nd Nov 2024}}.
%Type = Article
\bibitem[{Mejia-Escobar et~al.(2023)Mejia-Escobar, Cazorla and Martinez-Martin}]{mejia-escobar2023towards}
\bibinfo{author}{Mejia-Escobar, C.}, \bibinfo{author}{Cazorla, M.}, \bibinfo{author}{Martinez-Martin, E.}, \bibinfo{year}{2023}.
\newblock \bibinfo{title}{Towards a better performance in facial expression recognition: A data-centric approach}.
\newblock \bibinfo{journal}{Computational Intelligence and Neuroscience} \bibinfo{volume}{2023}, \bibinfo{pages}{1394882}.
\newblock \DOIprefix\doi{10.1155/2023/1394882}.
%Type = Article
\bibitem[{Meuwissen et~al.(2017)Meuwissen, Anderson and Zelazo}]{meuwissen2017creation}
\bibinfo{author}{Meuwissen, A.S.}, \bibinfo{author}{Anderson, J.E.}, \bibinfo{author}{Zelazo, P.D.}, \bibinfo{year}{2017}.
\newblock \bibinfo{title}{The creation and validation of the developmental emotional faces stimulus set}.
\newblock \bibinfo{journal}{Behavior Research Methods} \bibinfo{volume}{49}, \bibinfo{pages}{960--966}.
\newblock \DOIprefix\doi{10.3758/s13428-016-0756-7}.
%Type = Misc
\bibitem[{{Microsoft Azure}(2017)}]{MicrosoftFace}
\bibinfo{author}{{Microsoft Azure}}, \bibinfo{year}{2017}.
\newblock \bibinfo{title}{Microsoft face service}.
\newblock \URLprefix \url{https://azure.microsoft.com/en-us/products/ai-services/ai-vision}. \bibinfo{note}{{Accessed 22nd Nov 2024}}.
%Type = Inproceedings
\bibitem[{Milborrow and Nicolls(2014)}]{milborrow2014active}
\bibinfo{author}{Milborrow, S.}, \bibinfo{author}{Nicolls, F.}, \bibinfo{year}{2014}.
\newblock \bibinfo{title}{Active shape models with sift descriptors and mars}, in: \bibinfo{booktitle}{2014 International Conference on Computer Vision Theory and Applications (VISAPP)}, pp. \bibinfo{pages}{380--387}.
%Type = Article
\bibitem[{Minear and Park(2004)}]{minear2004lifespan}
\bibinfo{author}{Minear, M.}, \bibinfo{author}{Park, D.C.}, \bibinfo{year}{2004}.
\newblock \bibinfo{title}{A lifespan database of adult facial stimuli}.
\newblock \bibinfo{journal}{Behavior Research Methods, Instruments, \& Computers} \bibinfo{volume}{36}, \bibinfo{pages}{630--633}.
\newblock \DOIprefix\doi{10.3758/BF03206543}.
%Type = Article
\bibitem[{Modi and Patel(2022)}]{modi2022state-of-the-art}
\bibinfo{author}{Modi, P.}, \bibinfo{author}{Patel, S.}, \bibinfo{year}{2022}.
\newblock \bibinfo{title}{A state-of-the-art survey on face recognition methods}.
\newblock \bibinfo{journal}{International Journal of Computer Vision and Image Processing (IJCVIP)} \bibinfo{volume}{12}, \bibinfo{pages}{1--19}.
\newblock \DOIprefix\doi{10.4018/IJCVIP.2022010101}.
%Type = Inproceedings
\bibitem[{Mohammed et~al.(2020)Mohammed, Rawashdeh and Abdullah}]{mohammed2020machine}
\bibinfo{author}{Mohammed, R.}, \bibinfo{author}{Rawashdeh, J.}, \bibinfo{author}{Abdullah, M.}, \bibinfo{year}{2020}.
\newblock \bibinfo{title}{Machine learning with oversampling and undersampling techniques: Overview study and experimental results}, in: \bibinfo{booktitle}{2020 11th International Conference on Information and Communication Systems (ICICS)}, pp. \bibinfo{pages}{243--248}.
\newblock \DOIprefix\doi{10.1109/ICICS49469.2020.239556}.
%Type = Article
\bibitem[{Mohana and Subashini(2024)}]{mohana2024facial}
\bibinfo{author}{Mohana, M.}, \bibinfo{author}{Subashini, P.}, \bibinfo{year}{2024}.
\newblock \bibinfo{title}{Facial expression recognition using machine learning and deep learning techniques: A systematic review}.
\newblock \bibinfo{journal}{SN Computer Science} \bibinfo{volume}{5}, \bibinfo{pages}{432}.
\newblock \DOIprefix\doi{10.1007/s42979-024-02792-7}.
%Type = Article
\bibitem[{Mollahosseini et~al.(2019)Mollahosseini, Hasani and Mahoor}]{mollahosseini2017affectnet}
\bibinfo{author}{Mollahosseini, A.}, \bibinfo{author}{Hasani, B.}, \bibinfo{author}{Mahoor, M.H.}, \bibinfo{year}{2019}.
\newblock \bibinfo{title}{Affectnet: A database for facial expression, valence, and arousal computing in the wild}.
\newblock \bibinfo{journal}{IEEE Transactions on Affective Computing} \bibinfo{volume}{10}, \bibinfo{pages}{18--31}.
\newblock \DOIprefix\doi{10.1109/TAFFC.2017.2740923}.
%Type = Inproceedings
\bibitem[{Moolchandani et~al.(2021)Moolchandani, Dwivedi, Nigam and Gupta}]{moolchandani2021survey}
\bibinfo{author}{Moolchandani, M.}, \bibinfo{author}{Dwivedi, S.}, \bibinfo{author}{Nigam, S.}, \bibinfo{author}{Gupta, K.}, \bibinfo{year}{2021}.
\newblock \bibinfo{title}{A survey on: Facial emotion recognition and classification}, in: \bibinfo{booktitle}{2021 5th International Conference on Computing Methodologies and Communication (ICCMC)}, pp. \bibinfo{pages}{1677--1686}.
\newblock \DOIprefix\doi{10.1109/ICCMC51019.2021.9418349}.
%Type = Article
\bibitem[{Muazu et~al.(2021)Muazu, Manzi and Aminu}]{muazu2021systematic}
\bibinfo{author}{Muazu, A.}, \bibinfo{author}{Manzi, E.}, \bibinfo{author}{Aminu, J.}, \bibinfo{year}{2021}.
\newblock \bibinfo{title}{A systematic review of methods of emotion recognition by facial expressions}.
\newblock \bibinfo{journal}{International Journal of Advanced Research} \bibinfo{volume}{9}, \bibinfo{pages}{1141--1152}.
\newblock \DOIprefix\doi{10.21474/IJAR01/12951}.
%Type = Inproceedings
\bibitem[{Nojavanasghari et~al.(2016)Nojavanasghari, Baltru\v{s}aitis, Hughes and Morency}]{nojavanasghari2016emoreact}
\bibinfo{author}{Nojavanasghari, B.}, \bibinfo{author}{Baltru\v{s}aitis, T.}, \bibinfo{author}{Hughes, C.E.}, \bibinfo{author}{Morency, L.P.}, \bibinfo{year}{2016}.
\newblock \bibinfo{title}{Emoreact: a multimodal approach and dataset for recognizing emotional responses in children}, in: \bibinfo{booktitle}{Proceedings of the 18th ACM International Conference on Multimodal Interaction}, \bibinfo{publisher}{Association for Computing Machinery}, \bibinfo{address}{New York, NY, USA}. p. \bibinfo{pages}{137–144}.
\newblock \DOIprefix\doi{10.1145/2993148.2993168}.
%Type = Misc
\bibitem[{{Noldus Information Technology b.v}(2021)}]{noldus}
\bibinfo{author}{{Noldus Information Technology b.v}}, \bibinfo{year}{2021}.
\newblock \bibinfo{title}{Facial expression recognition software | {FaceReader}}.
\newblock \URLprefix \url{https://www.noldus.com/facereader}. \bibinfo{note}{{Accessed 22nd Nov 2024}}.
%Type = Article
\bibitem[{Oliver and Amengual~Alcover(2020)}]{oliver2020uibvfed}
\bibinfo{author}{Oliver, M.M.}, \bibinfo{author}{Amengual~Alcover, E.}, \bibinfo{year}{2020}.
\newblock \bibinfo{title}{Uibvfed: Virtual facial expression dataset}.
\newblock \bibinfo{journal}{PLOS ONE} \bibinfo{volume}{15}, \bibinfo{pages}{1--10}.
\newblock \DOIprefix\doi{10.1371/journal.pone.0231266}.
%Type = Inproceedings
\bibitem[{O'Mahony et~al.(2020)O'Mahony, Campbell, Carvalho, Harapanahalli, Hernandez, Krpalkova, Riordan and Walsh}]{omahony2020deep}
\bibinfo{author}{O'Mahony, N.}, \bibinfo{author}{Campbell, S.}, \bibinfo{author}{Carvalho, A.}, \bibinfo{author}{Harapanahalli, S.}, \bibinfo{author}{Hernandez, G.V.}, \bibinfo{author}{Krpalkova, L.}, \bibinfo{author}{Riordan, D.}, \bibinfo{author}{Walsh, J.}, \bibinfo{year}{2020}.
\newblock \bibinfo{title}{Deep learning vs. traditional computer vision}, in: \bibinfo{editor}{Arai, K.}, \bibinfo{editor}{Kapoor, S.} (Eds.), \bibinfo{booktitle}{Advances in Computer Vision}, \bibinfo{publisher}{Springer International Publishing}, \bibinfo{address}{Cham}. pp. \bibinfo{pages}{128--144}.
\newblock \DOIprefix\doi{10.1007/978-3-030-17795-9\_10}.
%Type = Misc
\bibitem[{OpenAI et~al.(2024)OpenAI, Achiam, Adler and et~al.}]{openai2024gpt-4}
\bibinfo{author}{OpenAI}, \bibinfo{author}{Achiam, J.}, \bibinfo{author}{Adler, S.}, \bibinfo{author}{et~al.}, \bibinfo{year}{2024}.
\newblock \bibinfo{title}{Gpt-4 technical report}.
\newblock \DOIprefix\doi{10.48550/arXiv.2303.08774}.
%Type = Inproceedings
\bibitem[{Pantic et~al.(2005)Pantic, Valstar, Rademaker and Maat}]{pantic2005web-based}
\bibinfo{author}{Pantic, M.}, \bibinfo{author}{Valstar, M.}, \bibinfo{author}{Rademaker, R.}, \bibinfo{author}{Maat, L.}, \bibinfo{year}{2005}.
\newblock \bibinfo{title}{Web-based database for facial expression analysis}, in: \bibinfo{booktitle}{2005 IEEE International Conference on Multimedia and Expo}, pp. \bibinfo{pages}{5 pp.--}.
\newblock \DOIprefix\doi{10.1109/ICME.2005.1521424}.
%Type = Article
\bibitem[{Park et~al.(2022)Park, Shin, Song, Yun and Jang}]{park2022facial}
\bibinfo{author}{Park, H.}, \bibinfo{author}{Shin, Y.}, \bibinfo{author}{Song, K.}, \bibinfo{author}{Yun, C.}, \bibinfo{author}{Jang, D.}, \bibinfo{year}{2022}.
\newblock \bibinfo{title}{Facial emotion recognition analysis based on age-biased data}.
\newblock \bibinfo{journal}{Applied Sciences} \bibinfo{volume}{12}.
\newblock \DOIprefix\doi{10.3390/app12167992}.
%Type = Article
\bibitem[{Patel et~al.(2020)Patel, Mehta, Mistry, Gupta, Tanwar, Kumar and Alazab}]{patel2020facial}
\bibinfo{author}{Patel, K.}, \bibinfo{author}{Mehta, D.}, \bibinfo{author}{Mistry, C.}, \bibinfo{author}{Gupta, R.}, \bibinfo{author}{Tanwar, S.}, \bibinfo{author}{Kumar, N.}, \bibinfo{author}{Alazab, M.}, \bibinfo{year}{2020}.
\newblock \bibinfo{title}{Facial sentiment analysis using ai techniques: State-of-the-art, taxonomies, and challenges}.
\newblock \bibinfo{journal}{IEEE Access} \bibinfo{volume}{8}, \bibinfo{pages}{90495--90519}.
\newblock \DOIprefix\doi{10.1109/ACCESS.2020.2993803}.
%Type = Book
\bibitem[{Pearl(1988)}]{pearl1988probabilistic}
\bibinfo{author}{Pearl, J.}, \bibinfo{year}{1988}.
\newblock \bibinfo{title}{Probabilistic Reasoning in Intelligent Systems: Networks of Plausible Inference}.
\newblock \bibinfo{publisher}{Morgan Kaufmann Publishers Inc.}, \bibinfo{address}{San Francisco, CA, USA}.
%Type = Inproceedings
\bibitem[{Petrou et~al.(2023)Petrou, Christodoulou, Avgerinakis and Kosmides}]{petrou2023lightweight}
\bibinfo{author}{Petrou, N.}, \bibinfo{author}{Christodoulou, G.}, \bibinfo{author}{Avgerinakis, K.}, \bibinfo{author}{Kosmides, P.}, \bibinfo{year}{2023}.
\newblock \bibinfo{title}{Lightweight mood estimation algorithm for faces under partial occlusion}, in: \bibinfo{booktitle}{Proceedings of the 16th International Conference on PErvasive Technologies Related to Assistive Environments}, \bibinfo{publisher}{Association for Computing Machinery}. pp. \bibinfo{pages}{402--407}.
\newblock \DOIprefix\doi{10.1145/3594806.3596553}.
%Type = Misc
\bibitem[{Petsiuk et~al.(2018)Petsiuk, Das and Saenko}]{petsiuk2018rise}
\bibinfo{author}{Petsiuk, V.}, \bibinfo{author}{Das, A.}, \bibinfo{author}{Saenko, K.}, \bibinfo{year}{2018}.
\newblock \bibinfo{title}{Rise: Randomized input sampling for explanation of black-box models}.
\newblock \DOIprefix\doi{10.48550/arXiv.1806.07421}.
%Type = Inproceedings
\bibitem[{Picón et~al.(2022)Picón, Roig-Maimó, Oliver, Alcover and Mas-Sansó}]{carretopicón2022do}
\bibinfo{author}{Picón, G.C.}, \bibinfo{author}{Roig-Maimó, M.F.}, \bibinfo{author}{Oliver, M.M.}, \bibinfo{author}{Alcover, E.A.}, \bibinfo{author}{Mas-Sansó, R.}, \bibinfo{year}{2022}.
\newblock \bibinfo{title}{Do machines better understand synthetic facial expressions than people?}, in: \bibinfo{booktitle}{Proceedings of the XXII International Conference on Human Computer Interaction}, \bibinfo{publisher}{Association for Computing Machinery}.
\newblock \DOIprefix\doi{10.1145/3549865.3549908}.
%Type = Article
\bibitem[{Pinto et~al.(2023)Pinto, Alves, Medeiros, da~Silva~Costa, Pires, Costa and da~Rocha~Seruffo}]{vinicioslopespinto2023systematic}
\bibinfo{author}{Pinto, L.V.L.}, \bibinfo{author}{Alves, A.V.N.}, \bibinfo{author}{Medeiros, A.M.}, \bibinfo{author}{da~Silva~Costa, S.W.}, \bibinfo{author}{Pires, Y.P.}, \bibinfo{author}{Costa, F.A.R.}, \bibinfo{author}{da~Rocha~Seruffo, M.C.}, \bibinfo{year}{2023}.
\newblock \bibinfo{title}{A systematic review of facial expression detection methods}.
\newblock \bibinfo{journal}{IEEE Access} \bibinfo{volume}{11}, \bibinfo{pages}{61881--61891}.
\newblock \DOIprefix\doi{10.1109/ACCESS.2023.3287090}.
%Type = Misc
\bibitem[{Prados-Torreblanca et~al.(2022)Prados-Torreblanca, Buenaposada and Baumela}]{prados-torreblanca2022shape}
\bibinfo{author}{Prados-Torreblanca, A.}, \bibinfo{author}{Buenaposada, J.M.}, \bibinfo{author}{Baumela, L.}, \bibinfo{year}{2022}.
\newblock \bibinfo{title}{Shape preserving facial landmarks with graph attention networks}.
\newblock \DOIprefix\doi{10.48550/arXiv.2210.07233}.
%Type = Inproceedings
\bibitem[{Pranathi et~al.(2021)Pranathi, Lakshmi and Suneetha}]{pranathi2021review}
\bibinfo{author}{Pranathi, P.}, \bibinfo{author}{Lakshmi, C.}, \bibinfo{author}{Suneetha, M.}, \bibinfo{year}{2021}.
\newblock \bibinfo{title}{A review on various facial expression recognition techniques}, in: \bibinfo{booktitle}{2021 Fifth International Conference on I-SMAC (IoT in Social, Mobile, Analytics and Cloud) (I-SMAC)}, pp. \bibinfo{pages}{1246--1254}.
\newblock \DOIprefix\doi{10.1109/I-SMAC52330.2021.9640733}.
%Type = Article
\bibitem[{Qin et~al.(2020)Qin, Zhang, Huang, Dehghan, Zaiane and Jagersand}]{qin2020u2-net}
\bibinfo{author}{Qin, X.}, \bibinfo{author}{Zhang, Z.}, \bibinfo{author}{Huang, C.}, \bibinfo{author}{Dehghan, M.}, \bibinfo{author}{Zaiane, O.R.}, \bibinfo{author}{Jagersand, M.}, \bibinfo{year}{2020}.
\newblock \bibinfo{title}{U2-net: Going deeper with nested u-structure for salient object detection}.
\newblock \bibinfo{journal}{Pattern Recognition} \bibinfo{volume}{106}, \bibinfo{pages}{107404}.
\newblock \DOIprefix\doi{10.1016/j.patcog.2020.107404}.
%Type = Inproceedings
\bibitem[{Rajeswari and IthayaRani(2018)}]{rajeswari2018literature}
\bibinfo{author}{Rajeswari, G.}, \bibinfo{author}{IthayaRani, P.}, \bibinfo{year}{2018}.
\newblock \bibinfo{title}{Literature survey on facial expression recognition techniques}, in: \bibinfo{booktitle}{2018 3rd International Conference on Communication and Electronics Systems (ICCES)}, pp. \bibinfo{pages}{137--142}.
\newblock \DOIprefix\doi{10.1109/CESYS.2018.8723953}.
%Type = Article
\bibitem[{Ramis et~al.(2022)Ramis, Buades, Perales and Manresa-Yee}]{SilNet2022}
\bibinfo{author}{Ramis, S.}, \bibinfo{author}{Buades, J.M.}, \bibinfo{author}{Perales, F.J.}, \bibinfo{author}{Manresa-Yee, C.}, \bibinfo{year}{2022}.
\newblock \bibinfo{title}{A novel approach to cross dataset studies in facial expression recognition}.
\newblock \bibinfo{journal}{Multimedia Tools Appl.} \bibinfo{volume}{81}, \bibinfo{pages}{39507–39544}.
\newblock \DOIprefix\doi{10.1007/s11042-022-13117-2}.
%Type = Inproceedings
\bibitem[{Redmon et~al.(2016)Redmon, Divvala, Girshick and Farhadi}]{YOLO}
\bibinfo{author}{Redmon, J.}, \bibinfo{author}{Divvala, S.}, \bibinfo{author}{Girshick, R.}, \bibinfo{author}{Farhadi, A.}, \bibinfo{year}{2016}.
\newblock \bibinfo{title}{You only look once: Unified, real-time object detection}, in: \bibinfo{booktitle}{2016 IEEE Conference on Computer Vision and Pattern Recognition (CVPR)}, pp. \bibinfo{pages}{779--788}.
\newblock \DOIprefix\doi{10.1109/CVPR.2016.91}.
%Type = Article
\bibitem[{Ren et~al.(2017)Ren, He, Girshick and Sun}]{ren2017faster}
\bibinfo{author}{Ren, S.}, \bibinfo{author}{He, K.}, \bibinfo{author}{Girshick, R.}, \bibinfo{author}{Sun, J.}, \bibinfo{year}{2017}.
\newblock \bibinfo{title}{Faster r-cnn: Towards real-time object detection with region proposal networks}.
\newblock \bibinfo{journal}{IEEE Transactions on Pattern Analysis and Machine Intelligence} \bibinfo{volume}{39}, \bibinfo{pages}{1137--1149}.
\newblock \DOIprefix\doi{10.1109/TPAMI.2016.2577031}.
%Type = Article
\bibitem[{Revina and Emmanuel(2021)}]{revina2021survey}
\bibinfo{author}{Revina, I.}, \bibinfo{author}{Emmanuel, W.R.S.}, \bibinfo{year}{2021}.
\newblock \bibinfo{title}{A survey on human face expression recognition techniques}.
\newblock \bibinfo{journal}{Journal of King Saud University - Computer and Information Sciences} \bibinfo{volume}{33}, \bibinfo{pages}{619--628}.
\newblock \DOIprefix\doi{10.1016/j.jksuci.2018.09.002}.
%Type = Inproceedings
\bibitem[{Ribeiro et~al.(2016)Ribeiro, Singh and Guestrin}]{ribeiro2016why}
\bibinfo{author}{Ribeiro, M.T.}, \bibinfo{author}{Singh, S.}, \bibinfo{author}{Guestrin, C.}, \bibinfo{year}{2016}.
\newblock \bibinfo{title}{``{Why} should {I} trust you?'': Explaining the predictions of any classifier}, in: \bibinfo{booktitle}{Proceedings of the 22nd ACM SIGKDD International Conference on Knowledge Discovery and Data Mining}, \bibinfo{publisher}{ACM}, \bibinfo{address}{New York}. p. \bibinfo{pages}{1135–1144}.
\newblock \DOIprefix\doi{10.1145/2939672.2939778}.
%Type = Inproceedings
\bibitem[{Rokkones et~al.(2019)Rokkones, Uddin and Torresen}]{skibelirokkones2019facial}
\bibinfo{author}{Rokkones, A.S.}, \bibinfo{author}{Uddin, M.Z.}, \bibinfo{author}{Torresen, J.}, \bibinfo{year}{2019}.
\newblock \bibinfo{title}{Facial expression recognition using robust local directional strength pattern features and recurrent neural network}, in: \bibinfo{editor}{Velikic, G.}, \bibinfo{editor}{Gross, C.} (Eds.), \bibinfo{booktitle}{2019 IEEE 9TH International Conference on Consumer Electronics (ICCE-BERLIN)}, \bibinfo{publisher}{IEEE}. pp. \bibinfo{pages}{283--288}.
\newblock \DOIprefix\doi{10.1109/icce-berlin47944.2019.8966234}.
%Type = Inproceedings
\bibitem[{Ruan et~al.(2024)Ruan, Wei, Guo, Xie and Yuan}]{ruan2024scatnet}
\bibinfo{author}{Ruan, Z.}, \bibinfo{author}{Wei, Y.}, \bibinfo{author}{Guo, Y.}, \bibinfo{author}{Xie, Y.}, \bibinfo{author}{Yuan, Y.}, \bibinfo{year}{2024}.
\newblock \bibinfo{title}{Scatnet: A novel self-supervised contrastive framework with spatial-channel attention and temporal transformer for few-shot action recognition}, in: \bibinfo{booktitle}{Proceedings of the 2023 6th International Conference on Algorithms, Computing and Artificial Intelligence}, \bibinfo{publisher}{Association for Computing Machinery}, \bibinfo{address}{New York, NY, USA}. p. \bibinfo{pages}{135–140}.
\newblock \DOIprefix\doi{10.1145/3639631.3639654}.
%Type = Article
\bibitem[{Ruffman et~al.(2008)Ruffman, Henry, Livingstone and Phillips}]{ruffman2008meta}
\bibinfo{author}{Ruffman, T.}, \bibinfo{author}{Henry, J.D.}, \bibinfo{author}{Livingstone, V.}, \bibinfo{author}{Phillips, L.H.}, \bibinfo{year}{2008}.
\newblock \bibinfo{title}{A meta-analytic review of emotion recognition and aging: Implications for neuropsychological models of aging}.
\newblock \bibinfo{journal}{Neuroscience \& Biobehavioral Reviews} \bibinfo{volume}{32}, \bibinfo{pages}{863--881}.
\newblock \DOIprefix\doi{10.1016/j.neubiorev.2008.01.001}.
%Type = Article
\bibitem[{Saez-Pons et~al.(2015)Saez-Pons, Syrdal and Dautenhahn}]{joan2015what}
\bibinfo{author}{Saez-Pons, J.}, \bibinfo{author}{Syrdal, D.S.}, \bibinfo{author}{Dautenhahn, K.}, \bibinfo{year}{2015}.
\newblock \bibinfo{title}{What has happened today? memory visualisation of a robot companion to assist user’s memory}.
\newblock \bibinfo{journal}{Journal of Assistive Technologies} \bibinfo{volume}{9}, \bibinfo{pages}{207--218}.
\newblock \DOIprefix\doi{10.1108/JAT-02-2015-0004}.
%Type = Inproceedings
\bibitem[{Sandler et~al.(2018)Sandler, Howard, Zhu, Zhmoginov and Chen}]{sandler2018mobilenetv2}
\bibinfo{author}{Sandler, M.}, \bibinfo{author}{Howard, A.}, \bibinfo{author}{Zhu, M.}, \bibinfo{author}{Zhmoginov, A.}, \bibinfo{author}{Chen, L.C.}, \bibinfo{year}{2018}.
\newblock \bibinfo{title}{Mobilenetv2: Inverted residuals and linear bottlenecks}, in: \bibinfo{booktitle}{2018 IEEE/CVF Conference on Computer Vision and Pattern Recognition}, pp. \bibinfo{pages}{4510--4520}.
\newblock \DOIprefix\doi{10.1109/CVPR.2018.00474}.
%Type = Article
\bibitem[{Santoso and Kusuma(2022)}]{ekosantoso2022facial}
\bibinfo{author}{Santoso, B.E.}, \bibinfo{author}{Kusuma, G.P.}, \bibinfo{year}{2022}.
\newblock \bibinfo{title}{Facial emotion recognition on fer2013 using vggspinalnet}.
\newblock \bibinfo{journal}{Journal of Theoretical and Applied Information Technology} \bibinfo{volume}{100}, \bibinfo{pages}{2088 – 2102}.
%Type = Article
\bibitem[{Sari et~al.(2020)Sari, Moussaouı and Hadid}]{sari2020automated}
\bibinfo{author}{Sari, M.}, \bibinfo{author}{Moussaouı, A.}, \bibinfo{author}{Hadid, A.}, \bibinfo{year}{2020}.
\newblock \bibinfo{title}{Automated facial expression recognition using deep learning techniques: An overview}.
\newblock \bibinfo{journal}{International Journal of Informatics and Applied Mathematics} \bibinfo{volume}{3}, \bibinfo{pages}{39--53}.
%Type = Inproceedings
\bibitem[{Selvaraju et~al.(2017)Selvaraju, Cogswell, Das, Vedantam, Parikh and Batra}]{selvaraju2017grad}
\bibinfo{author}{Selvaraju, R.R.}, \bibinfo{author}{Cogswell, M.}, \bibinfo{author}{Das, A.}, \bibinfo{author}{Vedantam, R.}, \bibinfo{author}{Parikh, D.}, \bibinfo{author}{Batra, D.}, \bibinfo{year}{2017}.
\newblock \bibinfo{title}{Grad-cam: Visual explanations from deep networks via gradient-based localization}, in: \bibinfo{booktitle}{2017 IEEE International Conference on Computer Vision (ICCV)}, \bibinfo{publisher}{IEEE}, \bibinfo{address}{New York}. pp. \bibinfo{pages}{618--626}.
\newblock \DOIprefix\doi{10.1109/ICCV.2017.74}.
%Type = Inproceedings
\bibitem[{Sharma et~al.(2020)Sharma, Joshi, Zeghari and Guerchouche}]{sharma2020audio-visual}
\bibinfo{author}{Sharma, G.}, \bibinfo{author}{Joshi, J.}, \bibinfo{author}{Zeghari, R.}, \bibinfo{author}{Guerchouche, R.}, \bibinfo{year}{2020}.
\newblock \bibinfo{title}{Audio-visual weakly supervised approach for apathy detection in the elderly}, in: \bibinfo{booktitle}{2020 International Joint Conference on Neural Networks (IJCNN)}, pp. \bibinfo{pages}{1--7}.
\newblock \DOIprefix\doi{10.1109/IJCNN48605.2020.9206829}.
%Type = Misc
\bibitem[{{Sighthound Inc}(2015)}]{Sighthound}
\bibinfo{author}{{Sighthound Inc}}, \bibinfo{year}{2015}.
\newblock \bibinfo{title}{Sighthound cloud api}.
\newblock \URLprefix \url{https://www.sighthound.com/products/alpr}. \bibinfo{note}{{Accessed 22nd Nov 2024}}.
%Type = Inproceedings
\bibitem[{Simonyan and Zisserman(2015)}]{simonyan2015very}
\bibinfo{author}{Simonyan, K.}, \bibinfo{author}{Zisserman, A.}, \bibinfo{year}{2015}.
\newblock \bibinfo{title}{Very deep convolutional networks for large-scale image recognition}, in: \bibinfo{booktitle}{3rd International Conference on Learning Representations (ICLR 2015)}, \bibinfo{publisher}{Computational and Biological Learning Society}. pp. \bibinfo{pages}{1--14}.
%Type = Inproceedings
\bibitem[{Song et~al.(2014)Song, Kim and Jeon}]{Song2014}
\bibinfo{author}{Song, I.}, \bibinfo{author}{Kim, H.J.}, \bibinfo{author}{Jeon, P.B.}, \bibinfo{year}{2014}.
\newblock \bibinfo{title}{Deep learning for real-time robust facial expression recognition on a smartphone}, in: \bibinfo{booktitle}{2014 IEEE International Conference on Consumer Electronics (ICCE)}, pp. \bibinfo{pages}{564--567}.
\newblock \DOIprefix\doi{10.1109/ICCE.2014.6776135}.
%Type = Article
\bibitem[{Sreevidya et~al.(2022)Sreevidya, Veni and Murthy}]{sreevidya2022elder}
\bibinfo{author}{Sreevidya, P.}, \bibinfo{author}{Veni, S.}, \bibinfo{author}{Murthy, O.V.R.}, \bibinfo{year}{2022}.
\newblock \bibinfo{title}{Elder emotion classification through multimodal fusion of intermediate layers and cross-modal transfer learning}.
\newblock \bibinfo{journal}{Signal Image Video Process.} \bibinfo{volume}{16}, \bibinfo{pages}{1281--1288}.
%Type = Inproceedings
\bibitem[{Szegedy et~al.(2015)Szegedy, Liu, Jia, Sermanet, Reed, Anguelov, Erhan, Vanhoucke and Rabinovich}]{GoogLeNet}
\bibinfo{author}{Szegedy, C.}, \bibinfo{author}{Liu, W.}, \bibinfo{author}{Jia, Y.}, \bibinfo{author}{Sermanet, P.}, \bibinfo{author}{Reed, S.}, \bibinfo{author}{Anguelov, D.}, \bibinfo{author}{Erhan, D.}, \bibinfo{author}{Vanhoucke, V.}, \bibinfo{author}{Rabinovich, A.}, \bibinfo{year}{2015}.
\newblock \bibinfo{title}{Going deeper with convolutions}, in: \bibinfo{booktitle}{2015 IEEE Conference on Computer Vision and Pattern Recognition (CVPR)}, pp. \bibinfo{pages}{1--9}.
%Type = Inproceedings
\bibitem[{Szegedy et~al.(2016)Szegedy, Vanhoucke, Ioffe, Shlens and Wojna}]{szegedy2016rethinking}
\bibinfo{author}{Szegedy, C.}, \bibinfo{author}{Vanhoucke, V.}, \bibinfo{author}{Ioffe, S.}, \bibinfo{author}{Shlens, J.}, \bibinfo{author}{Wojna, Z.}, \bibinfo{year}{2016}.
\newblock \bibinfo{title}{Rethinking the inception architecture for computer vision}, in: \bibinfo{booktitle}{2016 IEEE Conference on Computer Vision and Pattern Recognition (CVPR)}, pp. \bibinfo{pages}{2818--2826}.
\newblock \DOIprefix\doi{10.1109/CVPR.2016.308}.
%Type = Article
\bibitem[{Sönmez(2019)}]{battinisonmez2019computational}
\bibinfo{author}{Sönmez, E.B.}, \bibinfo{year}{2019}.
\newblock \bibinfo{title}{A computational study on aging effect for facial expression recognition}.
\newblock \bibinfo{journal}{Turkish Journal of Electrical Engineering and Computer Sciences} \bibinfo{volume}{27}, \bibinfo{pages}{2430--2443}.
\newblock \DOIprefix\doi{10.3906/elk-1811-70}.
%Type = Misc
\bibitem[{Tan and Le(2020a)}]{EfficientNet}
\bibinfo{author}{Tan, M.}, \bibinfo{author}{Le, Q.V.}, \bibinfo{year}{2020}a.
\newblock \bibinfo{title}{Efficientnet: Rethinking model scaling for convolutional neural networks}.
\newblock \DOIprefix\doi{10.48550/arXiv.1905.11946}, \href{http://arxiv.org/abs/1905.11946}{{\tt arXiv:1905.11946}}.
%Type = Misc
\bibitem[{Tan and Le(2020b)}]{tan2020efficientnet}
\bibinfo{author}{Tan, M.}, \bibinfo{author}{Le, Q.V.}, \bibinfo{year}{2020}b.
\newblock \bibinfo{title}{Efficientnet: Rethinking model scaling for convolutional neural networks}.
\newblock \DOIprefix\doi{10.48550/arXiv.1905.11946}.
%Type = Article
\bibitem[{Uddin et~al.(2017a)Uddin, Hassan, Almogren, Alamri, Alrubaian and Fortino}]{ziauddin2017facial2}
\bibinfo{author}{Uddin, M.Z.}, \bibinfo{author}{Hassan, M.M.}, \bibinfo{author}{Almogren, A.}, \bibinfo{author}{Alamri, A.}, \bibinfo{author}{Alrubaian, M.}, \bibinfo{author}{Fortino, G.}, \bibinfo{year}{2017}a.
\newblock \bibinfo{title}{Facial expression recognition utilizing local direction-based robust features and deep belief network}.
\newblock \bibinfo{journal}{IEEE Access} \bibinfo{volume}{5}, \bibinfo{pages}{4525--4536}.
\newblock \DOIprefix\doi{10.1109/ACCESS.2017.2676238}.
%Type = Article
\bibitem[{Uddin et~al.(2017b)Uddin, Hassan, Almogren, Zuair, Fortino and Torresen}]{ziauddin2017facial1}
\bibinfo{author}{Uddin, M.Z.}, \bibinfo{author}{Hassan, M.M.}, \bibinfo{author}{Almogren, A.}, \bibinfo{author}{Zuair, M.}, \bibinfo{author}{Fortino, G.}, \bibinfo{author}{Torresen, J.}, \bibinfo{year}{2017}b.
\newblock \bibinfo{title}{A facial expression recognition system using robust face features from depth videos and deep learning}.
\newblock \bibinfo{journal}{Computers \& Electrical Engineering} \bibinfo{volume}{63}, \bibinfo{pages}{114--125}.
\newblock \DOIprefix\doi{10.1016/j.compeleceng.2017.04.019}.
%Type = Inproceedings
\bibitem[{Ullah and Tian(2021)}]{ullah2021systematic}
\bibinfo{author}{Ullah, S.}, \bibinfo{author}{Tian, W.}, \bibinfo{year}{2021}.
\newblock \bibinfo{title}{A systematic literature review of recognition of compound facial expression of emotions}, in: \bibinfo{booktitle}{Proceedings of the 2020 4th International Conference on Video and Image Processing}, \bibinfo{publisher}{Association for Computing Machinery}. pp. \bibinfo{pages}{116--121}.
\newblock \DOIprefix\doi{10.1145/3447450.3447469}.
%Type = Inproceedings
\bibitem[{Vaswani et~al.(2017)Vaswani, Shazeer, Parmar, Uszkoreit, Jones, Gomez, Kaiser and Polosukhin}]{Transformer}
\bibinfo{author}{Vaswani, A.}, \bibinfo{author}{Shazeer, N.}, \bibinfo{author}{Parmar, N.}, \bibinfo{author}{Uszkoreit, J.}, \bibinfo{author}{Jones, L.}, \bibinfo{author}{Gomez, A.N.}, \bibinfo{author}{Kaiser, L.}, \bibinfo{author}{Polosukhin, I.}, \bibinfo{year}{2017}.
\newblock \bibinfo{title}{Attention is all you need}, in: \bibinfo{booktitle}{Proceedings of the 31st International Conference on Neural Information Processing Systems}, \bibinfo{publisher}{Curran Associates Inc.}, \bibinfo{address}{Red Hook, NY, USA}. p. \bibinfo{pages}{6000–6010}.
%Type = Article
\bibitem[{Vincent et~al.(2010)Vincent, Larochelle, Lajoie, Bengio and Manzagol}]{vincent2010stacked}
\bibinfo{author}{Vincent, P.}, \bibinfo{author}{Larochelle, H.}, \bibinfo{author}{Lajoie, I.}, \bibinfo{author}{Bengio, Y.}, \bibinfo{author}{Manzagol, P.A.}, \bibinfo{year}{2010}.
\newblock \bibinfo{title}{Stacked denoising autoencoders: Learning useful representations in a deep network with a local denoising criterion}.
\newblock \bibinfo{journal}{Journal of Machine Learning Research} \bibinfo{volume}{11}, \bibinfo{pages}{3371–3408}.
%Type = Inproceedings
\bibitem[{Viola and Jones(2001)}]{viola2001rapid}
\bibinfo{author}{Viola, P.}, \bibinfo{author}{Jones, M.}, \bibinfo{year}{2001}.
\newblock \bibinfo{title}{Rapid object detection using a boosted cascade of simple features}, in: \bibinfo{booktitle}{Proceedings of the 2001 IEEE Computer Society Conference on Computer Vision and Pattern Recognition. CVPR 2001}, pp. \bibinfo{pages}{I--I}.
\newblock \DOIprefix\doi{10.1109/CVPR.2001.990517}.
%Type = Inproceedings
\bibitem[{Vyas et~al.(2019)Vyas, Prajapati and Dabhi}]{svyas2019survey}
\bibinfo{author}{Vyas, A.S.}, \bibinfo{author}{Prajapati, H.B.}, \bibinfo{author}{Dabhi, V.K.}, \bibinfo{year}{2019}.
\newblock \bibinfo{title}{Survey on face expression recognition using cnn}, in: \bibinfo{booktitle}{2019 5th International Conference on Advanced Computing \& Communication Systems (ICACCS)}, pp. \bibinfo{pages}{102--106}.
\newblock \DOIprefix\doi{10.1109/ICACCS.2019.8728330}.
%Type = Inproceedings
\bibitem[{Wang et~al.(2014)Wang, Subagdja, Kang, Tan and Zhang}]{wang2014towards}
\bibinfo{author}{Wang, D.}, \bibinfo{author}{Subagdja, B.}, \bibinfo{author}{Kang, Y.}, \bibinfo{author}{Tan, A.H.}, \bibinfo{author}{Zhang, D.}, \bibinfo{year}{2014}.
\newblock \bibinfo{title}{Towards intelligent caring agents for aging-in-place: Issues and challenges}, in: \bibinfo{booktitle}{2014 IEEE Symposium on Computational Intelligence for Human-like Intelligence (CIHLI)}, pp. \bibinfo{pages}{1--8}.
\newblock \DOIprefix\doi{10.1109/CIHLI.2014.7013393}.
%Type = Article
\bibitem[{Wang et~al.(2015)Wang, Wu, Gao and Ji}]{wang2015facial}
\bibinfo{author}{Wang, S.}, \bibinfo{author}{Wu, S.}, \bibinfo{author}{Gao, Z.}, \bibinfo{author}{Ji, Q.}, \bibinfo{year}{2015}.
\newblock \bibinfo{title}{Facial expression recognition through modeling age-related spatial patterns}.
\newblock \bibinfo{journal}{Multimedia Tools and Applications} \bibinfo{volume}{75}, \bibinfo{pages}{3937--3954}.
\newblock \DOIprefix\doi{10.1007/s11042-015-3107-2}.
%Type = Inproceedings
\bibitem[{Wang et~al.(2021)Wang, Xie, Li, Fan, Song, Liang, Lu, Luo and Shao}]{wang2021pyramid}
\bibinfo{author}{Wang, W.}, \bibinfo{author}{Xie, E.}, \bibinfo{author}{Li, X.}, \bibinfo{author}{Fan, D.P.}, \bibinfo{author}{Song, K.}, \bibinfo{author}{Liang, D.}, \bibinfo{author}{Lu, T.}, \bibinfo{author}{Luo, P.}, \bibinfo{author}{Shao, L.}, \bibinfo{year}{2021}.
\newblock \bibinfo{title}{Pyramid vision transformer: A versatile backbone for dense prediction without convolutions}, in: \bibinfo{booktitle}{2021 IEEE/CVF International Conference on Computer Vision (ICCV)}, pp. \bibinfo{pages}{548--558}.
\newblock \DOIprefix\doi{10.1109/ICCV48922.2021.00061}.
%Type = Misc
\bibitem[{WHO(2024)}]{WHO2024}
\bibinfo{author}{WHO}, \bibinfo{year}{2024}.
\newblock \bibinfo{title}{World health organization report 2024}.
\newblock \bibinfo{howpublished}{\url{https://2024-wpds.prb.org/}}.
%Type = Inproceedings
\bibitem[{Wu et~al.(2021)Wu, Xiao, Codella, Liu, Dai, Yuan and Zhang}]{wu2021cvt}
\bibinfo{author}{Wu, H.}, \bibinfo{author}{Xiao, B.}, \bibinfo{author}{Codella, N.}, \bibinfo{author}{Liu, M.}, \bibinfo{author}{Dai, X.}, \bibinfo{author}{Yuan, L.}, \bibinfo{author}{Zhang, L.}, \bibinfo{year}{2021}.
\newblock \bibinfo{title}{Cvt: Introducing convolutions to vision transformers}, in: \bibinfo{booktitle}{2021 IEEE/CVF International Conference on Computer Vision (ICCV)}, pp. \bibinfo{pages}{22--31}.
\newblock \DOIprefix\doi{10.1109/ICCV48922.2021.00009}.
%Type = Inproceedings
\bibitem[{Wu et~al.(2015)Wu, Wang and Wang}]{wu2015enhanced}
\bibinfo{author}{Wu, S.}, \bibinfo{author}{Wang, S.}, \bibinfo{author}{Wang, J.}, \bibinfo{year}{2015}.
\newblock \bibinfo{title}{Enhanced facial expression recognition by age}, in: \bibinfo{booktitle}{2015 11th IEEE International Conference and Workshops on Automatic Face and Gesture Recognition (FG)}, pp. \bibinfo{pages}{1--6}.
\newblock \DOIprefix\doi{10.1109/FG.2015.7163117}.
%Type = Article
\bibitem[{Wu et~al.(2012)Wu, Turaga and Chellappa}]{wu2012age}
\bibinfo{author}{Wu, T.}, \bibinfo{author}{Turaga, P.}, \bibinfo{author}{Chellappa, R.}, \bibinfo{year}{2012}.
\newblock \bibinfo{title}{Age estimation and face verification across aging using landmarks}.
\newblock \bibinfo{journal}{IEEE Transactions on Information Forensics and Security} \bibinfo{volume}{7}, \bibinfo{pages}{1780--1788}.
\newblock \DOIprefix\doi{10.1109/TIFS.2012.2213812}.
%Type = Inproceedings
\bibitem[{Yan et~al.(2020)Yan, Xu, Klopfer and Nürnberger}]{yan2020deep}
\bibinfo{author}{Yan, C.}, \bibinfo{author}{Xu, J.}, \bibinfo{author}{Klopfer, B.}, \bibinfo{author}{Nürnberger, A.}, \bibinfo{year}{2020}.
\newblock \bibinfo{title}{A deep neural network based multimodal video recognition system for caring}, in: \bibinfo{booktitle}{2020 IEEE International Conference on Human-Machine Systems (ICHMS)}, pp. \bibinfo{pages}{1--5}.
\newblock \DOIprefix\doi{10.1109/ICHMS49158.2020.9209395}.
%Type = Article
\bibitem[{Yang et~al.(2018)Yang, Lin, Chang and Chen}]{yang2018joint}
\bibinfo{author}{Yang, H.F.}, \bibinfo{author}{Lin, B.Y.}, \bibinfo{author}{Chang, K.Y.}, \bibinfo{author}{Chen, C.S.}, \bibinfo{year}{2018}.
\newblock \bibinfo{title}{Joint estimation of age and expression by combining scattering and convolutional networks}.
\newblock \bibinfo{journal}{ACM Trans. Multimedia Comput. Commun. Appl.} \bibinfo{volume}{14}.
\newblock \DOIprefix\doi{10.1145/3152118}.
%Type = Article
\bibitem[{Yang et~al.(2020)Yang, Yang, Xu, Gao, Zhang, Wang, Liu, Han, Zhu, Tian, Huang, Zhao, Zhong, Shi, Li, Fu, Liang, Banissy and Sun}]{yang2020tsinghua}
\bibinfo{author}{Yang, T.}, \bibinfo{author}{Yang, Z.}, \bibinfo{author}{Xu, G.}, \bibinfo{author}{Gao, D.}, \bibinfo{author}{Zhang, Z.}, \bibinfo{author}{Wang, H.}, \bibinfo{author}{Liu, S.}, \bibinfo{author}{Han, L.}, \bibinfo{author}{Zhu, Z.}, \bibinfo{author}{Tian, Y.}, \bibinfo{author}{Huang, Y.}, \bibinfo{author}{Zhao, L.}, \bibinfo{author}{Zhong, K.}, \bibinfo{author}{Shi, B.}, \bibinfo{author}{Li, J.}, \bibinfo{author}{Fu, S.}, \bibinfo{author}{Liang, P.}, \bibinfo{author}{Banissy, M.J.}, \bibinfo{author}{Sun, P.}, \bibinfo{year}{2020}.
\newblock \bibinfo{title}{Tsinghua facial expression database – a database of facial expressions in chinese young and older women and men: Development and validation}.
\newblock \bibinfo{journal}{PLOS ONE} \bibinfo{volume}{15}, \bibinfo{pages}{1--14}.
\newblock \DOIprefix\doi{10.1371/journal.pone.0231304}.
%Type = Article
\bibitem[{Zhalehpour et~al.(2017)Zhalehpour, Onder, Akhtar and Erdem}]{zhalehpour2017baum-1}
\bibinfo{author}{Zhalehpour, S.}, \bibinfo{author}{Onder, O.}, \bibinfo{author}{Akhtar, Z.}, \bibinfo{author}{Erdem, C.E.}, \bibinfo{year}{2017}.
\newblock \bibinfo{title}{Baum-1: A spontaneous audio-visual face database of affective and mental states}.
\newblock \bibinfo{journal}{IEEE Transactions on Affective Computing} \bibinfo{volume}{8}, \bibinfo{pages}{300--313}.
\newblock \DOIprefix\doi{10.1109/TAFFC.2016.2553038}.
%Type = Article
\bibitem[{Zhang et~al.(2020)Zhang, Huang and Tian}]{zhang2020facial}
\bibinfo{author}{Zhang, H.}, \bibinfo{author}{Huang, B.}, \bibinfo{author}{Tian, G.}, \bibinfo{year}{2020}.
\newblock \bibinfo{title}{Facial expression recognition based on deep convolution long short-term memory networks of double-channel weighted mixture}.
\newblock \bibinfo{journal}{Pattern Recognition Letters} \bibinfo{volume}{131}, \bibinfo{pages}{128 – 134}.
\newblock \DOIprefix\doi{10.1016/j.patrec.2019.12.013}.
%Type = Article
\bibitem[{Zhang et~al.(2016)Zhang, Zhang, Li and Qiao}]{zhang2016joint}
\bibinfo{author}{Zhang, K.}, \bibinfo{author}{Zhang, Z.}, \bibinfo{author}{Li, Z.}, \bibinfo{author}{Qiao, Y.}, \bibinfo{year}{2016}.
\newblock \bibinfo{title}{Joint face detection and alignment using multitask cascaded convolutional networks}.
\newblock \bibinfo{journal}{IEEE Signal Processing Letters} \bibinfo{volume}{23}, \bibinfo{pages}{1499--1503}.
\newblock \DOIprefix\doi{10.1109/LSP.2016.2603342}.
%Type = Article
\bibitem[{Zhao et~al.(2011)Zhao, Huang, Taini, Li and Pietikäinen}]{zhao2011facial}
\bibinfo{author}{Zhao, G.}, \bibinfo{author}{Huang, X.}, \bibinfo{author}{Taini, M.}, \bibinfo{author}{Li, S.Z.}, \bibinfo{author}{Pietikäinen, M.}, \bibinfo{year}{2011}.
\newblock \bibinfo{title}{Facial expression recognition from near-infrared videos}.
\newblock \bibinfo{journal}{Image and Vision Computing} \bibinfo{volume}{29}, \bibinfo{pages}{607--619}.
\newblock \DOIprefix\doi{10.1016/j.imavis.2011.07.002}.
%Type = Article
\bibitem[{Zhu et~al.(2024)Zhu, Boonipat, Cherukuri and Bite}]{zhu2024defining}
\bibinfo{author}{Zhu, A.}, \bibinfo{author}{Boonipat, T.}, \bibinfo{author}{Cherukuri, S.}, \bibinfo{author}{Bite, U.}, \bibinfo{year}{2024}.
\newblock \bibinfo{title}{Defining standard values for facereader facial expression software output}.
\newblock \bibinfo{journal}{Aesthetic Plastic Surgery} \bibinfo{volume}{48}, \bibinfo{pages}{785--792}.
\newblock \DOIprefix\doi{10.1007/s00266-023-03468-y}.

\end{thebibliography}

%% else use the following coding to input the bibitems directly in the
%% TeX file.

%% Refer following link for more details about bibliography and citations.
%% https://en.wikibooks.org/wiki/LaTeX/Bibliography_Management

% \begin{thebibliography}{00}

%% For authoryear reference style
% \bibitem[Author(year)]{label}
%% Text of bibliographic item

% \bibitem[Lamport(1994)]{lamport94}
%   Leslie Lamport,
%   \textit{\LaTeX: a document preparation system},
%   Addison Wesley, Massachusetts,
%   2nd edition,
%   1994.

% \end{thebibliography}
\end{document}